\documentclass[10pt,twocolumn,letterpaper]{article}

\usepackage[pagenumbers]{cvpr} 
\usepackage[many]{tcolorbox}  
\usepackage{xltabular}    
\usepackage{graphicx}
\usepackage{listings}
\usepackage{booktabs}    
\usepackage{caption}
\usepackage{adjustbox}    
\usepackage{CJKutf8}    

\usepackage{tabularx}

\usepackage{siunitx}
\usepackage{amsmath}
\usepackage{float}
\usepackage{array}
\usepackage{multirow}
\usepackage{placeins}
\usepackage{ragged2e}
\definecolor{cvprblue}{rgb}{0.21,0.49,0.74}
\usepackage[pagebackref,breaklinks,colorlinks,allcolors=cvprblue]{hyperref}

\title{ProactiveMobile: A Comprehensive Benchmark for Boosting Proactive Intelligence on Mobile Devices}

\author{
Dezhi Kong$^{1}$\thanks{These authors contributed equally.} \quad Zhengzhao Feng$^{1,2}$\footnotemark[1] \quad Qiliang Liang$^{1,3}$\footnotemark[1] \quad
Hao Wang$^{1}$ \quad Haofei Sun$^{1}$ \quad Changpeng Yang$^{1}$ \\
Yang Li$^{1}$ \quad Peng Zhou$^{1}$ \quad Shuai Nie$^{1}$ \quad
Hongzhen Wang$^{1}$ \quad Linfeng Zhou$^{1,4}$ \\
Hao Jia$^{1}$ \quad Jiaming Xu$^{1}$ \quad
Runyu Shi$^{1}$\thanks{Corresponding authors.} \quad Ying Huang$^{1}$ \\
\normalsize\texttt\{kongdezhi1, zhoupeng11, nieshuai, xujiaming1, shirunyu\}@xiaomi.com \quad \\
{\normalsize
$^{1}$HyperAI Team, Xiaomi Corporation \quad
$^{2}$Zhejiang University \quad
$^{3}$Peking University \quad
$^{4}$Northeastern University \quad
}
}

\begin{document}
\maketitle
\begin{abstract}
Multimodal large language models (MLLMs) have made significant progress in mobile agent development, yet their capabilities are predominantly confined to a reactive paradigm, where they merely execute explicit user commands.
The emerging paradigm of \textbf{proactive intelligence}, where agents autonomously anticipate needs and initiate actions, represents the next frontier for mobile agents. 
However, its development is critically bottlenecked by the lack of benchmarks that can address real-world complexity and enable objective, executable evaluation. To overcome these challenges, we introduce \textbf{ProactiveMobile}, a comprehensive benchmark designed to systematically advance research in this domain. 
ProactiveMobile formalizes the proactive task as inferring latent user intent across four dimensions of on-device contextual signals and generating an executable function sequence from a comprehensive function pool of 63 APIs. The benchmark features over 3,660 instances of 14 scenarios that embrace real-world complexity through multi-answer annotations. To ensure quality, a team of 30 experts conducts a final audit of the benchmark, verifying factual accuracy, logical consistency, and action feasibility, and correcting non-compliant entries.
Extensive experiments demonstrate that our fine-tuned Qwen2.5-VL-7B-Instruct achieves a success rate of 20.82\%, outperforming o1 (17.02\%) and GPT-5 (11.37\%). This result indicates that proactivity is a critical competency widely lacking in current MLLMs, yet it is learnable, emphasizing the importance of the proposed benchmark for proactivity evaluation. Data and code are available at \small \url{https://github.com/xiaomi-research/proactive-mobile}.

\end{abstract}
\section{Introduction}
\begin{figure*}[!h]
    \centering 
    \includegraphics[width=0.8\textwidth, angle=0]{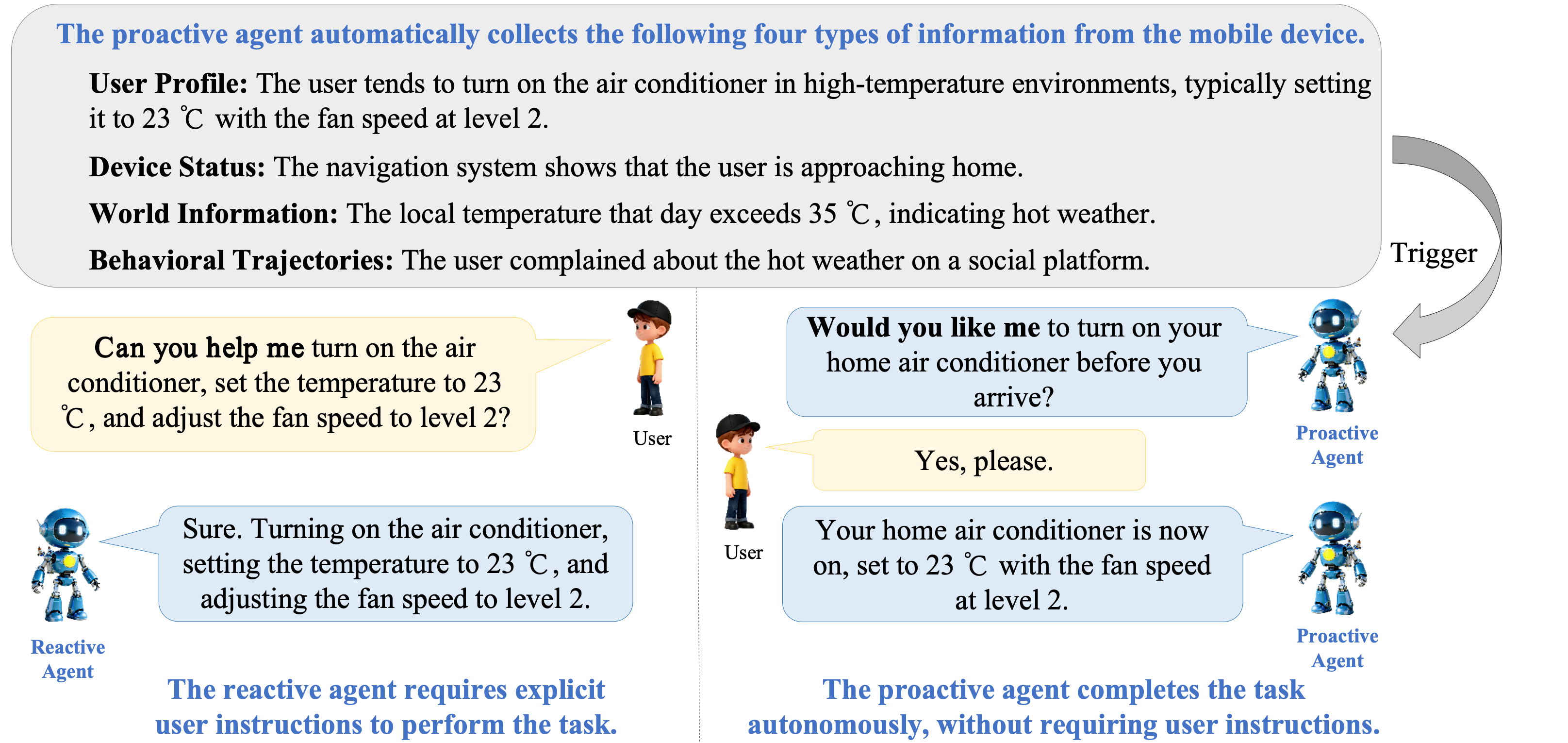}
    \vspace{-7pt}
    \caption{A comparison of proactive and reactive paradigms in mobile agents.}
    \vspace{-17pt}
    \label{fig:proactive_intelligence}    
\end{figure*}

Fueled by rapid advancements in MLLMs \citep{yin2024survey, zhang2024mmllm, bai2025qwen25vl}, mobile agents have achieved substantial breakthroughs \citep{yao2025surveyagenticmultimodallarge,hu2025agents} such as interface comprehension \citep{gou2025navigatingdigitalworldhumans,zhou2025gui}, conversational interaction \citep{wang2024mobileagentv2,li2024appagent}, and task planning \citep{deng-etal-2024-mobile,wang2025mobileagente}.

However, these agents share a fundamental constraint: they are confined to a reactive paradigm, functioning as passive executors of direct user commands \citep{tang2025surveymllmbasedguiagents, deng2025proactive}. These models place the entire cognitive burden on the user, from need identification to goal articulation \citep{peng2025navigating}, thereby relegating the agent to the role of a high-level tool and fundamentally limiting its potential for seamless integration into daily life \citep{liu2025proactiveconversationalagent}.

The limitations of the reactive paradigm are becoming a critical bottleneck, propelling a fundamental shift towards \textbf{proactive intelligence}—the undisputed next frontier for mobile agents. This vision represents not merely an incremental improvement, but a complete re-imagining of the agent's role: rather than being a passive tool, it evolves into a genuinely helpful assistant by autonomously anticipating user needs and initiating actions \citep{lu2024proactiveagentthu,yang2025contextagentcontextawareproactivellm,yang2025fingertip}. The profound implication is a future of human-agent collaboration where cognitive burden is minimized, and interaction feels seamlessly intuitive \citep{brachten2020ability, yang2025contextagentcontextawareproactivellm, peng2025navigating}. Recognizing this transformative potential, pioneering studies have indeed validated the core premise of proactivity \citep{deng2023surveyproactivedialoguesystems,lu2024proactiveagentthu,yang2025contextagentcontextawareproactivellm,yang2025fingertip}. 

Despite these promising initial steps, the current research landscape for proactive agents remains fragmented and lacks a unified foundation. 
A core deficiency is that existing benchmarks \citep{lu2024proactiveagentthu,yang2025fingertip} oversimplify the task: they rely on abstracted contexts and crucially assume a single ``correct'' action per scenario. This ignores the inherent subjectivity and diversity of user preferences, forcing the complex one-to-many mapping of proactive suggestions into an unrealistic one-to-one paradigm.
This flawed premise is exacerbated by the metrics used for evaluation.
For instance, \textit{ProactiveAgent}'s \citep{lu2024proactiveagentthu} binary reward model is too coarse to differentiate partial from complete failures, while \textit{FingerTip-20K} \citep{yang2025fingertip} relies on cosine similarity, which captures semantic relevance but ignores functional correctness and executability.
Beyond definition and evaluation, a third major shortcoming lies in the output format. Both benchmarks rely on generating natural language recommendations, a format that is inherently ambiguous and lacks a direct path to on-device execution, creating a critical gap between suggesting a task and actually performing it.
This confluence of issues (an ill-defined task rooted in oversimplification, superficial evaluation, and a non-executable output format) critically bottlenecks the systematic advancement of the field.

To address these critical gaps, we introduce \textbf{ProactiveMobile}, a comprehensive benchmark designed to systematically advance research on proactive mobile agents. 
To mitigate oversimplification, ProactiveMobile formalizes the proactive task by requiring agents to predict actions based on four dimensions of on-device contextual signals: user profile, device status, world information, and behavioral trajectories. Figure \ref{fig:proactive_intelligence} clearly illustrates this process and contrasts it with the reactive agent paradigm.
To tackle the ignorance of user preferences, ProactiveMobile embraces the one-to-many nature of proactivity: each instance is annotated and manually verified with one to three target actions. The resulting benchmark is substantial, comprising 3,660 instances of 14 scenarios spanning a diverse range of real-world scenarios. Furthermore, to overcome the inherent ambiguity and subjectivity of evaluating natural language suggestions, we introduce a crucial constraint: models must translate their intents into executable actions. We achieve this by constructing a comprehensive function pool of 63 APIs, requiring models to output specific function sequences. This approach transforms the evaluation from a subjective text-matching problem into an objective, structured task.

To establish baselines on our benchmark, we fine-tuned Qwen2.5-VL-7B-Instruct \citep{bai2025qwen25vl} and MiMo-VL-7B-SFT-2508 \citep{coreteam2025mimovltechnicalreport} on the training set. We then evaluated their performance on ProactiveMobile alongside a suite of leading closed-source models, including o1 \citep{jaech2024openaio1}, GPT-5 \citep{openai_gpt5_systemcard}, and Gemini-2.5-Pro \citep{comanici2025gemini}. Our fine-tuned Qwen2.5-VL-7B-Instruct achieves a success rate of 20.82\% on the exact function sequence matching, significantly outperforming closed‑source models, including o1 (17.02\%), GPT-5 (11.37\%), and Gemini-2.5-Pro (9.62\%).

These results offer two critical insights. First, this superior performance supports our hypothesis: proactivity is a specialized capability that requires targeted, domain-specific training, as provided by ProactiveMobile. Even the most powerful general-purpose models fail to master it out-of-the-box. Second, although proactivity is learnable, the performance of trained models still fails to meet the requirements for on-device deployment. This indicates that proactive intelligence is a highly challenging research problem, which in turn underscores the significance of our work and the necessity of the proposed benchmark.
Our contributions are as follows:
\begin{enumerate}
    \item We propose a novel and comprehensive task formalization for proactive mobile agents, grounding the problem in rich, multi-dimensional real-world context.
    \item We construct and open-source ProactiveMobile, comprising 3,660 multi-intent instances across 14 scenarios. To facilitate deployment and fine-grained evaluation, all intents are mapped to corresponding function sequences via a predefined function pool of 63 APIs.
    \item We provide an in-depth empirical analysis, establishing strong baselines and revealing that proactivity is a specialized capability lacking in current general models, thereby highlighting critical challenges and future research directions. Notably, we will release our model weights to foster progress within the research community.
\end{enumerate}
\section{Related Work}
\label{sec:related_work}

\subsection{LLM-Based Mobile Interaction}

The advent of MLLMs has ushered in a new era of mobile agents capable of understanding natural language instructions and visual UI elements to perform actions autonomously. These LLM-based mobile agents represent a paradigm shift, enabling users to accomplish intricate, multi-step tasks on web, mobile, or desktop applications through simple conversational commands \citep{zhang2025largelanguagemodelbrainedgui,tang2025surveymllmbasedguiagents}.

A core capability of these agents is \textbf{GUI understanding}, or GUI grounding, where MLLMs interpret screen layouts by combining visual perception with textual information. To enhance this, specialized models \citep{wu2024osatlasfoundationactionmodel,yang2024aria, wang2025mp} and methods \citep{gou2025navigatingdigitalworldhumans,zhou2025gui, tang2025guig2gaussianrewardmodeling,chen2025v2pbackgroundsuppressioncenter} have been developed to better process GUI-specific modalities. The rapid progress in this area is also fueled by the development of specialized datasets, including large-scale annotated datasets \citep{wu2024osatlasfoundationactionmodel, hui2025winspot,li2025screenspot, li2025autogui, liu2025uie2isynthadvancingguigrounding} and data pipelines \citep{yang2024aria, li2025autogui, liu2025uie2isynthadvancingguigrounding}.

Another critical area is \textbf{task planning and execution}. LLMs excel at decomposing high-level natural language commands into a series of executable actions. However, methods based on static prompting often struggle with long-horizon tasks and dynamic environments \citep{zhang2024dynamic, tang2025surveymllmbasedguiagents, xie2025mirage1augmentingupdatinggui,wu2026atlas}. Some research explores fine-tuning or reinforcement learning to enhance the reasoning and prediction capabilities of MLLMs in related tasks \citep{wu2024beyond,CAGUI,luo2025guir1generalistr1style, lu2025ui,gu2025uivenustechnicalreportbuilding}.
The maturation of the field is also marked by comprehensive benchmarks \citep{gui_odyssey,AITZ,zhao2025worldgui,CAGUI}, which provide standardized environments for evaluating agent performance on realistic tasks.

\subsection{Proactive Agents}

The paradigm of intelligent agents is undergoing a significant shift, moving from reactive systems that await explicit user commands to proactive agents that anticipate user needs \citep{deng2023surveyproactivedialoguesystems}. By inferring likely intentions and preemptively offering or executing useful actions, these agents can enhance user engagement and task efficiency \citep{tabalba2024articulateprocomparativestudyproactive}. Research in proactivity has evolved through several distinct stages.

Initial explorations in this domain have largely focused on proactive conversational agents. Instead of passively responding, these systems actively guide the dialogue by asking clarifying questions, suggesting relevant topics, or steering the conversation towards a productive goal \citep{liao2023proactive,deng2023surveyproactivedialoguesystems,deng2025proactive}. While foundational, their proactivity is primarily confined to the conversational level.

Building on this, subsequent research has delved deeper into proactive intent inference, where the agent's goal is to predict a user's next action or ultimate goal from their behavior. These approaches can be broadly categorized into two types: those that explicitly prompt the user for clarification to confirm intent \citep{qian2024tell, zhang2024ask}, and those that implicitly infer intent from contextual cues and behavioral history \citep{kim2024auto,yang2025fingertip}. This line of work is crucial for understanding user needs before they are articulated.

The most advanced form of proactivity involves agents that not only anticipate needs but also autonomously execute or propose complete tasks. This represents the ultimate goal of delivering value to the user with minimal friction. However, existing work in this advanced stage often faces significant limitations. Some studies are confined to narrow, specific domains like smart home control, limiting their generalizability \citep{cao2024smart}. Others predict overly simplistic, single-step tasks, often within simulated or artificial scenarios that do not capture the nuances of genuine user interactions \citep{lu2024proactiveagentthu,yang2025fingertip}. Our work addresses these gaps by introducing a benchmark where the data is deeply grounded in diverse, realistic scenarios, designed to evaluate an agent's ability to recommend complex, multi-step tasks.
\section{Benchmark}

\begin{figure*}[!ht]
	\centering 
	\includegraphics[width=0.83\textwidth, angle=0]{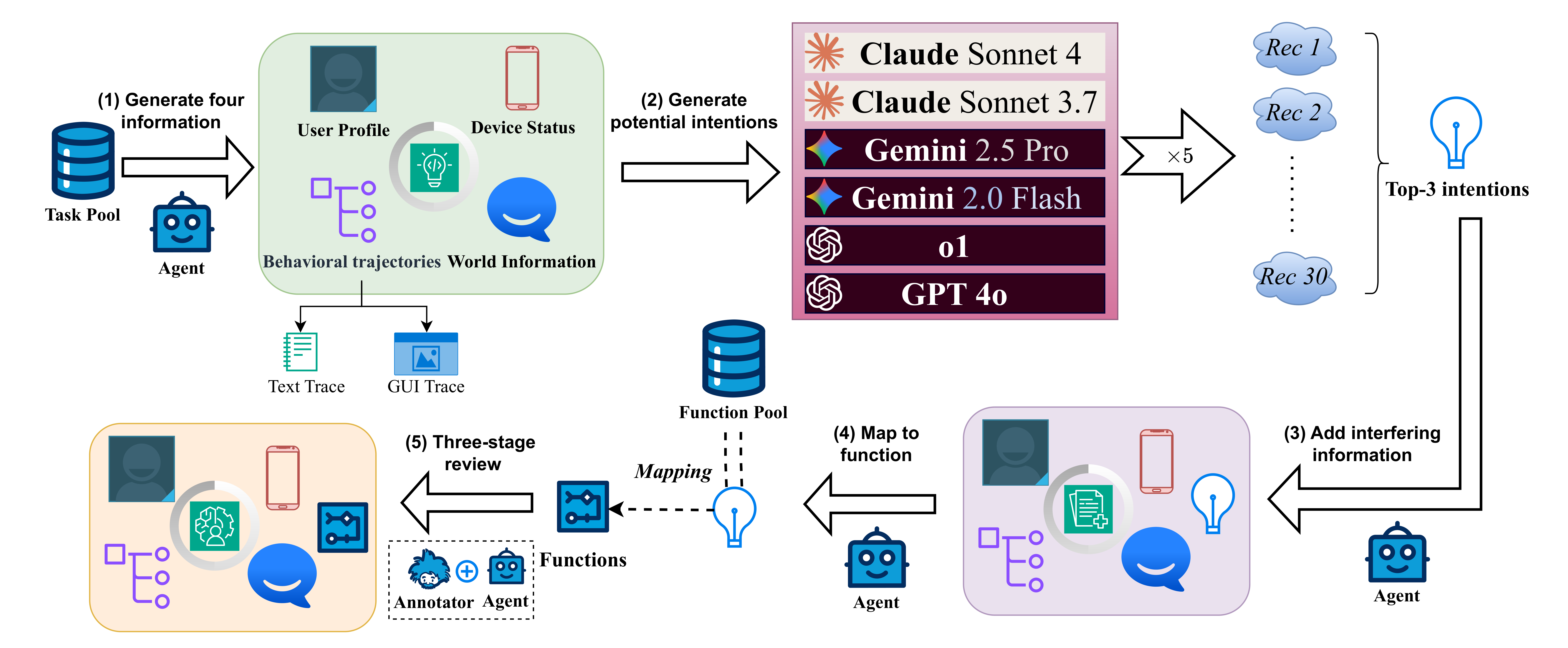}
    \vspace{-7pt}
	\caption{The overview of data generation.}
    \vspace{-15pt}
	\label{fig:overview}
\end{figure*}

In this section, we define the task and detail the benchmark construction process. Due to space limitations, we provide comprehensive implementation details in 
the Appendix, including the prompt templates used for data generation, the 
design of the annotation platform, annotator training materials, annotation 
guidelines, quality control procedures, and other technical specifications.

\subsection{Task Definition}
We define \textbf{proactive intelligence} as the task of predicting users' latent intents based on their user profile, device status, world information, and behavioral trajectories. The details of these four categories are as follows:
\begin{itemize}
    \item \textbf{User Profile.} The user’s static attributes and dynamic behavioral characteristics encompass basic information, long-term behavioral habits, and personal preferences.
    \item \textbf{Device Status.} Real-time device and immediate environmental states include hardware, battery level, network status, location, and notifications.
    \item \textbf{World Information.} External circumstances, including weather, time of day, and public holidays. 
    \item \textbf{Behavioral Trajectories.} A temporal sequence of user-device interactions that reveals evolving intent.
\end{itemize}

In terms of representation, user profile, device status, and world information are expressed in natural language, while behavioral trajectories are represented either as textual descriptions or sequences of GUI screenshots. 
To facilitate command execution, all intents are mapped into a unified sequence of executable functions. 
Consequently, the complete proactive intelligence task can be formalized as:
\begin{equation}
   \mathcal{T}=\{(\pmb{I}_k, \pmb{F}_k)\}_{k=1}^{a} 
= \text{Predict}(\mathbf{U}, \mathbf{D}, \mathbf{W}, \mathbf{B}).
\end{equation}
Here, $\mathbf{U}$, $\mathbf{D}$, $\mathbf{W}$, and $\mathbf{B}$ represent the user profile, device status, world information, and behavioral trajectories at the decision moment. 
For each decision point, there may exist multiple valid intent–function pairs, denoted as the ground-truth set $\mathcal{T}$.
The model generates a single predicted pair:
\begin{equation}
\begin{aligned}
&(\hat{\pmb{I}}, \hat{\pmb{F}}) = \pmb{M}_\theta(\mathbf{U}, \mathbf{D}, \mathbf{W}, \mathbf{B}), \\
   &\hat{\pmb{F}} = 
    \begin{cases}
    (f_1,\dots,f_n),\quad \hat{\pmb{I}}\neq\varnothing\land\hat{\pmb{I}}\Rightarrow\hat{\pmb{F}}, \\
    \varnothing,\quad \hat{\pmb{I}}=\varnothing\lor\hat{\pmb{I}}\nRightarrow\hat{\pmb{F}},
    \end{cases}
\end{aligned}
\end{equation}
where $\hat{\pmb{F}}$ is non-empty only if $\hat{\pmb{I}}$ is actionable and can be mapped to at least one function from the predefined function pool $\mathbb{F}$, i.e., $\hat{\pmb{F}} \subseteq \mathbb{F}$; otherwise, $\hat{\pmb{F}} = \varnothing$.  

The prediction is considered correct if the model output matches any ground-truth pair:
\begin{equation}
(\hat{\pmb{I}}, \hat{\pmb{F}}) \in \mathcal{T}.
\end{equation}

\subsection{Dataset Construction}
This section is organized into three main parts. First, we elaborate on the acquisition methods for behavioral trajectories. Second, we outline the end-to-end data generation process. Finally, we provide a detailed account of the data auditing mechanism.

\subsubsection{Acquisition of Behavioral Trajectories}

User behavioral trajectories serve as the foundation for intent prediction. In cases where direct access to user actions is unavailable, screenshots are used as substitutes. We define these two modalities as:
\begin{itemize}
    \item \textbf{Multimodal trajectories:} Sequences of mobile screenshots captured during user interactions, combined with corresponding text commands from both public and self-built datasets, summarized in Table~\ref{tb:gui_datasets}.
    \item \textbf{Text trajectories:} Textual logs of user actions are derived from GUI traces. We first categorize and deduplicate text commands, then employ Claude-Sonnet-4 to generate text-based action trajectories via prompt-based expansion. 
\end{itemize}

\begin{table}[h!]
    \small
    \centering
    \scalebox{0.75}{
    \begin{tabular}{c c c}
        \hline
        Dataset & Description & Usage\\
        \hline
         GUI-Odyssey~\cite{gui_odyssey} & A dataset for cross-app mobile GUI navigation & 21,669\\
         AITZ~\cite{AITZ} & Largest dataset in the Android GUI navigation field & 9,413 \\
         CAGUI~\cite{CAGUI} & A real-world Chinese Android GUI benchmark & 3,196 \\
         MobileAgentBench & Self-collected GUI data from mobile devices & 12,481 \\
        \hline
    \end{tabular}
    }
    \vspace{-5pt}
    \caption{Summary of GUI datasets.}
    \vspace{-15pt}
    \label{tb:gui_datasets}
\end{table}

\subsubsection{Generation Pipeline}

The data generation process involves five key steps, as illustrated in Figure~\ref{fig:overview}.

\begin{enumerate}
    \item \textbf{Generate contextual information.} Based on user behavioral trajectories and relevant commands, we randomly employ Claude-Sonnet-4\citep{claude4-anthropic-2025}, Gemini-2.5-Pro \citep{comanici2025gemini}, and GPT-5 \citep{openai_gpt5_systemcard} to generate three complementary information components—user profile, device status, and world information—thereby constructing a comprehensive contextual information stream. The generated information then undergoes a plausibility check using the o1 \citep{jaech2024openaio1}. If the content is deemed implausible, it is discarded and subsequently regenerated.
    \item \textbf{Generate potential intentions.} Leveraging the contextual information, multiple MLLM models simulate users' potential next-step intentions. These intentions represent tasks that agents can recommend proactively, triggered by specific conditions and personalized according to the user profile. To ensure both diversity and quality of generated intentions, we select six state-of-the-art closed-source MLLMs (Claude-Sonnet-4 \citep{claude4-anthropic-2025}, Claude-Sonnet-3.7 \citep{claude37-anthropic-2025}, Gemini-2.5-Pro \citep{comanici2025gemini}, Gemini-2.0-Flash \citep{gemini20flash-google-2025}, o1 \citep{jaech2024openaio1}, and GPT-4o \citep{hurst2024gpt4o}) with strong multimodal understanding and reasoning capabilities. 
    To unify their outputs, Gemini-2.5-Flash \citep{comanici2025gemini} is prompted to semantically cluster the 30 generated candidates and extract the top-3 representative intentions (cluster centroids) ranked by overall support across models.
    \item \textbf{Add interfering information.} To enhance model robustness, we intentionally inject irrelevant textual noise into the user profile, device states, and environmental information. The injected noise consists of task-irrelevant yet semantically coherent text generated by Gemini-2.5-Pro \citep{comanici2025gemini} through carefully designed prompts. This process preserves overall logical consistency while training the model to focus on salient and task-relevant signals. On average, the amount of injected noise is approximately 5–20 times the volume of task-relevant information.
    \item \textbf{Map to function.} Convert intentions into function calls. We uniformly convert textual instructions generated by multiple MLLMs into executable function-call sequences. This conversion is performed by Claude-Sonnet-4 \citep{claude4-anthropic-2025}, which is prompted to select appropriate functions from a predefined function pool to fulfill each recommended task. The resulting sequence may include one or more functions, while a zero-function sequence indicates that no action is required and triggers the no-recommendation logic.
    \item \textbf{Three-stage review.} A three-stage review mechanism—comprising rule-based checks, agent evaluations, and expert reviews—is adopted to filter and validate generated data, ensuring reliability and accuracy.
\end{enumerate}

\subsubsection{Three-Stage Review}

To ensure data quality, we implement a comprehensive quality control process spanning three stages: rule-based filtering, agent evaluation, and expert review. 

\begin{enumerate}
    \item \textbf{Rule-based filtering.} Automatically removes entries that fail to meet format and consistency requirements.
    \item \textbf{Agent evaluation.} We employ Gemini-2.5-Pro \citep{comanici2025gemini} to assess the internal consistency among textual information, action trajectories, and recommended actions.  Using a prompt-based evaluation framework, the model examines textual information for authenticity and naturalness, trajectories for realism and temporal coherence, and recommendations for contextual appropriateness and executability.  
    \item \textbf{Expert review.} Experts verify the remaining entries for factual accuracy, internal logic feasibility, and action feasibility. A team of 30 trained annotators, each with prior experience in human–computer interaction and data annotation, conducts the verification process. All annotators undergo standardized training and trial labeling sessions to align annotation criteria and resolve ambiguities. To ensure data quality, each data point is independently annotated by three annotators. An item is considered valid and accepted for the final dataset only if at least two annotators are in agreement on its label. This extensive cleaning and correction process represents a four-month effort with a total investment of \$210,000. Throughout this period, experts collaboratively refine and validate the dataset to ensure its reliability and consistency.
\end{enumerate}

\subsection{Function Pool Construction}
To facilitate on-device execution and standardized evaluation, we transformed textual instructions into a unified function call format by creating a predefined function pool. Our construction process involved a multi-stage pipeline. First, we manually categorized instructions into 14 scenarios. Then, we employed LLMs to initially generate function sequences and subsequently refine them by merging similar functions and parameters while pruning infrequent ones. Following this automated phase, we defined a formal schema for each function, annotating parameter data types and specifying required arguments. Finally, the entire function pool underwent a rigorous manual verification by five experienced doctoral researchers specializing in AI agents and system design, who cross-checked all definitions to ensure semantic consistency, correctness, and overall coherence.

\subsection{Difficulty Definition}

To systematically evaluate model performance across different levels of challenge, we establish a three-tier difficulty system. We classify each data item based on the number of correct predictions from a panel of five powerful models: Claude-Sonnet-4 \citep{claude4-anthropic-2025}, Claude-Sonnet-3.7 \citep{claude37-anthropic-2025}, GPT-4o \citep{hurst2024gpt4o}, Gemini-2.5-Pro \citep{comanici2025gemini}, and Gemini-2.5-Flash \citep{comanici2025gemini}. The difficulty level is defined as follows:
\begin{itemize}
\item \textbf{Level 1 (Easy)}: Correctly solved by 4–5 out of 5 reference models.
\item \textbf{Level 2 (Medium)}: Correctly solved by 2–3 out of 5 reference models.
\item \textbf{Level 3 (Hard)}: Correctly solved by 0–1 out of 5 reference models.
\end{itemize}
To validate this automatic classification, a group of \textbf{five experienced doctoral researchers} independently assessed a stratified sample of data items. The resulting difficulty annotations showed an inter-rater agreement of over \textbf{95\%} with our model-based difficulty levels, confirming the reliability and consistency of the proposed three-tier system.

\begin{figure}[!bt]
	\centering 
	\includegraphics[width=0.9\linewidth, angle=0]{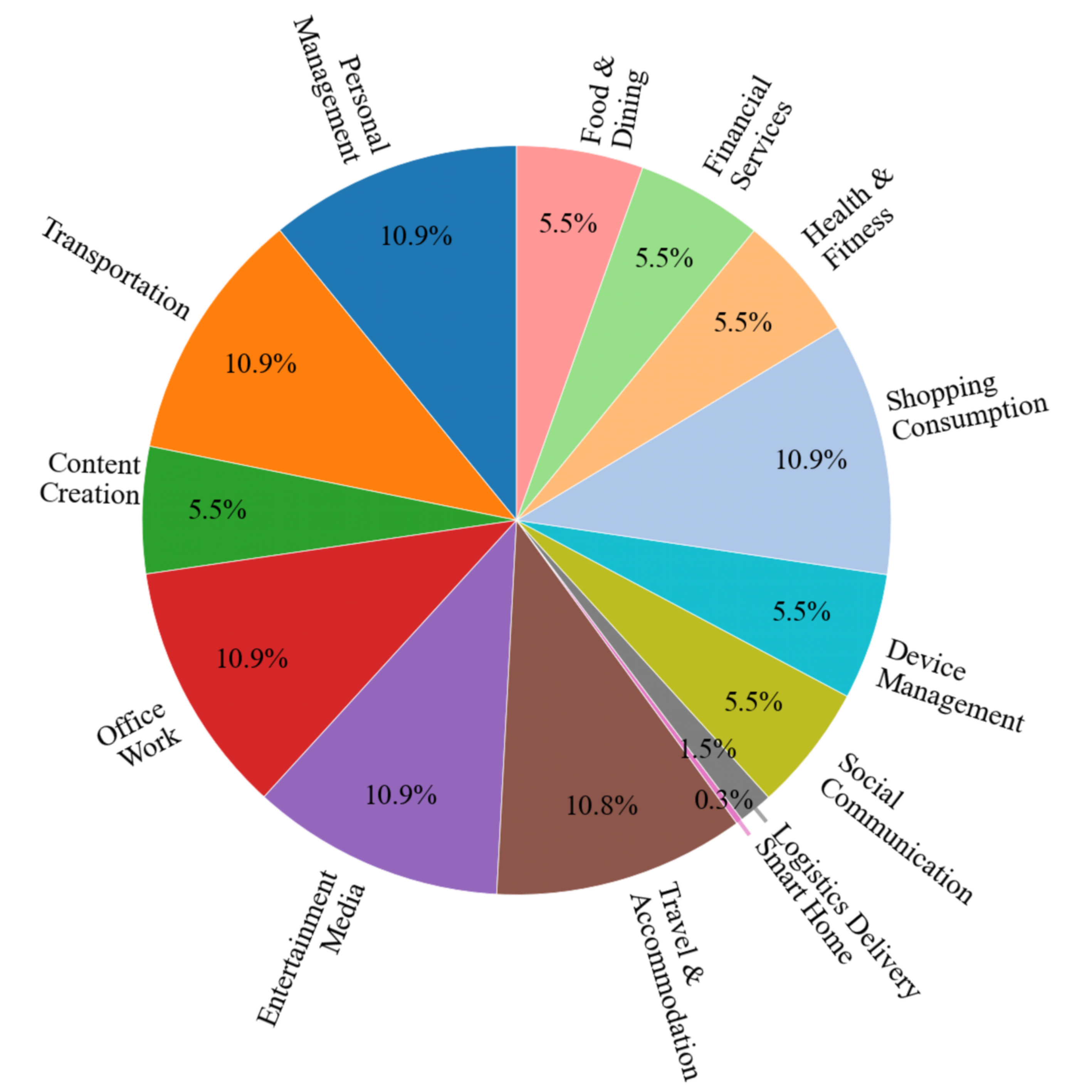}
	\vspace{-5pt}
	\caption{ Distribution of the 14 primary user intent categories, demonstrating the benchmark's broad scenario coverage (e.g., Personal Management, Office Work, Food \& Dining). }
	\vspace{-15pt}
	\label{fig:Dataset statistics and distribution}
\end{figure}

Finally, in Figure \ref{fig:Dataset statistics and distribution} and Table \ref{tb:data_overview}, we illustrate the dataset information for our benchmark.

\begin{table*}[h!]
    \small
    \centering
    \scalebox{0.95}{
    \begin{tabular}{lccccccccccc}
        \toprule
        Split & Data Type & Scenarios & Items & Intents & Images & Ave. Image & Functions & Ave. Functions & L1 & L2 & L3\\
        \midrule
        \multirow{2}{*}{Train} & Multimodal & 12 & 4,438 & 8,977 & 32,418 & 7.30 & 9,964 & 1.11 & 308 & 1,376 & 2,754 \\
        & Text & 12 & 4,438 & 4,438 & - & -& 8,259 & 1.86 & 372 & 1,208 & 2,858 \\
        \midrule
        \multirow{2}{*}{Test} & Multimodal & 14 & 1,832 & 3,711 & 14,341 & 7.83 & 4,173 & 1.24 & 118 & 613 & 1,101 \\
        & Text & 14 & 1,828 & 2,676 & - & - & 2,266 & 0.85  & 259 & 711 & 858 \\
        \bottomrule
    \end{tabular}
    }
    \vspace{-7pt}
    \caption{Statistics of the ProactiveMobile dataset, broken down by Train and Test splits and data modality. The table details the composition of our benchmark, including the number of scenarios, items, intents, functions, and distribution of different difficulties. Notably, the test set includes two additional scenarios (14 vs. 12) not present in the training set, forming our dedicated out-of-distribution evaluation split.}
    \vspace{-5pt}
    \label{tb:data_overview}
\end{table*}

\section{Experiments}
This section evaluates our approach by benchmarking against state-of-the-art closed-source MLLMs.

\begin{table*}[t!]
\small
\centering
\scalebox{0.95}{
\begin{tabular}{l!{\vrule width 1pt}cc|cc|cc!{\vrule width 1pt}cc|cc|cc}
\hline
\multirow{3}{*}{Model}
& \multicolumn{6}{c!{\vrule width 1pt}}{\textbf{L1}} 
& \multicolumn{6}{c}{\textbf{L2}} \\
\cline{2-13}
& \multicolumn{2}{c|}{Multimodal} & \multicolumn{2}{c|}{Text} & \multicolumn{2}{c!{\vrule width 1pt}}{All}
& \multicolumn{2}{c|}{Multimodal} & \multicolumn{2}{c|}{Text} & \multicolumn{2}{c}{All} \\
& SR$^\uparrow$ & FTR$^\downarrow$ & SR$^\uparrow$ & FTR$^\downarrow$ & SR$^\uparrow$ & FTR$^\downarrow$
& SR$^\uparrow$ & FTR$^\downarrow$ & SR$^\uparrow$ & FTR$^\downarrow$ & SR$^\uparrow$ & FTR$^\downarrow$ \\
\hline

GPT-5 & 12.71 & 75.00 & 24.32 & 25.57 & 20.69 & 30.45 
& 7.50 & 69.44 & 17.42 & 30.24 & 12.81 & 37.79 \\

GPT-4o & 11.86 & 100.00 & 13.13 & 47.06 & 12.73 & 51.79 
& 5.71 & 95.60 & 10.58 & 56.29 & 8.32 & 62.87 \\

o1 & \textbf{22.03} & 64.29 & \underline{30.50} & \textbf{2.73} & \underline{27.85} & \textbf{9.68}
& 11.91 & \underline{37.76} & \underline{24.90} & \textbf{6.23} & \underline{18.88} & \underline{13.09} \\

Gemini-2.5-Pro & \underline{16.10} & 69.57 & 13.51 & 59.09 & 14.32 & 60.18 
& 12.07 & 70.53 & 6.61 & 84.74 & 9.14 & 81.90 \\

Qwen2.5-VL-7B-Instruct & 0.00 & 66.67 & 2.70 & 60.00 & 1.86 & 60.71 
& 0.49 & 68.42 & 2.53 & 71.20 & 1.59 & 70.55 \\

MiMo-VL-7B-SFT-2508 & 2.54 & 76.92 & 1.93 & 82.26 & 2.12 & 81.33 
& 1.63 & 68.29 & 1.27 & 85.39 & 1.44 & 81.29 \\

Qwen2.5-VL-7B\textbf{+Proactive} 
& 10.17 & \textbf{42.31} & \textbf{37.07} & \underline{6.82} & \textbf{28.65} & \underline{10.57}
& \textbf{17.62} & \textbf{23.78} & \textbf{32.49} & \underline{9.07} & \textbf{25.60} & \textbf{12.61} \\

MiMo-VL-7B-SFT\textbf{+Proactive} 
& 11.86 & \underline{44.44} & 18.53 & 47.85 & 16.45 & 47.42
& \underline{12.40} & 40.16 & 16.32 & 49.34 & 14.50 & 47.09 \\

\hline
\multirow{3}{*}{Model} 
& \multicolumn{6}{c!{\vrule width 1pt}}{\textbf{L3}} 
& \multicolumn{6}{c}{\textbf{Avg}} \\
\cline{2-13}
& \multicolumn{2}{c|}{Multimodal} & \multicolumn{2}{c|}{Text} & \multicolumn{2}{c!{\vrule width 1pt}}{All}
& \multicolumn{2}{c|}{Multimodal} & \multicolumn{2}{c|}{Text} & \multicolumn{2}{c}{All} \\
& SR$^\uparrow$ & FTR$^\downarrow$ & SR$^\uparrow$ & FTR$^\downarrow$ & SR$^\uparrow$ & FTR$^\downarrow$
& SR$^\uparrow$ & FTR$^\downarrow$ & SR$^\uparrow$ & FTR$^\downarrow$ & SR$^\uparrow$ & FTR$^\downarrow$ \\
\hline

GPT-5 & 7.90 & 51.56 & 9.50 & 38.26 & 8.60 & 44.81 
& 8.08 & 57.99 & 14.69 & 31.41 & 11.37 & 39.20 \\

GPT-4o & 3.09 & 94.56 & 5.77 & 54.08 & 4.25 & 74.58 
& 4.53 & 95.14 & 8.69 & 53.60 & 6.60 & 65.32 \\

o1 & 11.81 & \underline{23.30} & \underline{16.08} & \underline{9.61} & \underline{13.68} & \underline{16.64}
& 12.50 & \underline{29.45} & \underline{21.55} & \textbf{6.56} & \underline{17.02} & \underline{14.09} \\

Gemini-2.5-Pro & 9.45 & 66.52 & 8.51 & 85.71 & 9.04 & 74.94 
& 10.75 & 67.81 & 8.48 & 78.29 & 9.62 & 74.98 \\

Qwen2.5-VL-7B-Instruct & 1.09 & 76.29 & 2.21 & 54.29 & 1.58 & 67.07 
& 0.82 & 73.76 & 2.41 & 64.08 & 1.61 & 67.62 \\

MiMo-VL-7B-SFT-2508  & 1.09 & 80.44 & 1.05 & 73.49 & 1.07 & 77.14 
& 1.36 & 76.71 & 1.26 & 81.09 & 1.31 & 79.57 \\

Qwen2.5-VL-7B\textbf{+Proactive} 
& \textbf{15.08} & \textbf{22.63} & \textbf{17.37} & \textbf{8.75} & \textbf{16.08} & \textbf{16.08}
& \textbf{15.61} & \textbf{23.91} & \textbf{26.04} & \underline{8.51} & \textbf{20.82} & \textbf{13.76} \\

MiMo-VL-7B-SFT\textbf{+Proactive} 
& \underline{13.62} & 43.16 & 10.37 & 51.26 & 12.20 & 46.49
& \underline{13.10} & 42.40 & 13.84 & 49.48 & 13.47 & 46.91
\\
\hline
\end{tabular}
}
\vspace{-7pt}
\caption{Overall performance comparison of our fine-tuned model (\textbf{+Proactive}) against baselines on the ProactiveMobile test set. We report two key metrics: SR$^\uparrow$, where higher is better, and FTR$^\downarrow$, where lower is better. The comparison is broken down by task difficulty (L1-L3) and data modality. For each metric, the best result is in \textbf{bold} and the second-best is \underline{underlined}. All scores are in percentage (\%).}
\vspace{-15pt}
\label{tab:all_results}
\end{table*}

\subsection{Setting}
\label{sec:setting}

\noindent\textbf{Fine-tuned Model.}
To create a specialized proactive agent, we perform full-parameter supervised fine-tuning (SFT) on Qwen2.5-VL-7B-Instruct \citep{bai2025qwen25vl} and MiMo-VL-7B-SFT-2508 \citep{coreteam2025mimovltechnicalreport}. A core aspect of our methodology is the defined output format: the model is trained to co-generate both a natural language recommendation instruction and the corresponding executable function sequence. We utilize 8,876 instances from the training split of ProactiveMobile. Further details regarding data pre-processing, specific hyperparameters, and the hardware environment are provided in Appendix. 

\noindent\textbf{Baseline Models.}
To benchmark against the current state-of-the-art, we evaluate several leading proprietary MLLMs, including GPT-5 \citep{openai_gpt5_systemcard}, GPT-4o \citep{hurst2024gpt4o}, Gemini-2.5-Pro \citep{comanici2025gemini}, o1 \citep{jaech2024openaio1}, unfinetuned Qwen2.5-VL-7B-Instruct \citep{bai2025qwen25vl}, and unfinetuned MiMo-VL-7B-SFT-2508 \citep{coreteam2025mimovltechnicalreport}. All baseline models are evaluated in a zero-shot setting. To ensure a fair comparison, the standardized prompt instructs these models to adopt the same output format. We design a standardized prompt that provides each model with the same multi-dimensional context (user profile, device status, etc.) and the list of available functions from our API pool, instructing them to output the appropriate function call sequence.

\subsection{Metrics}

Evaluating proactive intelligence presents unique challenges, especially given the one-to-many nature of valid actions in ProactiveMobile, where a single context can map to multiple ground-truth sequences. A naive evaluation metric would either be too brittle (penalizing functionally correct but formally different predictions) or too lenient. To address this, we define two core metrics, Success Rate and False Trigger Rate, whose final values are determined by a dedicated evaluation protocol designed specifically for this one-to-many context, as detailed below.

\begin{table}[ht!]
    \small
    \centering
    \scalebox{1}{
       \begin{tabular}{l|cc}
            \hline
             Model & SR$^\uparrow$ & FTR$^\downarrow$  \\
            \hline
            GPT-5 & 14.55 & 36.84\\
            GPT-4o & 5.26 & 47.37\\
            o1  & \underline{15.63} & \textbf{13.33}\\
            Gemini-2.5-Pro & 9.38 & 45.46 \\
            Qwen2.5-VL-7B-Instruct & 4.69 & 42.86\\
            MiMo-VL-7B-SFT-2508 & 0.00 & 100.00\\
            Qwen2.5-VL-7B-Instruct\textbf{+Proactive} & \textbf{20.31} & \underline{16.67}\\
            MiMo-VL-7B-SFT-2508\textbf{+Proactive} & 9.38 & 57.90 \\
            \hline
        \end{tabular}
      } 
    \vspace{-5pt}
    \caption{Performance on the OOD test set. This set comprises 64 instances from two scenarios (Logistics Delivery and Smart Home) that are entirely absent from the training data.}
    \vspace{-15pt}
    \label{tab:ood_result}
\end{table}


\begin{table}[ht!]
    \small
    \centering
    \scalebox{0.85}{
        \begin{tabular}{l|cc|cc|cc}
            \hline
            \multirow{2}{*}{\centering Training Strategy} & \multicolumn{2}{c|}{Multimodal} & \multicolumn{2}{c|}{Text} & \multicolumn{2}{c}{All}  \\
            &  SR$^\uparrow$ & FTR$^\downarrow$  & SR$^\uparrow$ & FTR$^\downarrow$  & SR$^\uparrow$ & FTR$^\downarrow$ \\
            \hline
            Func. & \underline{12.87} & 95.07 & 5.69 & 85.19 & \underline{9.18} & 93.16 \\
            \textbf{Rec.+Func.}\textbf{(ours)} &  \textbf{15.61} & \underline{23.91} & \textbf{26.04}  & \underline{8.51}  & \textbf{20.82} &  \underline{13.76}\\
            Think+Func. &  8.81 & 95.07 & 4.04 & 85.04 &  6.36 & 93.06 \\
            Think+Rec.+Func. & 5.53 & \textbf{2.87} &  \underline{10.50}  & \textbf{1.53} & 8.02  & \textbf{2.06} \\
            \hline
        \end{tabular}
        }
    \vspace{-5pt}
    \caption{Ablation study on the impact of different output formats. We compare our primary \textbf{Text Recommendation + Function} strategy against variants that only output the \textbf{Function}, or include an additional reasoning (\textbf{Think}) step. }
    \vspace{-5pt}
    \label{tab:l5_results}
\end{table}

\subsubsection{Core Metrics}

\noindent\textbf{1. Success Rate (SR).} This is our primary, binary success metric, designed to measure perfect functional equivalence. A prediction is considered accurate not based on simple string comparison, but on whether it is semantically and functionally identical to a valid ground truth. To make this judgment, we employ a powerful LLM judge (Gemini-2.5-Pro \citep{comanici2025gemini}) \footnote{To ensure the validity of judgment, we verified the consistency between the model and human experts, and achieved a consistency of 98\%.}. An instance receives a final SR score of 1 only if the model's prediction is deemed functionally equivalent to one of the valid ground-truth answers; otherwise, it is 0. Given ProactiveMobile's one-to-many nature, the precise procedure for selecting the ``best'' ground truth to compare against is critical, and is elaborated in our Best-Match Selection Protocol (Section \ref{subsubsection:best_match}).

\noindent\textbf{2. False Trigger Rate (FTR).} This metric measures the model's reliability in non-trigger scenarios. It quantifies the rate at which the model incorrectly generates an action when the ground truth specifies that no action should be taken. Let $N_{no-action}$ be the total number of instances where the ground-truth set is empty ($G = {\emptyset}$), and $N_{ft}$ be the number of those instances where the model falsely triggers a non-empty $S_{pred}$. The FTR is calculated as:
$\text{FTR} = \frac{N_{ft}}{N_{no-action}}$

\subsubsection{Best-Match Selection Protocol}
\label{subsubsection:best_match}
The aforementioned protocol dictates how a model's prediction ($S_{pred}$) is scored against the set of ground-truth candidates ($G = {S_{label,1}, ...}$) to yield the final SR score. It is a two-stage process designed to be both rigorous and fair:

\noindent\textbf{Stage 1: Prioritize Perfect Functional Equivalence.}
We first check if the model's prediction is functionally equivalent (as judged by our LLM referee) to any of the ground-truth sequences. If one or more such "perfect matches" are found, the SR for this instance is immediately set to 1, and the protocol terminates for this instance. One of these perfect matches is randomly selected as the best match ($S_{label}^*$) for any further analysis.

\noindent\textbf{Stage 2: F1-Score Fallback for Imperfect Predictions.}
If no perfect match is found in Stage 1, the SR for this instance is definitively 0. However, for consistent and fair analysis, we still need to select a single ``closest'' ground truth. In this scenario, we identify the ground-truth candidate that maximizes the F1-score (calculated on the sets of function names) when compared with $S_{pred}$. To compute this score, we treat both the prediction and the ground truth as unordered sets of function names, thus ignoring parameters and sequence order. This allows us to calculate the harmonic mean of precision (the fraction of predicted functions that are correct) and recall (the fraction of correct functions that are predicted). This F1-maximizing sequence is then designated as the best match ($S_{label}^*$).

\noindent\textbf{Why this protocol?} This two-stage design serves a crucial purpose. It establishes perfect functional correctness as the unambiguous gold standard for success, which is directly reflected in our primary SR metric. The F1-fallback mechanism, meanwhile, ensures a robust and consistent process for handling failures, providing a fair basis for comparison and deeper analysis even when the primary success condition is not met.

\subsection{Overall Performance}

Table \ref{tab:all_results} presents a comprehensive performance analysis, revealing several critical insights into the current landscape of proactive intelligence.

\noindent\textbf{Fine-tuning on ProactiveMobile consistently unlocks SOTA capabilities.}
The most striking result is the significant impact of fine-tuning on our benchmark. This effect is consistent across different base models: fine-tuning boosts the Qwen2.5-VL-7B-Instruct from 1.61\% to a state-of-the-art 20.82\% Success Rate, and similarly elevates the MiMo-VL-7B-SFT-2508 from 1.31\% to 13.47\%. Our fine-tuned Qwen model establishes a new benchmark, significantly outperforming the top-performing proprietary model, o1 (17.02\%). This gap unequivocally demonstrates that proactivity is a specialized, learnable skill requiring domain-specific adaptation, validating ProactiveMobile as an essential training resource.

\noindent\textbf{Multimodal reasoning remains a key bottleneck.}
The performance disparity across data types reveals a core challenge. For our top-performing model (Qwen2.5-VL-7B + Proactive), the SR on Text tasks (26.04\%) is substantially higher than on Multimodal tasks (15.61\%). Additionally, we find that for some multimodal tasks, certain scenarios exhibit better performance when visual information is missing, as detailed in the Appendix. This performance delta suggests that grounding abstract intents within noisy, real-world GUI screenshots introduces significant complexity, highlighting robust visual comprehension as a critical area for future advancement in on-device proactive intelligence.

\noindent\textbf{The low absolute scores validate the task's inherent difficulty.}
Despite the strong relative performance of our fine-tuned model, the absolute SR scores remain modest across the board. The fact that the state-of-the-art sits just around 21\% confirms that reliable, functionally correct proactive intelligence is a highly challenging and largely unsolved problem. This finding validates ProactiveMobile not as a benchmark for a saturated task, but as a challenging and indispensable testbed designed to catalyze genuine breakthroughs in the field.

\subsection{Generalization to Out-of-Distribution Scenarios}

To assess generalization, we evaluate all models on an out-of-distribution (OOD) test set comprising 64 instances from two scenarios—Logistics Delivery and Smart Home—that are entirely absent from the training data. The results in Table \ref{tab:ood_result} reveal a clear performance gap. Our fine-tuned Qwen2.5-VL-7B-Instruct + Proactive model achieves the best performance with 20.31\% SR, significantly outperforming other powerful generalists like o1 (15.63\%), GPT-5 (14.55\%), Gemini-2.5-Pro (9.38\%), and GPT-4o (5.26\%). This demonstrates that while immense scale offers one path to generalization, our fine-tuning approach effectively imparts a more robust and transferable understanding of proactive logic. It validates that the skills learned on ProactiveMobile are not mere pattern matching, but represent a promising step toward truly generalizable proactive intelligence.


\subsection{Ablation Study}
To validate our training and output format (Recommendation + Function), we conduct an ablation study comparing it with variants that either omit the recommendation or add an explicit Think step. The results in Table \ref{tab:l5_results} are decisive. Our chosen strategy achieves the highest SR (20.82\%), indicating that compelling the model to articulate a user-facing intent acts as an effective reasoning scaffold. Critically, formats trained without this textual recommendation (Function only and Think + Function) exhibit extremely poor safety behavior, with False Trigger Rate (FTR) rates near 100\%. This demonstrates that generating the intent is indispensable for teaching the model the crucial skill of when not to act.

The study also reveals a crucial trade-off between SR and safety. While adding a Think step (Think + Recommendation + Function) significantly lowers SR, it drastically enhances safety, slashing the FTR rate to a 2.06\%. This highlights the Think step as a promising direction for building maximally safe agents. Nevertheless, our primary Recommendation + Function approach offers the best-performing balance between SR and reliability, thus validating our core design choice. Further ablation studies, including an analysis of the impact of different contextual dimensions, are detailed in the Appendix.

\section{Conclusion}

In this work, we address the critical bottleneck hindering the transition of mobile agents from a reactive to a proactive paradigm: the lack of an executable, objective, and realistic benchmark. We introduce ProactiveMobile, a comprehensive benchmark that formalizes the proactive task around a four-dimensional context model, incorporates multi-answer annotations, and uniquely mandates an executable function-call sequence output. Our extensive experiments validate that proactivity is a specialized, learnable capability. This is consistently demonstrated as fine-tuning on our benchmark boosts the performance of different models, with our top-performing model achieving a 20.82\% success rate—establishing a new state-of-the-art that surpasses even leading proprietary models like o1 (17.02\%). This demonstrates the efficacy of ProactiveMobile as an essential tool for targeted training and highlights the significant gap in current models' out-of-the-box abilities.

While our work establishes a new SOTA, the modest absolute success rates underscore that proactive intelligence is a profoundly challenging research problem, opening up several promising future directions. Key priorities include enhancing models' multimodal reasoning to close the significant performance gap between text and multimodal tasks, and exploring advanced training methodologies like reinforcement learning for more robust decision-making. Furthermore, our ablation study on output formats reveals a rich trade-off between success rate and safety, warranting deeper investigation into creating agents that are not only effective but also trustworthy. By providing a foundational and challenging testbed, ProactiveMobile aims to catalyze these future innovations, steering the community toward the development of truly intelligent, anticipatory agents.
{
    \small
    \bibliographystyle{ieeenat_fullname}
    \bibliography{main}

@inproceedings{AITZ,
  title={Android in the Zoo: Chain-of-Action-Thought for GUI Agents},
  author={Zhang, Jiwen and Wu, Jihao and Yihua, Teng and Liao, Minghui and Xu, Nuo and Xiao, Xiao and Wei, Zhongyu and Tang, Duyu},
  booktitle={Findings of the Association for Computational Linguistics: EMNLP 2024},
  pages={12016--12031},
  year={2024}
}

@article{gui_odyssey,
  title={Gui odyssey: A comprehensive dataset for cross-app gui navigation on mobile devices},
  author={Lu, Quanfeng and Shao, Wenqi and Liu, Zitao and Meng, Fanqing and Li, Boxuan and Chen, Botong and Huang, Siyuan and Zhang, Kaipeng and Qiao, Yu and Luo, Ping},
  journal={arXiv preprint arXiv:2406.08451},
  year={2024}
}

@article{CAGUI,
      title={Agent{CPM}-{GUI}: Building Mobile-Use Agents with Reinforcement Fine-Tuning}, 
      author={Zhong Zhang and Yaxi Lu and Yikun Fu and Yupeng Huo and Shenzhi Yang and Yesai Wu and Han Si and Xin Cong and Haotian Chen and Yankai Lin and Jie Xie and Wei Zhou and Wang Xu and Yuanheng Zhang and Zhou Su and Zhongwu Zhai and Xiaoming Liu and Yudong Mei and Jianming Xu and Hongyan Tian and Chongyi Wang and Chi Chen and Yuan Yao and Zhiyuan Liu and Maosong Sun},
      year={2025},
      journal={arXiv preprint arXiv:2506.01391},
}

@misc{tang2025surveymllmbasedguiagents,
      title={A Survey on (M)LLM-Based GUI Agents}, 
      author={Fei Tang and Haolei Xu and Hang Zhang and Siqi Chen and Xingyu Wu and Yongliang Shen and Wenqi Zhang and Guiyang Hou and Zeqi Tan and Yuchen Yan and Kaitao Song and Jian Shao and Weiming Lu and Jun Xiao and Yueting Zhuang},
      year={2025},
      eprint={2504.13865},
      archivePrefix={arXiv},
      primaryClass={cs.HC},
      url={https://arxiv.org/abs/2504.13865}, 
}

@misc{zhang2025largelanguagemodelbrainedgui,
      title={Large Language Model-Brained GUI Agents: A Survey}, 
      author={Chaoyun Zhang and Shilin He and Jiaxu Qian and Bowen Li and Liqun Li and Si Qin and Yu Kang and Minghua Ma and Guyue Liu and Qingwei Lin and Saravan Rajmohan and Dongmei Zhang and Qi Zhang},
      year={2025},
      eprint={2411.18279},
      archivePrefix={arXiv},
      primaryClass={cs.AI},
      url={https://arxiv.org/abs/2411.18279}, 
}

@inproceedings{wang2025mp,
  title={Mp-gui: Modality perception with mllms for gui understanding},
  author={Wang, Ziwei and Chen, Weizhi and Yang, Leyang and Zhou, Sheng and Zhao, Shengchu and Zhan, Hanbei and Jin, Jiongchao and Li, Liangcheng and Shao, Zirui and Bu, Jiajun},
  booktitle={Proceedings of the Computer Vision and Pattern Recognition Conference},
  pages={29711--29721},
  year={2025}
}

@article{zhou2025gui,
  title={Gui-g1: Understanding r1-zero-like training for visual grounding in gui agents},
  author={Zhou, Yuqi and Dai, Sunhao and Wang, Shuai and Zhou, Kaiwen and Jia, Qinglin and Xu, Jun},
  journal={arXiv preprint arXiv:2505.15810},
  year={2025}
}

@misc{gou2025navigatingdigitalworldhumans,
      title={Navigating the Digital World as Humans Do: Universal Visual Grounding for GUI Agents}, 
      author={Boyu Gou and Ruohan Wang and Boyuan Zheng and Yanan Xie and Cheng Chang and Yiheng Shu and Huan Sun and Yu Su},
      year={2025},
      eprint={2410.05243},
      archivePrefix={arXiv},
      primaryClass={cs.AI},
      url={https://arxiv.org/abs/2410.05243}, 
}

@article{yang2024aria,
  title={Aria-ui: Visual grounding for gui instructions},
  author={Yang, Yuhao and Wang, Yue and Li, Dongxu and Luo, Ziyang and Chen, Bei and Huang, Chao and Li, Junnan},
  journal={arXiv preprint arXiv:2412.16256},
  year={2024}
}

@misc{tang2025guig2gaussianrewardmodeling,
      title={GUI-G$^2$: Gaussian Reward Modeling for GUI Grounding}, 
      author={Fei Tang and Zhangxuan Gu and Zhengxi Lu and Xuyang Liu and Shuheng Shen and Changhua Meng and Wen Wang and Wenqi Zhang and Yongliang Shen and Weiming Lu and Jun Xiao and Yueting Zhuang},
      year={2025},
      eprint={2507.15846},
      archivePrefix={arXiv},
      primaryClass={cs.LG},
      url={https://arxiv.org/abs/2507.15846}, 
}

@misc{chen2025v2pbackgroundsuppressioncenter,
      title={V2P: From Background Suppression to Center Peaking for Robust GUI Grounding Task}, 
      author={Jikai Chen and Long Chen and Dong Wang and Leilei Gan and Chenyi Zhuang and Jinjie Gu},
      year={2025},
      eprint={2508.13634},
      archivePrefix={arXiv},
      primaryClass={cs.AI},
      url={https://arxiv.org/abs/2508.13634}, 
}

@article{li2025screenspot,
  title={Screenspot-pro: Gui grounding for professional high-resolution computer use},
  author={Li, Kaixin and Meng, Ziyang and Lin, Hongzhan and Luo, Ziyang and Tian, Yuchen and Ma, Jing and Huang, Zhiyong and Chua, Tat-Seng},
  journal={arXiv preprint arXiv:2504.07981},
  year={2025}
}

@misc{wu2024osatlasfoundationactionmodel,
      title={OS-ATLAS: A Foundation Action Model for Generalist GUI Agents}, 
      author={Zhiyong Wu and Zhenyu Wu and Fangzhi Xu and Yian Wang and Qiushi Sun and Chengyou Jia and Kanzhi Cheng and Zichen Ding and Liheng Chen and Paul Pu Liang and Yu Qiao},
      year={2024},
      eprint={2410.23218},
      archivePrefix={arXiv},
      primaryClass={cs.CL},
      url={https://arxiv.org/abs/2410.23218}, 
}

@article{li2025autogui,
  title={Autogui: Scaling gui grounding with automatic functionality annotations from llms},
  author={Li, Hongxin and Chen, Jingfan and Su, Jingran and Chen, Yuntao and Li, Qing and Zhang, Zhaoxiang},
  journal={arXiv preprint arXiv:2502.01977},
  year={2025}
}

@inproceedings{hui2025winspot,
  title={WinSpot: GUI grounding benchmark with multimodal large language models},
  author={Hui, Zheng and Li, Yinheng and Zhao, Dan and Banbury, Colby and Chen, Tianyi and Koishida, Kazuhito},
  booktitle={Proceedings of the 63rd Annual Meeting of the Association for Computational Linguistics (Volume 2: Short Papers)},
  pages={1086--1096},
  year={2025}
}

@misc{liu2025uie2isynthadvancingguigrounding,
      title={UI-E2I-Synth: Advancing GUI Grounding with Large-Scale Instruction Synthesis}, 
      author={Xinyi Liu and Xiaoyi Zhang and Ziyun Zhang and Yan Lu},
      year={2025},
      eprint={2504.11257},
      archivePrefix={arXiv},
      primaryClass={cs.HC},
      url={https://arxiv.org/abs/2504.11257}, 
}

@inproceedings{zhang2024dynamic,
  title={Dynamic Planning for LLM-based Graphical User Interface Automation},
  author={Zhang, Shaoqing and Zhang, Zhuosheng and Chen, Kehai and Ma, Xinbei and Yang, Muyun and Zhao, Tiejun and Zhang, Min},
  booktitle={Findings of the Association for Computational Linguistics: EMNLP 2024},
  pages={1304--1320},
  year={2024}
}

@misc{xie2025mirage1augmentingupdatinggui,
      title={Mirage-1: Augmenting and Updating GUI Agent with Hierarchical Multimodal Skills}, 
      author={Yuquan Xie and Zaijing Li and Rui Shao and Gongwei Chen and Kaiwen Zhou and Yinchuan Li and Dongmei Jiang and Liqiang Nie},
      year={2025},
      eprint={2506.10387},
      archivePrefix={arXiv},
      primaryClass={cs.AI},
      url={https://arxiv.org/abs/2506.10387}, 
}

@misc{luo2025guir1generalistr1style,
      title={GUI-R1 : A Generalist R1-Style Vision-Language Action Model For GUI Agents}, 
      author={Run Luo and Lu Wang and Wanwei He and Longze Chen and Jiaming Li and Xiaobo Xia},
      year={2025},
      eprint={2504.10458},
      archivePrefix={arXiv},
      primaryClass={cs.CV},
      url={https://arxiv.org/abs/2504.10458}, 
}

@article{lu2025ui,
  title={UI-R1: Enhancing Efficient Action Prediction of GUI Agents by Reinforcement Learning},
  author={Lu, Zhengxi and Chai, Yuxiang and Guo, Yaxuan and Yin, Xi and Liu, Liang and Wang, Hao and Xiao, Han and Ren, Shuai and Xiong, Guanjing and Li, Hongsheng},
  journal={arXiv preprint arXiv:2503.21620},
  year={2025}
}

@misc{gu2025uivenustechnicalreportbuilding,
      title={UI-Venus Technical Report: Building High-performance UI Agents with RFT}, 
      author={Zhangxuan Gu and Zhengwen Zeng and Zhenyu Xu and Xingran Zhou and Shuheng Shen and Yunfei Liu and Beitong Zhou and Changhua Meng and Tianyu Xia and Weizhi Chen and Yue Wen and Jingya Dou and Fei Tang and Jinzhen Lin and Yulin Liu and Zhenlin Guo and Yichen Gong and Heng Jia and Changlong Gao and Yuan Guo and Yong Deng and Zhenyu Guo and Liang Chen and Weiqiang Wang},
      year={2025},
      eprint={2508.10833},
      archivePrefix={arXiv},
      primaryClass={cs.CV},
      url={https://arxiv.org/abs/2508.10833}, 
}

@article{zhao2025worldgui,
  title={WorldGUI: An Interactive Benchmark for Desktop GUI Automation from Any Starting Point},
  author={Zhao, Henry Hengyuan and Yang, Kaiming and Yu, Wendi and Gao, Difei and Shou, Mike Zheng},
  journal={arXiv preprint arXiv:2502.08047},
  year={2025}
}

@misc{tabalba2024articulateprocomparativestudyproactive,
      title={ArticulatePro: A Comparative Study on a Proactive and Non-Proactive Assistant in a Climate Data Exploration Task}, 
      author={Roderick Tabalba and Christopher J. Lee and Giorgio Tran and Nurit Kirshenbaum and Jason Leigh},
      year={2024},
      eprint={2409.10797},
      archivePrefix={arXiv},
      primaryClass={cs.HC},
      url={https://arxiv.org/abs/2409.10797}, 
}

@inproceedings{liao2023proactive,
  title={Proactive conversational agents in the post-chatgpt world},
  author={Liao, Lizi and Yang, Grace Hui and Shah, Chirag},
  booktitle={Proceedings of the 46th international ACM SIGIR conference on research and development in information retrieval},
  pages={3452--3455},
  year={2023}
}

@misc{deng2023surveyproactivedialoguesystems,
      title={A Survey on Proactive Dialogue Systems: Problems, Methods, and Prospects}, 
      author={Yang Deng and Wenqiang Lei and Wai Lam and Tat-Seng Chua},
      year={2023},
      eprint={2305.02750},
      archivePrefix={arXiv},
      primaryClass={cs.CL},
      url={https://arxiv.org/abs/2305.02750}, 
}

@article{deng2025proactive,
  title={Proactive conversational ai: A comprehensive survey of advancements and opportunities},
  author={Deng, Yang and Liao, Lizi and Lei, Wenqiang and Yang, Grace Hui and Lam, Wai and Chua, Tat-Seng},
  journal={ACM Transactions on Information Systems},
  volume={43},
  number={3},
  pages={1--45},
  year={2025},
  publisher={ACM New York, NY}
}

@inproceedings{qian2024tell,
  title={Tell Me More! Towards Implicit User Intention Understanding of Language Model Driven Agents},
  author={Qian, Cheng and He, Bingxiang and Zhuang, Zhong and Deng, Jia and Qin, Yujia and Cong, Xin and Zhang, Zhong and Zhou, Jie and Lin, Yankai and Liu, Zhiyuan and others},
  booktitle={Proceedings of the 62nd Annual Meeting of the Association for Computational Linguistics (Volume 1: Long Papers)},
  pages={1088--1113},
  year={2024}
}

@inproceedings{zhang2024ask,
  title={Ask-before-Plan: Proactive Language Agents for Real-World Planning},
  author={Zhang, Xuan and Deng, Yang and Ren, Zifeng and Ng, See Kiong and Chua, Tat-Seng},
  booktitle={Findings of the Association for Computational Linguistics: EMNLP 2024},
  pages={10836--10863},
  year={2024}
}

@inproceedings{kim2024auto,
  title={Auto-Intent: Automated Intent Discovery and Self-Exploration for Large Language Model Web Agents},
  author={Kim, Jaekyeom and Kim, Dong-Ki and Logeswaran, Lajanugen and Sohn, Sungryull and Lee, Honglak},
  booktitle={Findings of the Association for Computational Linguistics: EMNLP 2024},
  pages={16531--16541},
  year={2024}
}

@article{yang2025fingertip,
  title={Fingertip 20k: A benchmark for proactive and personalized mobile llm agents},
  author={Yang, Qinglong and Li, Haoming and Zhao, Haotian and Yan, Xiaokai and Ding, Jingtao and Xu, Fengli and Li, Yong},
  journal={arXiv preprint arXiv:2507.21071},
  year={2025}
}

@inproceedings{cao2024smart,
  title={Smart help: Strategic opponent modeling for proactive and adaptive robot assistance in households},
  author={Cao, Zhihao and Wang, Zidong and Xie, Siwen and Liu, Anji and Fan, Lifeng},
  booktitle={Proceedings of the IEEE/CVF Conference on Computer Vision and Pattern Recognition},
  pages={18091--18101},
  year={2024}
}

@article{lu2024proactiveagentthu,
  title={Proactive agent: Shifting llm agents from reactive responses to active assistance},
  author={Lu, Yaxi and Yang, Shenzhi and Qian, Cheng and Chen, Guirong and Luo, Qinyu and Wu, Yesai and Wang, Huadong and Cong, Xin and Zhang, Zhong and Lin, Yankai and others},
  journal={arXiv preprint arXiv:2410.12361},
  year={2024}
}

@article{yin2024survey,
  title={A survey on multimodal large language models},
  author={Yin, Shukang and Fu, Chaoyou and Zhao, Sirui and Li, Ke and Sun, Xing and Xu, Tong and Chen, Enhong},
  journal={National Science Review},
  volume={11},
  number={12},
  pages={nwae403},
  year={2024},
  publisher={Oxford University Press}
}

@article{bai2025qwen25vl,
  title={Qwen2. 5-vl technical report},
  author={Bai, Shuai and Chen, Keqin and Liu, Xuejing and Wang, Jialin and Ge, Wenbin and Song, Sibo and Dang, Kai and Wang, Peng and Wang, Shijie and Tang, Jun and others},
  journal={arXiv preprint arXiv:2502.13923},
  year={2025}
}

@article{zhang2024mmllm,
  title={Mm-llms: Recent advances in multimodal large language models},
  author={Zhang, Duzhen and Yu, Yahan and Dong, Jiahua and Li, Chenxing and Su, Dan and Chu, Chenhui and Yu, Dong},
  journal={arXiv preprint arXiv:2401.13601},
  year={2024}
}

@misc{yao2025surveyagenticmultimodallarge,
      title={A Survey on Agentic Multimodal Large Language Models}, 
      author={Huanjin Yao and Ruifei Zhang and Jiaxing Huang and Jingyi Zhang and Yibo Wang and Bo Fang and Ruolin Zhu and Yongcheng Jing and Shunyu Liu and Guanbin Li and Dacheng Tao},
      year={2025},
      eprint={2510.10991},
      archivePrefix={arXiv},
      primaryClass={cs.CV},
      url={https://arxiv.org/abs/2510.10991}, 
}

@inproceedings{hu2025agents,
  title={Os agents: A survey on mllm-based agents for computer, phone and browser use},
  author={Hu, Xueyu and Xiong, Tao and Yi, Biao and Wei, Zishu and Xiao, Ruixuan and Chen, Yurun and Ye, Jiasheng and Tao, Meiling and Zhou, Xiangxin and Zhao, Ziyu and others},
  booktitle={Proceedings of the 63rd Annual Meeting of the Association for Computational Linguistics (Volume 1: Long Papers)},
  pages={7436--7465},
  year={2025}
}

@inproceedings{deng-etal-2024-mobile,
    title = "Mobile-Bench: An Evaluation Benchmark for {LLM}-based Mobile Agents",
    author = "Deng, Shihan  and
      Xu, Weikai  and
      Sun, Hongda  and
      Liu, Wei  and
      Tan, Tao  and
      Liujianfeng, Liujianfeng  and
      Li, Ang  and
      Luan, Jian  and
      Wang, Bin  and
      Yan, Rui  and
      Shang, Shuo",
    editor = "Ku, Lun-Wei  and
      Martins, Andre  and
      Srikumar, Vivek",
    booktitle = "Proceedings of the 62nd Annual Meeting of the Association for Computational Linguistics (Volume 1: Long Papers)",
    month = aug,
    year = "2024",
    address = "Bangkok, Thailand",
    publisher = "Association for Computational Linguistics",
    url = "https://aclanthology.org/2024.acl-long.478/",
    doi = "10.18653/v1/2024.acl-long.478",
    pages = "8813--8831"
}

@article{li2024appagent,
  title={Appagent v2: Advanced agent for flexible mobile interactions},
  author={Li, Yanda and Zhang, Chi and Jiang, Wenjia and Yang, Wanqi and Fu, Bin and Cheng, Pei and Chen, Xin and Chen, Ling and Wei, Yunchao},
  journal={arXiv preprint arXiv:2408.11824},
  year={2024}
}

@article{wang2024mobileagentv2,
  title={Mobile-agent-v2: Mobile device operation assistant with effective navigation via multi-agent collaboration},
  author={Wang, Junyang and Xu, Haiyang and Jia, Haitao and Zhang, Xi and Yan, Ming and Shen, Weizhou and Zhang, Ji and Huang, Fei and Sang, Jitao},
  journal={Advances in Neural Information Processing Systems},
  volume={37},
  pages={2686--2710},
  year={2024}
}

@article{wang2025mobileagente,
  title={Mobile-agent-e: Self-evolving mobile assistant for complex tasks},
  author={Wang, Zhenhailong and Xu, Haiyang and Wang, Junyang and Zhang, Xi and Yan, Ming and Zhang, Ji and Huang, Fei and Ji, Heng},
  journal={arXiv preprint arXiv:2501.11733},
  year={2025}
}

@inproceedings{peng2025navigating,
  title={Navigating the Unknown: A Chat-Based Collaborative Interface for Personalized Exploratory Tasks},
  author={Peng, Yingzhe and Qin, Xiaoting and Zhang, Zhiyang and Zhang, Jue and Lin, Qingwei and Yang, Xu and Zhang, Dongmei and Rajmohan, Saravan and Zhang, Qi},
  booktitle={Proceedings of the 30th International Conference on Intelligent User Interfaces},
  pages={1048--1063},
  year={2025}
}

@inproceedings{liu2025proactiveconversationalagent,
  title={Proactive conversational agents with inner thoughts},
  author={Liu, Xingyu Bruce and Fang, Shitao and Shi, Weiyan and Wu, Chien-Sheng and Igarashi, Takeo and Chen, Xiang'Anthony'},
  booktitle={Proceedings of the 2025 CHI Conference on Human Factors in Computing Systems},
  pages={1--19},
  year={2025}
}

@misc{yang2025contextagentcontextawareproactivellm,
      title={ContextAgent: Context-Aware Proactive LLM Agents with Open-World Sensory Perceptions}, 
      author={Bufang Yang and Lilin Xu and Liekang Zeng and Kaiwei Liu and Siyang Jiang and Wenrui Lu and Hongkai Chen and Xiaofan Jiang and Guoliang Xing and Zhenyu Yan},
      year={2025},
      eprint={2505.14668},
      archivePrefix={arXiv},
      primaryClass={cs.AI},
      url={https://arxiv.org/abs/2505.14668}, 
}

@article{brachten2020ability,
  title={On the ability of virtual agents to decrease cognitive load: an experimental study},
  author={Brachten, Florian and Br{\"u}nker, Felix and Frick, Nicholas RJ and Ross, Bj{\"o}rn and Stieglitz, Stefan},
  journal={Information Systems and e-Business Management},
  volume={18},
  number={2},
  pages={187--207},
  year={2020},
  publisher={Springer}
}

@misc{openai_gpt5_systemcard,
  author       = {{OpenAI}},
  title        = {{GPT-5 System Card}},
  howpublished = {Technical report, OpenAI},
  note         = {Accessed: 2025-08-10},
  year         = {2025},
  month        = aug,
  day          = 7,
}

@article{comanici2025gemini,
  title={Gemini 2.5: Pushing the frontier with advanced reasoning, multimodality, long context, and next generation agentic capabilities},
  author={Comanici, Gheorghe and Bieber, Eric and Schaekermann, Mike and Pasupat, Ice and Sachdeva, Noveen and Dhillon, Inderjit and Blistein, Marcel and Ram, Ori and Zhang, Dan and Rosen, Evan and others},
  journal={arXiv preprint arXiv:2507.06261},
  year={2025}
}

@article{jaech2024openaio1,
  title={Openai o1 system card},
  author={Jaech, Aaron and Kalai, Adam and Lerer, Adam and Richardson, Adam and El-Kishky, Ahmed and Low, Aiden and Helyar, Alec and Madry, Aleksander and Beutel, Alex and Carney, Alex and others},
  journal={arXiv preprint arXiv:2412.16720},
  year={2024}
}

@article{hurst2024gpt4o,
  title={Gpt-4o system card},
  author={Hurst, Aaron and Lerer, Adam and Goucher, Adam P and Perelman, Adam and Ramesh, Aditya and Clark, Aidan and Ostrow, AJ and Welihinda, Akila and Hayes, Alan and Radford, Alec and others},
  journal={arXiv preprint arXiv:2410.21276},
  year={2024}
}

@misc{claude4-anthropic-2025,
  author = {{Anthropic}},
  title = {{Introducing Claude 4}},
  howpublished = {\url{https://www.anthropic.com/news/claude-4}},
  year = {2025},
  note = {Accessed: 2025-11-13}
}

@misc{claude37-anthropic-2025,
  author = {{Anthropic}},
  title = {{Claude 3.7 Sonnet and Claude Code}},
  howpublished = {\url{https://www.anthropic.com/news/claude-3-7-sonnet}},
  year = {2025},
  note = {Accessed: 2025-11-13}
}

@misc{gemini20flash-google-2025,
  author = {{Google}},
  title = {{Gemini 2.0: Flash, Flash-Lite and Pro}},
  howpublished = {\url{https://developers.googleblog.com/en/gemini-2-family-expands/}},
  year = {2025},
  note = {Accessed: 2025-11-13}
}

@misc{coreteam2025mimovltechnicalreport,
      title={MiMo-VL Technical Report}, 
      author={Core Team and Zihao Yue and Zhenru Lin and Yifan Song and Weikun Wang and Shuhuai Ren and Shuhao Gu and Shicheng Li and Peidian Li and Liang Zhao and Lei Li and Kainan Bao and Hao Tian and Hailin Zhang and Gang Wang and Dawei Zhu and Cici and Chenhong He and Bowen Ye and Bowen Shen and Zihan Zhang and Zihan Jiang and Zhixian Zheng and Zhichao Song and Zhenbo Luo and Yue Yu and Yudong Wang and Yuanyuan Tian and Yu Tu and Yihan Yan and Yi Huang and Xu Wang and Xinzhe Xu and Xingchen Song and Xing Zhang and Xing Yong and Xin Zhang and Xiangwei Deng and Wenyu Yang and Wenhan Ma and Weiwei Lv and Weiji Zhuang and Wei Liu and Sirui Deng and Shuo Liu and Shimao Chen and Shihua Yu and Shaohui Liu and Shande Wang and Rui Ma and Qiantong Wang and Peng Wang and Nuo Chen and Menghang Zhu and Kangyang Zhou and Kang Zhou and Kai Fang and Jun Shi and Jinhao Dong and Jiebao Xiao and Jiaming Xu and Huaqiu Liu and Hongshen Xu and Heng Qu and Haochen Zhao and Hanglong Lv and Guoan Wang and Duo Zhang and Dong Zhang and Di Zhang and Chong Ma and Chang Liu and Can Cai and Bingquan Xia},
      year={2025},
      eprint={2506.03569},
      archivePrefix={arXiv},
      primaryClass={cs.CL},
      url={https://arxiv.org/abs/2506.03569}, 
}

@article{wu2026atlas,
  title={Atlas: Orchestrating Heterogeneous Models and Tools for Multi-Domain Complex Reasoning},
  author={Wu, Jinyang and Zhai, Guocheng and Jin, Ruihan and Yuan, Jiahao and Shen, Yuhao and Zhang, Shuai and Wen, Zhengqi and Tao, Jianhua},
  journal={arXiv preprint arXiv:2601.03872},
  year={2026}
}

@article{wu2024beyond,
  title={Beyond examples: High-level automated reasoning paradigm in in-context learning via mcts},
  author={Wu, Jinyang and Feng, Mingkuan and Zhang, Shuai and Che, Feihu and Wen, Zengqi and Liao, Chonghua and Tao, Jianhua},
  journal={arXiv preprint arXiv:2411.18478},
  year={2024}
}
}
\clearpage 
\appendix 
\section{Implementation Details}
\subsection{Training Parameter Configuration}

This section describes the detailed settings for reproducing the fine-tuning experiments.

\noindent\textbf{Data Pre-processing.}
To mitigate potential noise in the training data, we applied the following two filtering criteria:
\begin{itemize}
\item Samples with function call sequences longer than 3 are excluded.
\item The visual trajectory for each instance is truncated to a maximum of 10 frames.
\end{itemize}

\noindent\textbf{Training Hyperparameters.}
Full-parameter supervised fine-tuning is configured with the following settings:
\begin{itemize}
\item \textbf{Optimizer:} AdamW
\item \textbf{Initial Learning Rate:} 1e-5
\item \textbf{LR Schedule:} Cosine Annealing
\item \textbf{Warmup Ratio:} 0.1
\item \textbf{Batch Size:} 64
\item \textbf{Epochs:} 4
\end{itemize}

\noindent\textbf{Inference Hyperparameters.}
To ensure a fair comparison, all hyperparameters are held constant during inference across all models.
\begin{itemize}
\item \textbf{Temperature:} 1
\item \textbf{Top P:} 0.7
\end{itemize}

\subsection{Training Format}
\noindent\textbf{Hardware and Training Time.}
The training is conducted on a 4-node cluster, utilizing a total of 32 NVIDIA H20 GPUs. The entire training process for 4 epochs took approximately 8 hours to complete.

\section{Data}

\subsection{Data statistics}

\begin{figure*}[t]
    \centering
    \includegraphics[width=1\textwidth, angle=0]{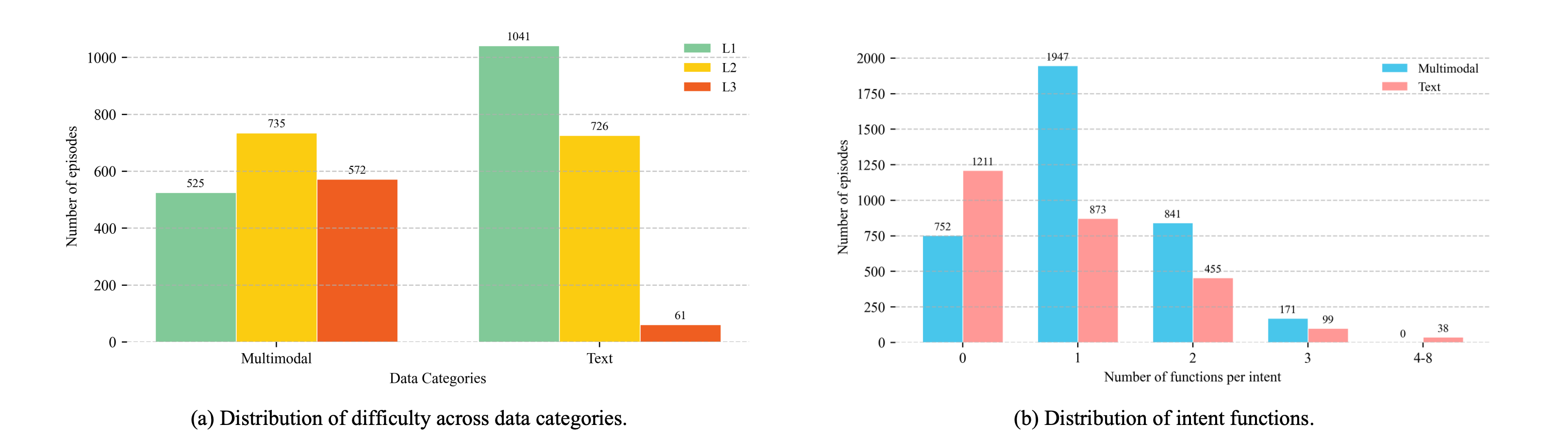}
    \vspace{-10pt}
    \caption{Detailed dataset statistics and distribution.}
    \vspace{-6pt}
    \label{fig:appendix_dataset_statistics}    
\end{figure*}

We present a more intuitive visualization of the data distribution of the proposed benchmark in Figure~\ref{fig:appendix_dataset_statistics}.

\subsection{Function}

We have constructed a comprehensive function pool comprising 63 APIs, which will be fully open-sourced. To better illustrate its structure, Figure~\ref{fig:function_sample} provides an example of a commonly used function.

\subsection{Data Annotation Platform}
\begin{figure*}[t!]
    \centering
    \includegraphics[width=0.8\textwidth, angle=0]{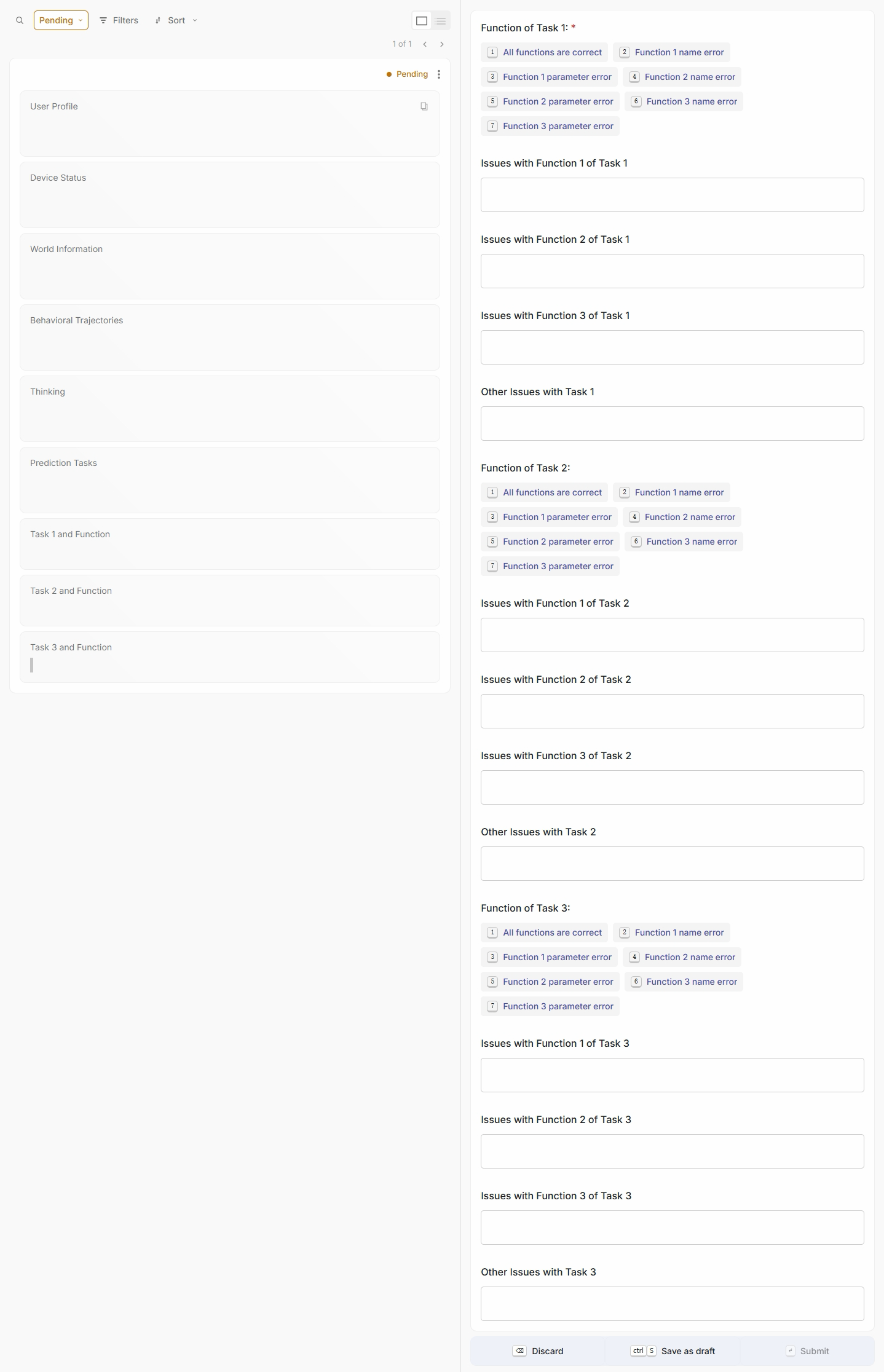}
    \vspace{0pt}
    \caption{The data annotation platform.}
    \vspace{-6pt}
    \label{fig:argilla}    
\end{figure*}

All data annotations are performed by domain experts on the Argilla platform. A screenshot of the Argilla platform is presented in Figure~\ref{fig:argilla}.

\begin{figure*}[t!]
\begin{CJK}{UTF8}{gbsn}
\centering
\scalebox{0.9}{
\begin{minipage}[t]{\linewidth}
\begin{lstlisting}[
language={},
basicstyle=\ttfamily\tiny,
breaklines=true,
frame=single,
framerule=0.4pt,
rulecolor=\color{gray!30},
xleftmargin=0.5em,
xrightmargin=0.5em,
aboveskip=0.2em,
belowskip=0.2em,
linewidth=\linewidth,
literate=
{‘}{{'}}1
{’}{{'}}1
{“}{{"}}1
{”}{{"}}1
]
"book_transport": {
    "name": "book_transport",
    "description": "A unified transportation reservation function that facilitates booking of multimodal transport services, including but not limited to aviation, railway, taxi, and ride-hailing services. Users may specify origin, destination, travel temporal parameters, and passenger demographics, with optional platform selection.",
    "similar": [
    "search_transport",
    "plan_route_and_navigate"
    ],
    "parameters": {
    "transport_type": {
        "description": "Mandatory parameter specifying the modality of transportation. Critical for distinguishing between long-haul aviation, intercity rail, urban taxi services, and short-distance mobility options. Valid enumerations include: 'flight', 'train', 'taxi' .",
        "type": "string",
        "must_fill": "required",
        "value": [
        "flight",
        "train",
        "high_speed_rail",
        "taxi",
        "ride_sharing",
        "bus",
        "subway",
        "bike",
        "rental_car",
        "ferry"
        ]
    },
    "start_location": {
        "description": "Mandatory parameter defining the geographical origin of the itinerary. Accepts structured addresses, municipal designations, IATA airport codes (e.g., 'PEK'), railway station nomenclature (e.g., 'Beijing South Railway Station'), or prominent landmarks. For example: ‘No. 1 Zhongguancun Avenue, Haidian District, Beijing’ or ‘Shanghai Hongqiao International Airport’.",
        "type": "string",
        "must_fill": "required",
        "value": "non-enumerable"
    },
    "end_location": {
        "description": "Mandatory parameter indicating the geographical terminus of the journey. Format specifications mirror those of the origin parameter, accommodating precise addresses, urban areas, transportation hubs, or notable landmarks. For example: ‘Shenzhen Nanshan Science and Technology Park’ or ‘Guangzhou Baiyun Airport’.",
        "type": "string",
        "must_fill": "required",
        "value": "non-enumerable"
    },
    "departure_time": {
        "description": "Optional temporal parameter for journey scheduling. Accepts standardized datetime formatting ('YYYY-MM-DD HH:MM:SS') or relative temporal expressions (e.g., '30 minutes hence', '09:00 tomorrow morning').",
        "type": "string",
        "must_fill": "optional",
        "value": "non-enumerable"
    },
    "passenger_num": {
        "description": "Optional parameter quantifying traveler count. Defaults to unitary occupancy if unspecified.",
        "type": "int",
        "must_fill": "optional",
        "value": "non-enumerable"
    },
    "platform_info": {
        "description": "Optional parameter designating the reservation platform or application. When specified, directs the booking to the designated service; otherwise, initiates cross-platform comparative analysis or employs default service providers. Supports mainstream mobility service platforms.For example: ‘DiDi’, ‘Ctrip’, ‘12306’.",
        "type": "string",
        "must_fill": "optional",
        "value": [
        "Didi",
        "Uber",
        "Amap",
        "CaoCao",
        "T3",
        "Ctrip",
        "Qunar",
        "Fliggy",
        "Tongcheng",
        "12306",
        "TravelSky"
        ]
    }
    }
}
\end{lstlisting}
\end{minipage}
}
\caption{Exemplary implementation of the book\_transport API.}
\label{fig:function_sample}
\end{CJK}
\end{figure*}

\section{Expanded Results with Granular Metrics}
\label{app:more_metrics}

This section provides a more granular analysis of model performance, moving beyond the primary, all-or-nothing ACC metric discussed in the main paper. Here, we report on set-based metrics that capture partial correctness and offer deeper insights into model behaviors. The results are presented in Table \ref{tab:more_metric_results}.

\subsection{Granular Metric Definitions}
The following metrics are used to generate the results in Table \ref{tab:more_metric_results}. Let $S_{pred}$ be the predicted function call sequence and $S_{label}$ be the best-match ground-truth sequence determined by our protocol.

\noindent\textbf{Type-Acc (Function Name Sequence Accuracy).} This metric focuses solely on the perfect match of the function name sequence.
\begin{equation}
\text{Type-Acc}(S_{pred}, S_{label}) = \mathbb{I}\left( \langle c_{pred,i}.\text{name} \rangle = \langle c_{label,j}.\text{name} \rangle \right)
\end{equation}

\noindent\textbf{F1-Score, Precision (P) \& Recall (R).} To provide a more forgiving, set-based evaluation, we treat the prediction and ground truth as unordered sets of function names. Let $P_{set} = \text{set}(S_{pred})$ and $L_{set} = \text{set}(S_{label})$.

\textbf{Precision (P)} measures the proportion of predicted functions that are relevant (i.e., how many of the model's suggestions are correct).
$ \text{P} = \frac{|P_{set} \cap L_{set}|}{|P_{set}|} $

\textbf{Recall (R)} measures the proportion of ground-truth functions that were successfully predicted (i.e., how many of the required actions the model found).
$ \text{R} = \frac{|P_{set} \cap L_{set}|}{|L_{set}|} $

\textbf{F1-Score (F1)} is the harmonic mean of Precision and Recall, providing a single balanced score for partial correctness.
$ \text{F1} = 2 \times \frac{\text{P} \times \text{R}}{\text{P} + \text{R}} $
\subsection{Analysis of Detailed Results}

The detailed results in Table \ref{tab:more_metric_results} reinforce the primary conclusions from the main paper while offering additional, nuanced insights:

\noindent\textbf{1. Fine-tuning Consistently Unlocks Proactive Skills.}
The most prominent trend is the dramatic performance uplift from fine-tuning on ProactiveMobile. This holds true for both model families. The Qwen2.5-VL-7B-Instruct model's average F1 score catapults from 4.50\% to 50.88\% after being trained, becoming Qwen2.5-VL-7B-Instruct + Proactive. Similarly, the MiMo-VL-7B-SFT-2508 + Proactive model significantly outperforms its base version. This strongly corroborates our main finding that proactivity is a learnable skill, and our benchmark is an effective resource for teaching it.

\noindent\textbf{2. A Clear Hierarchy Among Models Emerges.}
The granular metrics reveal a clear performance hierarchy. The o1 model consistently establishes itself as the top-performing baseline across nearly all metrics and settings, demonstrating its powerful general-purpose reasoning. Our fine-tuned Qwen2.5-VL-7B-Instruct + Proactive model emerges as a highly competitive contender, often securing the second-best performance, particularly in Text-based scenarios where its F1 score (60.84\%) approaches that of o1 (72.09\%). Other models like GPT-5, GPT-4o, and Gemini-2.5-Pro show varied performance, but generally trail behind the top two.


\begin{table*}[!htbp]
    \small
    \centering
    \scalebox{0.8}{
        \begin{tabular}{l|l|cccc|cccc|cccc}
        \hline
        \multirow{2}{*}{\centering Difficulty} & \multirow{2}{*}{\centering Model} & \multicolumn{4}{c|}{Multimodal} & \multicolumn{4}{c|}{Text} & \multicolumn{4}{c}{All}  \\
        & & Type-Acc$^\uparrow$ & F1$^\uparrow$  & P$^\uparrow$ & R$^\uparrow$ & Type-Acc$^\uparrow$ & F1$^\uparrow$  & P$^\uparrow$ & R$^\uparrow$ & Type-Acc$^\uparrow$ & F1$^\uparrow$  & P$^\uparrow$ & R$^\uparrow$ \\
        \hline
        \multicolumn{1}{c|}{\multirow{8}{*}{L1}} & GPT-5 & 33.05 & \underline{58.50} & 56.64 & \textbf{65.25} & 67.57 & 70.91 & 71.04 & 71.36 & 56.76 & 67.03 & 66.53 & 69.45 \\
         & GPT-4o & 30.51 & 49.70 & 46.85 & 57.91 & 45.17 & 48.82 & 48.46 & 49.81 & 40.58 & 49.09 & 47.95 & 52.34 \\
         & o1 & \underline{51.70} & 58.48 & \underline{59.75} & 58.05 & \textbf{91.12} & \textbf{92.41} & \textbf{92.47} & \textbf{92.47} & \textbf{78.78} & \textbf{81.79} & \textbf{82.23} & \textbf{81.70} \\
         & Gemini-2.5-Pro & \textbf{55.08} & \textbf{63.70} & \textbf{65.25} & \underline{63.14} & 40.93 & 43.50 & 44.02 & 43.44 & 45.36 & 49.82 & 50.66 & 49.60 \\
         & Qwen2.5-VL-7B-Instruct & 3.39 & 3.39 & 3.39 & 3.39 & 7.72 & 7.72 & 7.72 & 7.72 & 6.37 & 6.37 & 6.37 & 6.37 \\
         & MiMo-VL-7B-SFT-2508 & 5.93 & 8.45 & 8.33 & 9.32 & 4.63 & 4.89 & 4.83 & 5.02 & 5.04 & 6.00 & 5.92 & 6.37 \\
         & Qwen2.5-VL-7B\textbf{+Proactive} & 32.20 & 38.84 & 40.25 & 38.56 & \underline{82.24} & \underline{83.45} & \underline{83.46} & \underline{83.59} & \underline{66.58} & \underline{69.49} & \underline{69.94} & \underline{69.50} \\
         & MiMo-VL-7B-SFT\textbf{+Proactive} & 30.51 & 41.02 & 41.95 & 41.81 & 42.47 & 44.75 & 44.72 & 45.17 & 38.73 & 43.58 & 43.86 & 44.12 \\
        \midrule
        \multicolumn{1}{c|}{\multirow{8}{*}{L2}} & GPT-5 & 26.75 & 50.61 & 48.29 & \textbf{58.05} & 54.67 & 63.50 & 63.74 & 64.95 & 41.70 & 57.51 & 56.56 & \underline{61.74} \\
         & GPT-4o & 15.33 & 38.78 & 35.89 & 48.04 & 32.58 & 38.88 & 38.65 & 40.74 & 24.58 & 38.83 & 37.37 & 44.12 \\
         & o1 & \textbf{42.58} & \underline{51.78} & \underline{52.96} & 51.99 & \textbf{76.93} & \textbf{80.96} & \textbf{82.49} & \textbf{80.24} & \textbf{61.03} & \textbf{67.45} & \textbf{68.82} & \textbf{67.16} \\
         & Gemini-2.5-Pro & \underline{42.09} & \textbf{56.17} & \textbf{57.75} & \underline{57.04} & 24.61 & 31.59 & 33.68 & 30.95 & 32.70 & 42.97 & 44.83 & 43.03 \\
         & Qwen2.5-VL-7B-Instruct & 3.43 & 4.04 & 4.05 & 4.13 & 5.20 & 5.27 & 5.25 & 5.34 & 4.38 & 4.70 & 4.70 & 4.78 \\
         & MiMo-VL-7B-SFT-2508 & 5.55 & 8.16 & 8.10 & 8.97 & 3.80 & 3.98 & 3.97 & 4.10 & 4.61 & 5.92 & 5.88 & 6.36 \\
         & Qwen2.5-VL-7B\textbf{+Proactive} & 34.75 & 43.39 & 44.13 & 44.02 & \underline{69.06} & \underline{72.34} & \underline{72.60} & \underline{72.68} & \underline{53.17} & \underline{58.93} & \underline{59.42} & 59.41 \\
         & MiMo-VL-7B-SFT\textbf{+Proactive} & 30.02 & 43.21 & 44.56 & 44.32 & 34.32 & 41.23 & 41.40 & 42.35 & 32.33 & 42.15 & 42.86 & 43.26 \\
        \midrule
        \multicolumn{1}{c|}{\multirow{8}{*}{L3}} & GPT-5 & 23.98 & \underline{44.66} & 44.26 & \textbf{49.09} & 29.19 & \underline{46.94} & \underline{48.42} & \underline{48.73} & 26.25 & \underline{45.66} & 46.07 & \underline{48.93} \\
         & GPT-4o & 12.62 & 31.84 & 30.55 & 37.56 & 22.59 & 37.06 & 36.95 & 40.38 & 16.97 & 34.11 & 33.34 & 38.79 \\
         & o1 & \textbf{40.51} & \textbf{49.53} & \textbf{51.74} & \underline{48.99} & \textbf{50.35} & \textbf{58.61} & \textbf{61.11} & \textbf{57.67} & \textbf{44.82} & \textbf{53.50} & \textbf{55.84} & \textbf{52.79} \\
         & Gemini-2.5-Pro & 31.61 & 44.48 & \underline{47.02} & 44.19 & 27.16 & 41.98 & 45.38 & 41.38 & 29.66 & 43.38 & \underline{46.30} & 42.96 \\
         & Qwen2.5-VL-7B-Instruct & 2.72 & 3.80 & 3.93 & 3.84 & 4.08 & 4.26 & 4.23 & 4.31 & 3.32 & 4.00 & 4.06 & 4.04 \\
         & MiMo-VL-7B-SFT-2508 & 3.09 & 5.01 & 5.03 & 5.69 & 3.50 & 4.91 & 5.11 & 5.13 & 3.27 & 4.97 & 5.07 & 5.44 \\
         & Qwen2.5-VL-7B\textbf{+Proactive} & \underline{33.24} & 39.79 & 40.77 & 40.21 & \underline{39.04} & 44.50 & 45.28 & 44.85 & \underline{35.78} & 41.85 & 42.74 & 42.24 \\
         & MiMo-VL-7B-SFT\textbf{+Proactive} & 28.70 & 39.42 & 40.75 & 40.05 & 22.14 & 33.55 & 34.65 & 34.66 & 25.83 & 36.85 & 38.08 & 37.69 \\
        \midrule
        \multicolumn{1}{c|}{\multirow{8}{*}{Avg}} & GPT-5 & 25.49 & 47.54 & 46.40 & \textbf{53.13} & 44.55 & 56.79 & 57.59 & 58.25 & 34.99 & \underline{52.15} & \underline{51.98} & \underline{55.68} \\
         & GPT-4o & 14.68 & 35.31 & 33.38 & 42.38 & 29.70 & 39.44 & 39.25 & 41.86 & 22.16 & 37.37 & 36.30 & 42.12 \\
         & o1 & \textbf{41.92} & \textbf{50.86} & \textbf{52.67} & \underline{50.57} & \textbf{66.47} & \textbf{72.09} & \textbf{73.87} & \textbf{71.38} & \textbf{54.18} & \textbf{61.46} & \textbf{63.26} & \textbf{60.97} \\
         & Gemini-2.5-Pro & \underline{36.63} & \underline{49.63} & \underline{51.78} & 49.71 & 28.12 & 38.15 & 40.64 & 37.61 & 32.38 & 43.90 & 46.22 & 43.67 \\
         & Qwen2.5-VL-7B-Instruct & 3.00 & 3.85 & 3.94 & 3.91 & 5.03 & 5.15 & 5.12 & 5.20 & 4.02 & 4.50 & 4.53 & 4.55 \\
         & MiMo-VL-7B-SFT-2508 & 4.09 & 6.28 & 6.27 & 7.02 & 3.78 & 4.54 & 4.63 & 4.71 & 3.93 & 5.42 & 5.45 & 5.87 \\
         & Qwen2.5-VL-7B\textbf{+Proactive} & 33.68 & 40.93 & 41.86 & 41.38 & \underline{56.84} & \underline{60.84} & \underline{61.32} & \underline{61.16} & \underline{45.25} & 50.88 & 51.58 & 51.26 \\
         & MiMo-VL-7B-SFT\textbf{+Proactive} & 29.26 & 40.79 & 42.10 & 41.59 & 29.76 & 38.13 & 38.70 & 39.14 & 29.51 & 39.46 & 40.40 & 40.37 \\
        \hline
        \end{tabular}
        }
        
    \caption{\textbf{Detailed performance comparison using granular, set-based metrics.} We report on function name sequence accuracy (Type-Acc$^\uparrow$), F1-Score$^\uparrow$, Precision (P$^\uparrow$), and Recall (R$^\uparrow$) across different difficulties and modalities. Best results are in \textbf{bold}, and second-best are \underline{underlined}. All scores are in percentage (\%). Qwen2.5-VL-7B\textbf{+Proactive} represents Qwen2.5-VL-7B-Instruct + Proactive, and MiMo-VL-7B-SFT\textbf{+Proactive} represents MiMo-VL-7B-SFT-2508 + Proactive.}
    \label{tab:more_metric_results}
\end{table*}

\section{Ablation Study on Four-Dimension Information}

This section details an ablation study designed to quantify the contribution of each of the four contextual dimensions proposed in our benchmark: User Profile, Device Status, World Information, and Behavioral Trajectories.

\subsection{Experimental Setup}
To isolate the impact of each dimension, we systematically removed one dimension at a time from the input provided to the models and re-ran the evaluation. The ``All Info" condition, where models have access to the complete four-dimensional context, serves as the baseline for comparison. The results of this study are presented in Table \ref{tab:ablation_information}. It is worth noting that in the w/o Trajectories condition, the distinction between Multimodal and Text data disappears, as the visual trajectory is the sole differentiating element.

\subsection{Analysis of Results}

\textbf{The ProactiveMobile Benchmark demonstrates strong robustness.} Our experiments indicate that omitting any single input dimension leads to only minor fluctuations in performance. For instance, the SR of our fine-tuned Qwen2.5-VL-7B-Instruct + Proactive model varies by merely 1.4\%, while its FTR fluctuates by around 4\%. For the o1 model, the corresponding variations are approximately 1.5\% and 6.3\%, respectively.

These findings are consistent with our design motivation. In real-world environments, contextual signals often overlap or contain redundant information. Incorporating such redundancy during training enables models to learn to extract the most relevant signals from complex inputs. As a result, the absence of any single information source does not substantially degrade performance and, in some cases, can even lead to slight improvements.

\section{Case Study}
The complete task workflow is illustrated through a concrete case provided in Table~\ref{tab:case}. Additionally, we conducted a comparative evaluation of our model against strong counterparts, including GPT-5 and o1. Notably, for the purpose of clear demonstration, both case studies present only a single intent.

\section{Prompts for the LLM agents}
A suite of tailored prompts is utilized, encompassing the following categories: Prompt for Generating Contextual Information, Prompt for Generating Potential Intentions, Prompt for Adding Interfering Information, Prompt for Mapping to Function, Prompt for Three-stage Review, and Prompt for Training/Inference. The specific prompts are compiled in the prompt box below.

\begin{table*}[t!]
    \small
    \centering
    \scalebox{0.95}{
        \begin{tabular}{l|l|cc|cc|cc}
        \hline
        \multirow{2}{*}{\centering Input} & \multirow{2}{*}{\centering Model} & \multicolumn{2}{c|}{Multimodal} & \multicolumn{2}{c|}{Text} & \multicolumn{2}{c}{All}  \\
        & & SR$^\uparrow$ & FTR$^\downarrow$  & SR$^\uparrow$ & FTR$^\downarrow$  & SR$^\uparrow$ & FTR$^\downarrow$ \\
        \midrule
        \multicolumn{1}{c|}{\multirow{8}{*}{w/o Profile}}
        & GPT-5 & 8.41 & 40.30 & 14.00 & 27.76 & 11.20 & 31.87 \\
        & GPT-4o & 5.19 & 89.42 & 11.43 & 45.35 & 8.31 & 57.78 \\
        & o1 & 12.12 & \underline{17.28} & \underline{20.02} & \textbf{5.18} & \underline{16.07} & \textbf{9.43} \\
        & Gemini-2.5-Pro & 10.70 & 60.10 & 8.59 & 77.42 & 9.65 & 71.55 \\
        & Qwen2.5-VL-7B-Instruct & 1.91 & 55.86 & 2.52 & 60.16 & 2.21 & 58.60 \\
        & MiMo-VL-7B-SFT-2508 & 1.53 & 80.80 & 1.86 & 78.93 & 1.69 & 79.56 \\
        & Qwen2.5-VL-7B\textbf{+Proactive} & \textbf{14.25} & \textbf{17.00} & \textbf{28.12} & \underline{7.78} & \textbf{21.18} & \underline{11.05} \\
        & MiMo-VL-7B-SFT\textbf{+Proactive} & \underline{13.76} & 38.68 & 14.83 & 49.45 & 14.29 & 45.53 \\
        \midrule
        \multicolumn{1}{c|}{\multirow{8}{*}{w/o Device}}
        & GPT-5 & 8.19 & 51.22 & 13.73 & 29.05 & 10.96 & 35.70 \\
        & GPT-4o & 3.38 & 94.74 & 7.59 & 49.21 & 5.48 & 61.89 \\
        & o1 & 12.66 & \underline{24.02} & \underline{19.42} & \textbf{5.55} & \underline{16.04} & \textbf{11.68} \\
        & Gemini-2.5-Pro & 9.83 & 67.77 & 7.82 & 78.05 & 8.83 & 74.81 \\
        & Qwen2.5-VL-7B-Instruct & 1.31 & 74.82 & 1.86 & 66.06 & 1.59 & 69.38 \\
        & MiMo-VL-7B-SFT-2508 & 1.09 & 86.98 & 1.59 & 80.23 & 1.34 & 83.11 \\
        & Qwen2.5-VL-7B\textbf{+Proactive} & \textbf{15.61} & \textbf{20.98} & \textbf{26.37} & \underline{10.18} & \textbf{20.98} & \underline{13.96} \\
        & MiMo-VL-7B-SFT\textbf{+Proactive} & \underline{12.83} & 42.21 & 14.72 & 44.84 & 13.77 & 43.94 \\
        \midrule
        \multicolumn{1}{c|}{\multirow{8}{*}{w/o World}}
        & GPT-5 & 9.44 & 42.66 & 14.83 & 27.23 & 12.13 & 32.08 \\
        & GPT-4o & 3.99 & 91.04 & 10.61 & 42.49 & 7.30 & 55.56 \\
        & o1 & \underline{11.85} & \underline{25.66} & \underline{19.26} & \textbf{4.82} & \underline{15.55} & \underline{11.72} \\
        & Gemini-2.5-Pro & 9.88 & 63.20 & 8.92 & 75.35 & 9.40 & 71.40 \\
        & Qwen2.5-VL-7B-Instruct & 0.82 & 81.65 & 2.08 & 68.31 & 1.45 & 73.57 \\
        & MiMo-VL-7B-SFT-2508 & 1.15 & 77.78 & 1.81 & 77.29 & 1.48 & 77.46 \\
        & Qwen2.5-VL-7B\textbf{+Proactive} & \textbf{15.88} & \textbf{19.97} & \textbf{26.75} & \underline{6.97} & \textbf{21.31} & \textbf{11.57} \\
        & MiMo-VL-7B-SFT\textbf{+Proactive} & 13.65 & 36.50 & 13.73 & 49.40 & 13.69 & 44.72 \\
        \midrule
        \multicolumn{1}{c|}{\multirow{8}{*}{w/o Trajectories}}
        & GPT-5 & 8.30 & 31.48 & 15.70 & 15.85 & 12.00 & 20.98 \\
        & GPT-4o & 5.84 & 50.00 & 11.00 & 27.66 & 8.42 & 34.73 \\
        & o1 & 11.35 & \textbf{20.35} & \underline{19.86} & \textbf{1.58} & \underline{15.60} & \textbf{7.83} \\
        & Gemini-2.5-Pro & 8.68 & 76.34 & 8.10 & 66.59 & 8.39 & 69.43 \\
        & Qwen2.5-VL-7B-Instruct & 1.80 & 63.76 & 3.83 & 51.85 & 2.81 & 55.83 \\
        & MiMo-VL-7B-SFT-2508 & 1.31 & 79.63 & 1.20 & 79.10 & 1.26 & 79.30 \\
        & Qwen2.5-VL-7B\textbf{+Proactive} & \textbf{15.07} & \underline{25.28} & \textbf{29.32} & \underline{2.74} & \textbf{22.19} & \underline{9.92} \\
        & MiMo-VL-7B-SFT\textbf{+Proactive} & \underline{13.97} & 30.78 & 18.38 & 32.02 & 16.18 & 36.29 \\
        \midrule
        \multicolumn{1}{c|}{\multirow{8}{*}{All Info}}
        & GPT-5 & 8.08 & 57.99 & 14.69 & 31.41 & 11.37 & 39.20 \\
        & GPT-4o & 4.53 & 95.14 & 8.69 & 53.60 & 6.60 & 65.32 \\
        & o1 & 12.50 & \underline{29.45} & \underline{21.55} & \textbf{6.56} & \underline{17.02} & \underline{14.09} \\
        & Gemini-2.5-Pro & 10.75 & 67.81 & 8.48 & 78.29 & 9.62 & 74.98 \\
        & Qwen2.5-VL-7B-Instruct & 0.82 & 73.76 & 2.41 & 64.08 & 1.61 & 67.62 \\
        & MiMo-VL-7B-SFT-2508 & 1.37 & 76.71 & 1.26 & 81.09 & 1.31 & 79.57 \\
        & Qwen2.5-VL-7B\textbf{+Proactive} & \textbf{15.61} & \textbf{23.91} & \textbf{26.04} & \underline{8.51} & \textbf{20.82} & \textbf{13.76} \\
        & MiMo-VL-7B-SFT\textbf{+Proactive} & \underline{13.10} & 42.40 & 13.84 & 49.48 & 13.47 & 46.91 \\
        \bottomrule
        \end{tabular}
        }
        
    \caption{\textbf{Ablation study on the impact of contextual information dimensions.}  We evaluate model performance after systematically removing one of the four key dimensions: User Profile, Device Status, World Information, and Behavioral Trajectories. The ``All Info" row represents the full model performance with no information removed, serving as the baseline. Metrics are Success Rate (SR$^\uparrow$) and False Trigger Rate (FTR$^\downarrow$). Best results are in \textbf{bold}, and second-best are \underline{underlined}. All scores are in percentage (\%). Note that the `w/o Trajectories` experiment removes the distinction between Multimodal and Text data.}
    \label{tab:ablation_information}
\end{table*}

\begin{table*}[t!]
\centering
\small
\setlength{\tabcolsep}{3pt}
\renewcommand{\arraystretch}{1.05}
\begin{tabularx}{\textwidth}{
    >{\bfseries\centering\arraybackslash}m{1.6cm}
    >{\hsize=1.4\hsize}X
    >{\centering\arraybackslash}m{2.6cm}
    >{\centering\arraybackslash}m{2.6cm}
    >{\centering\arraybackslash}m{2.6cm}
}
\toprule
\begin{minipage}[t][1cm][c]{1.6cm}  
    \centering                    
    \vspace{0.1em}                 
    User Profile                 
    \vspace{0.01em}               
\end{minipage}                   
&                               
\multicolumn{4}{>{\hsize=\dimexpr4\hsize+8\tabcolsep\relax\raggedright\arraybackslash}X}{
Based on the user profile and historical behavior analysis, the user is 25 years old and a photography enthusiast who frequently browses digital camera review websites. Recently, the user has repeatedly searched for the “Sony Alpha 7 IV.” The user also has a habit of purchasing products from U.S. e-commerce platforms via cross-border shopping (“Haitao”). On the phone, the user has installed international remittance apps such as Wise and has a record of using them, showing a preference for payment channels with lower transaction fees. Recently, the user has also browsed content related to travel and geography, such as searching for “the capital of China.”
} \\
\midrule
\begin{minipage}[t][1cm][c]{1.6cm}  
    \centering                    
    \vspace{0.01em}                 
    Device Status                 
    \vspace{0.5em}               
\end{minipage}                   
&                               
\multicolumn{4}{>{\hsize=\dimexpr4\hsize+8\tabcolsep\relax\raggedright\arraybackslash}X}{
The current device time is 9:45 AM on Tuesday, October 11. The system’s primary language is set to English (Australia). The device is located in Sydney, Australia. The phone battery level is sufficient at 91\%, and it is connected to a stable Wi-Fi network. Installed applications include Amazon, Wise, Gmail, Chrome, and YouTube. Recent notifications include a shopping cart reminder from Amazon: “Items in your cart are still waiting for you!”
} \\
\midrule
\begin{minipage}[t][1cm][c]{1.6cm}  
    \centering                    
    \vspace{0.01em}                 
    World Information                 
    \vspace{0.5em}               
\end{minipage}                   
&                               
\multicolumn{4}{>{\hsize=\dimexpr4\hsize+8\tabcolsep\relax\raggedright\arraybackslash}X}{
Macroeconomic information indicates that the global financial market is currently in an active trading period. According to real-time data, the current USD/AUD interbank exchange rate is approximately 1.58, although most payment channels typically charge additional fees on top of this rate. A financial news report notes that cross-border payment services provided by emerging fintech companies often offer more favorable exchange rates than traditional banks. An international news summary also reports that the European Space Agency is preparing a new Mars exploration program.
} \\
\midrule
\begin{minipage}[t][0.1cm][c]{1.6cm}  
    \centering                    
    \vspace{0.01em}                 
    Behavioral Trajectories       
    \vspace{15em}               
\end{minipage}                   
&                               
\multicolumn{3}{c}{
  \hspace*{-1cm}
  \begin{minipage}[c]{\linewidth}
    \centering
    \includegraphics[width=0.16\linewidth, height=0.2\textheight, keepaspectratio]{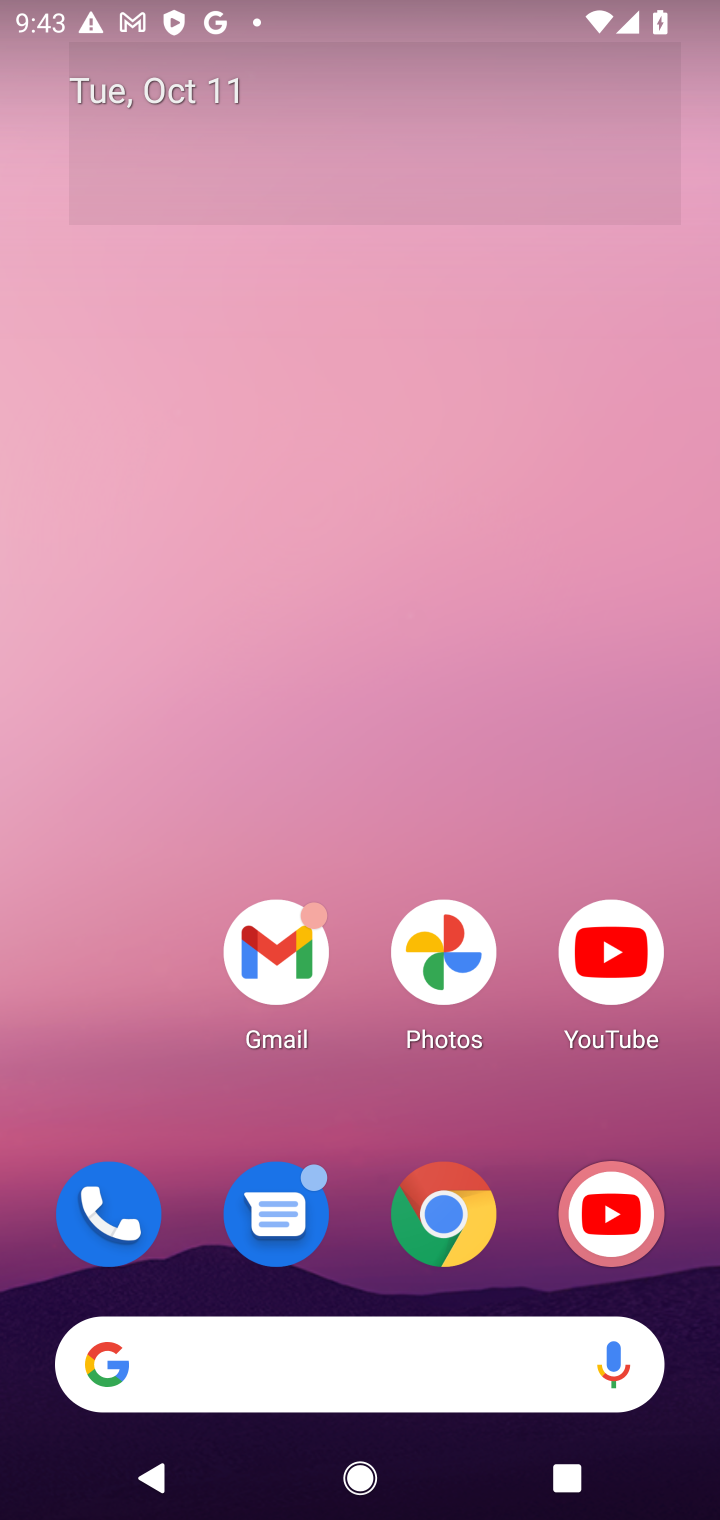}
    \includegraphics[width=0.16\linewidth, height=0.2\textheight, keepaspectratio]{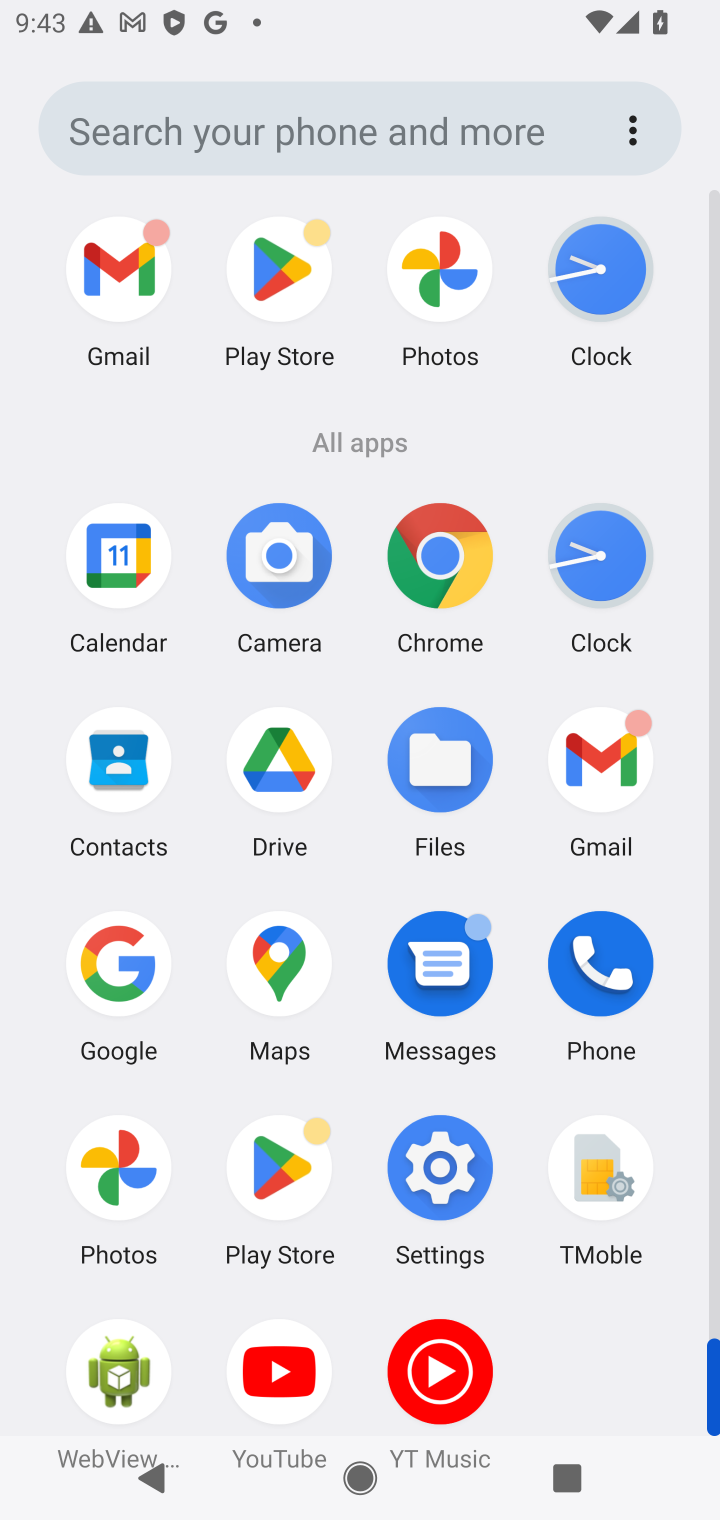}
    \includegraphics[width=0.16\linewidth, height=0.2\textheight, keepaspectratio]{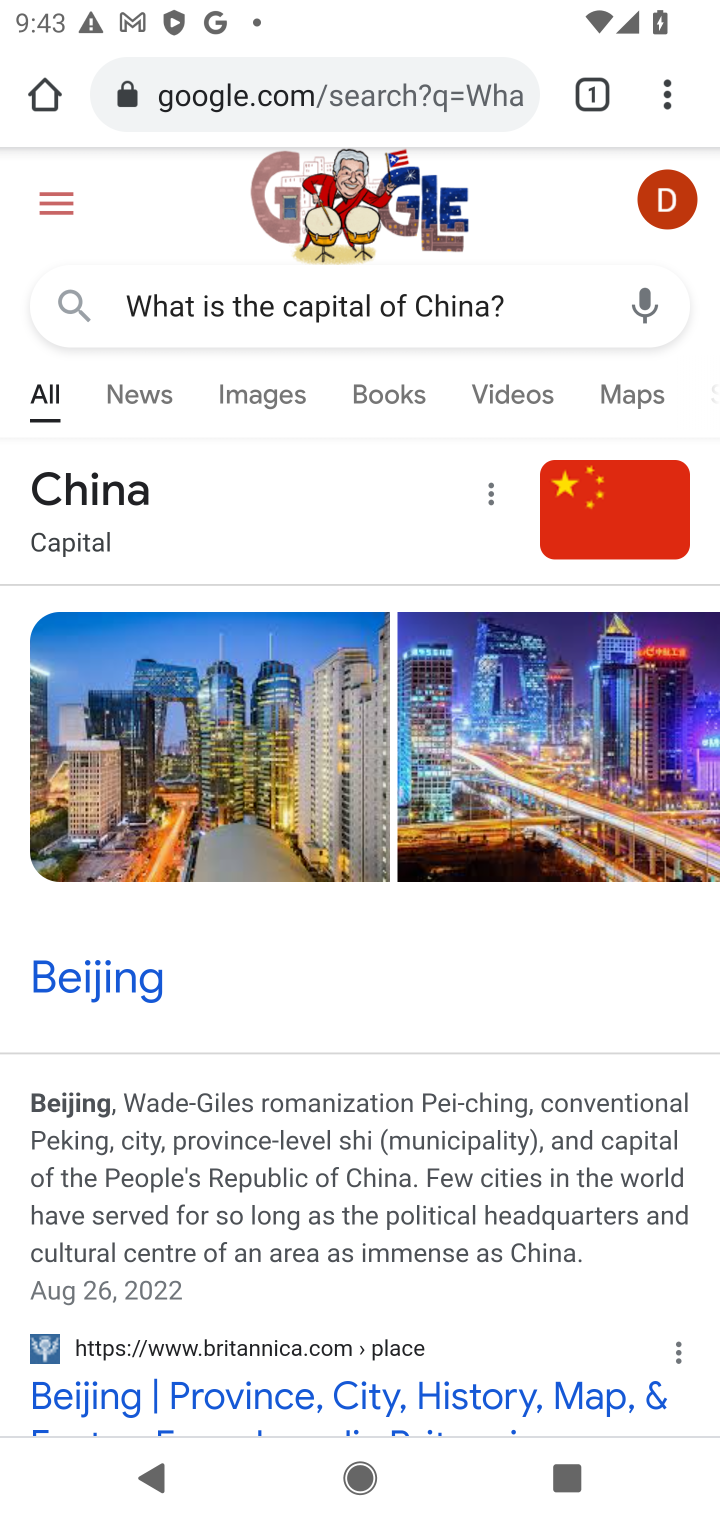}
    \includegraphics[width=0.16\linewidth, height=0.2\textheight, keepaspectratio]{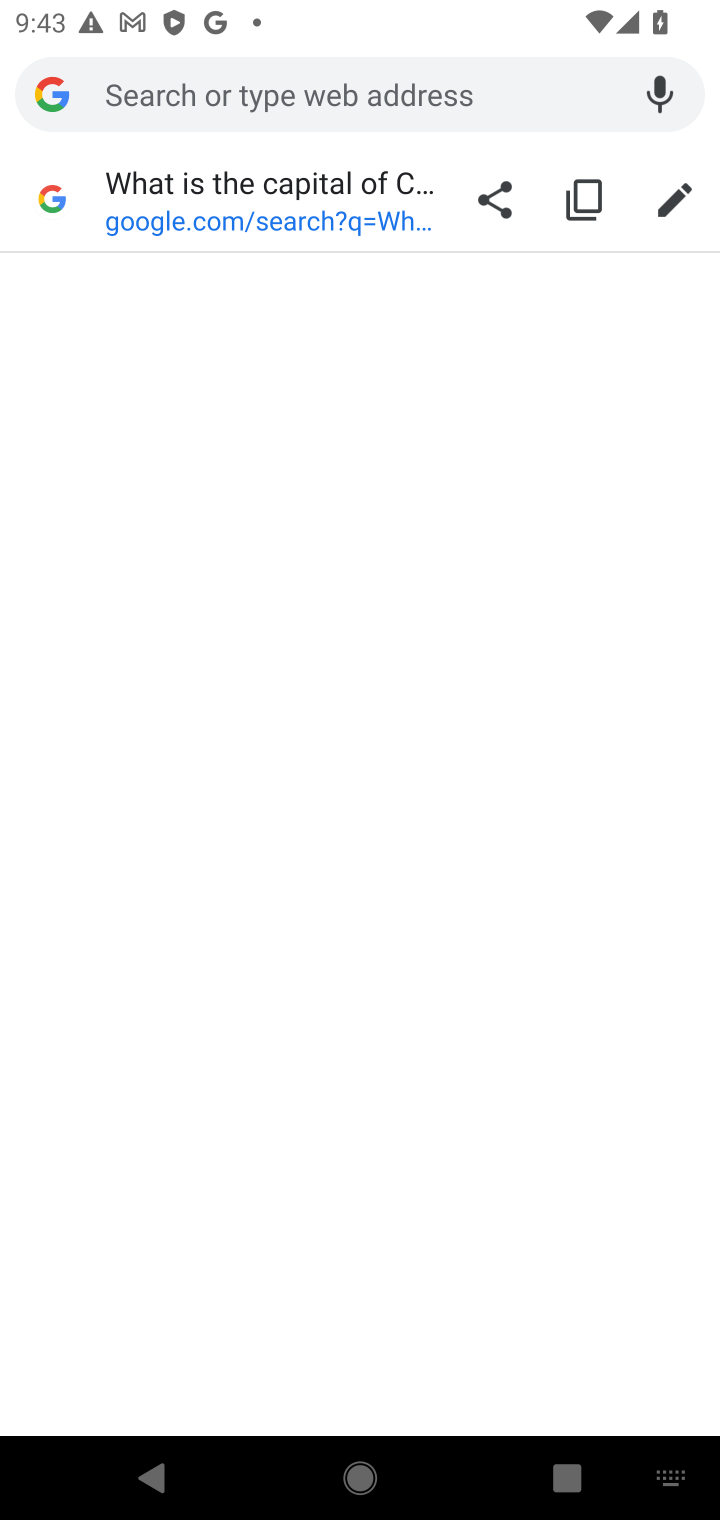}
    \includegraphics[width=0.16\linewidth, height=0.2\textheight, keepaspectratio]{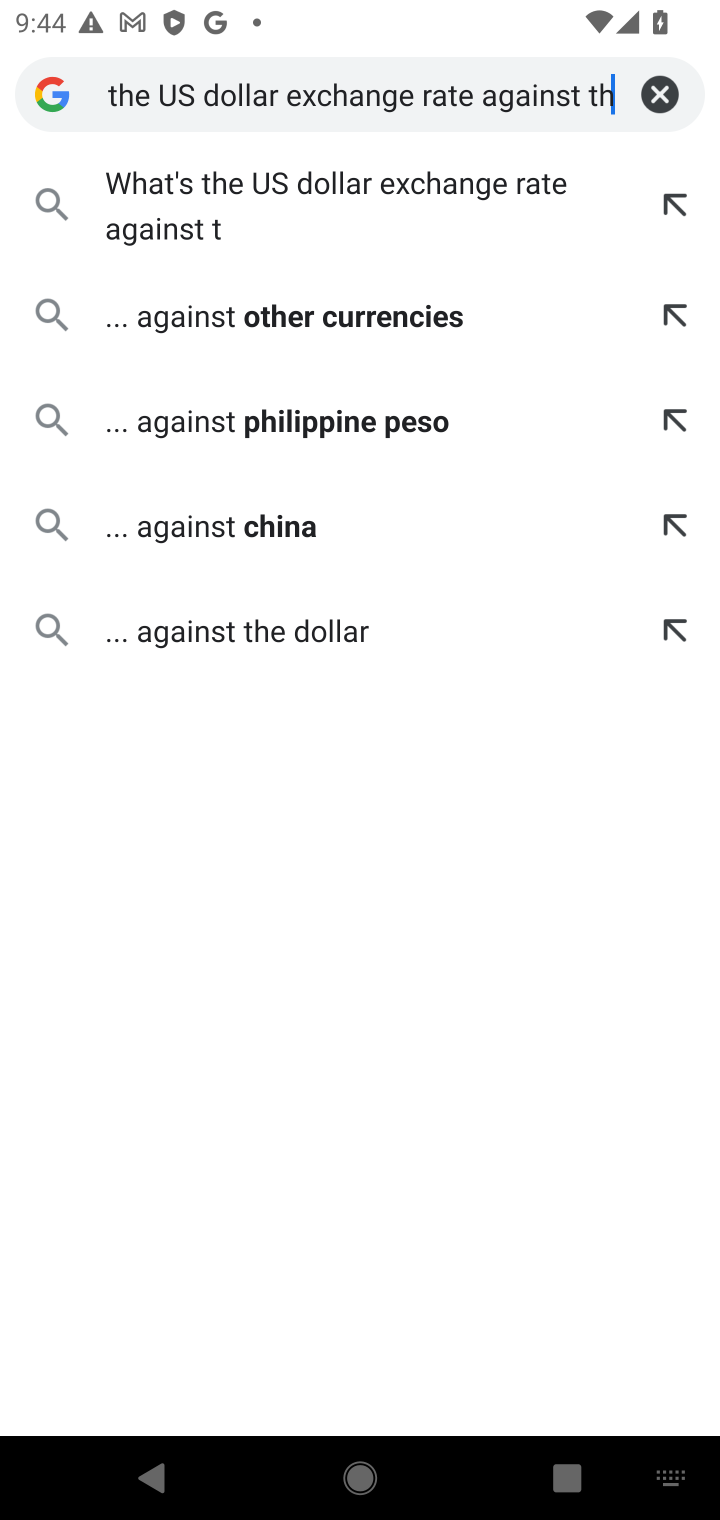}
    \includegraphics[width=0.16\linewidth, height=0.2\textheight, keepaspectratio]{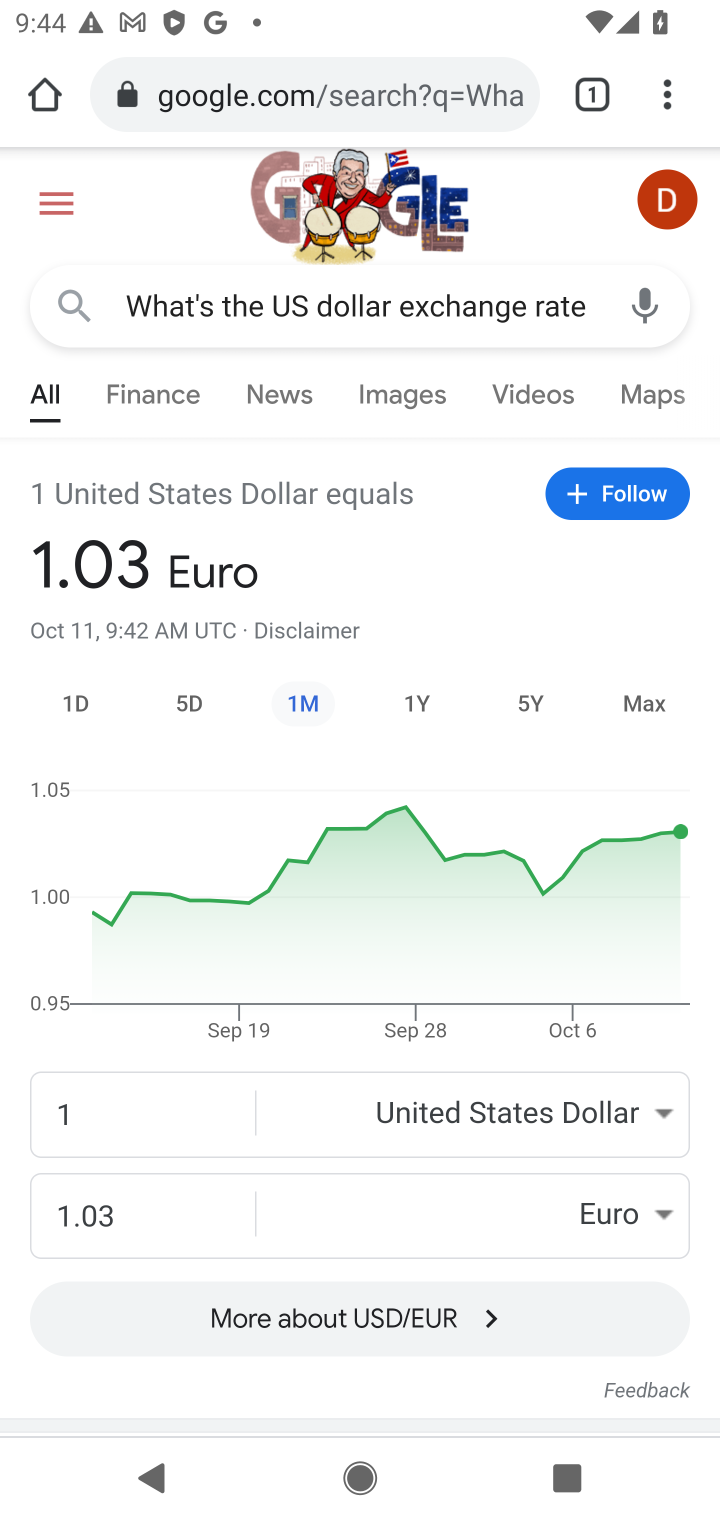}\\
    \vspace{4pt}
    \includegraphics[width=0.16\linewidth, height=0.2\textheight, keepaspectratio]{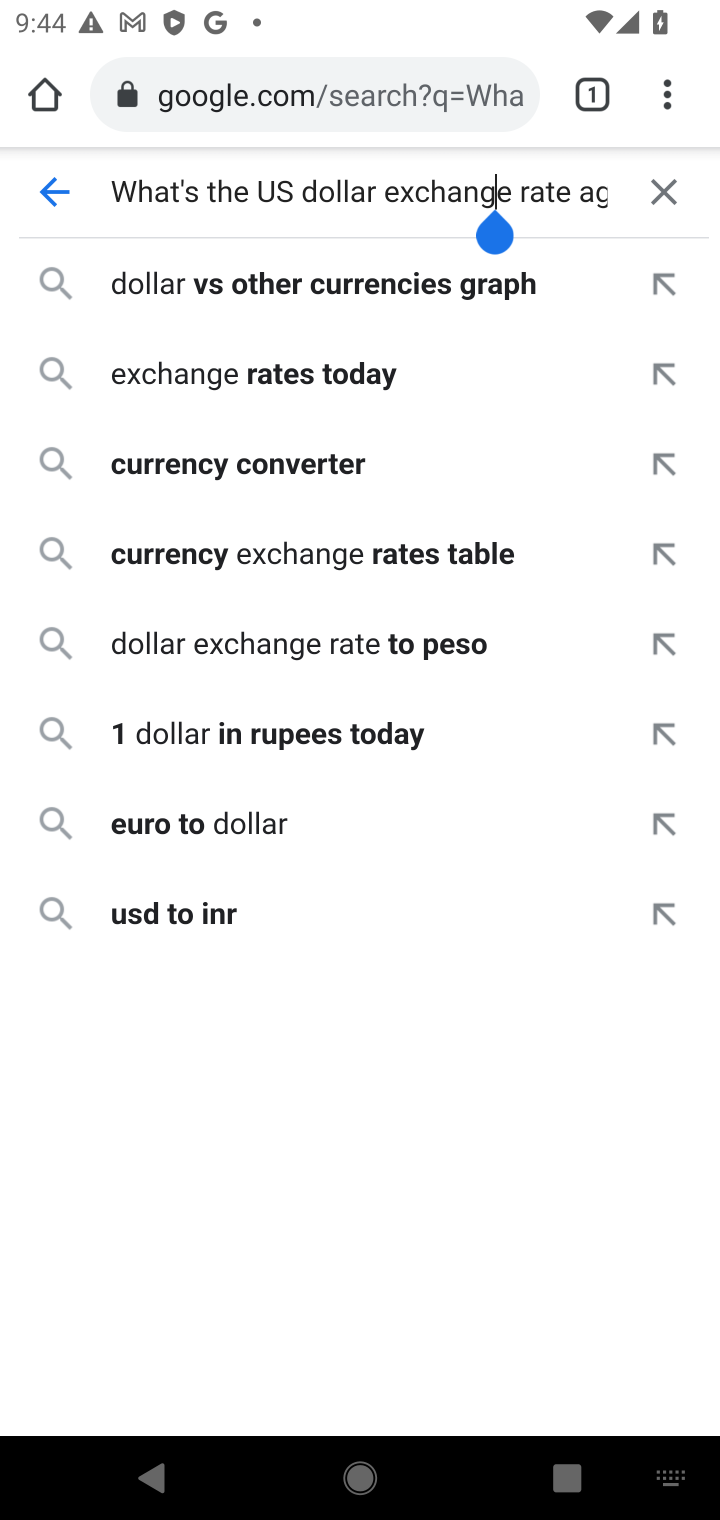}
    \includegraphics[width=0.16\linewidth, height=0.2\textheight, keepaspectratio]{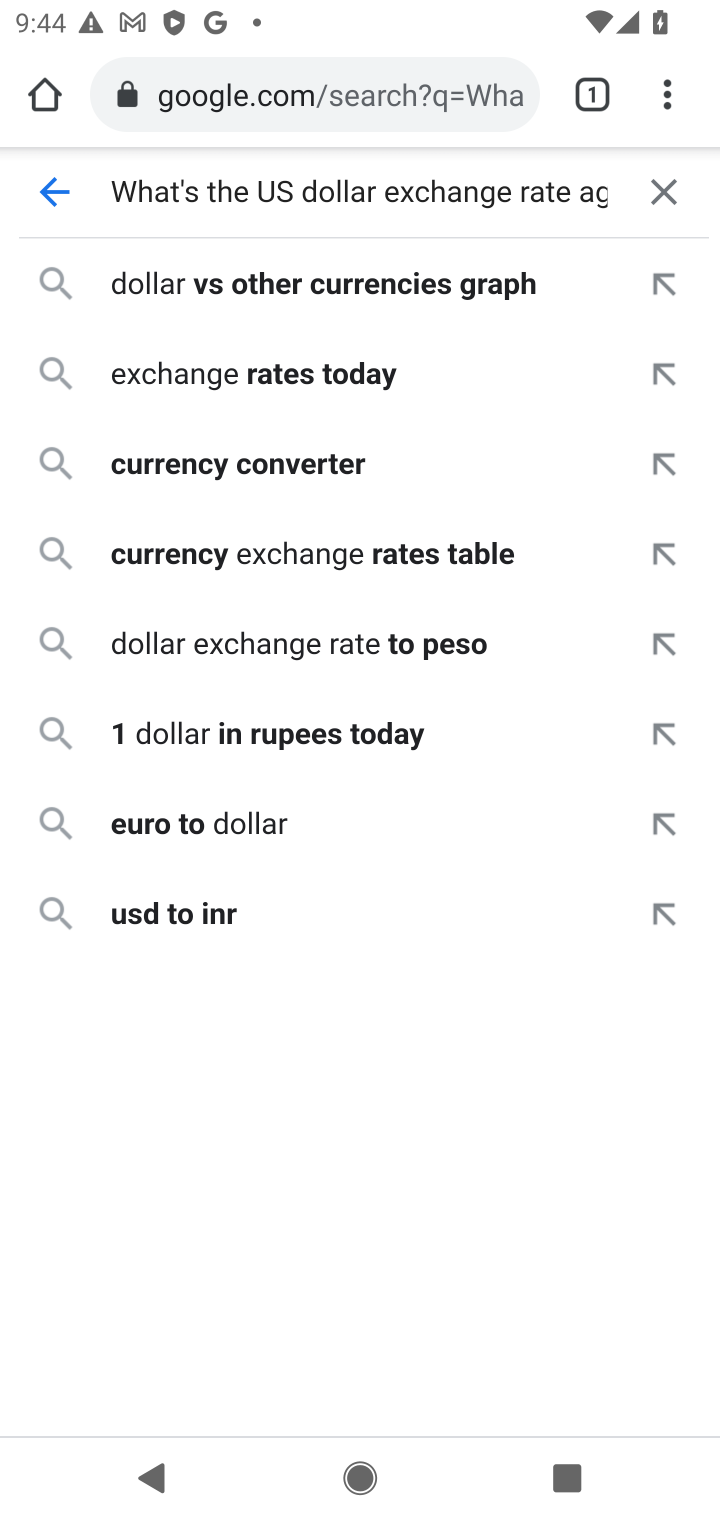}
    \includegraphics[width=0.16\linewidth, height=0.2\textheight, keepaspectratio]{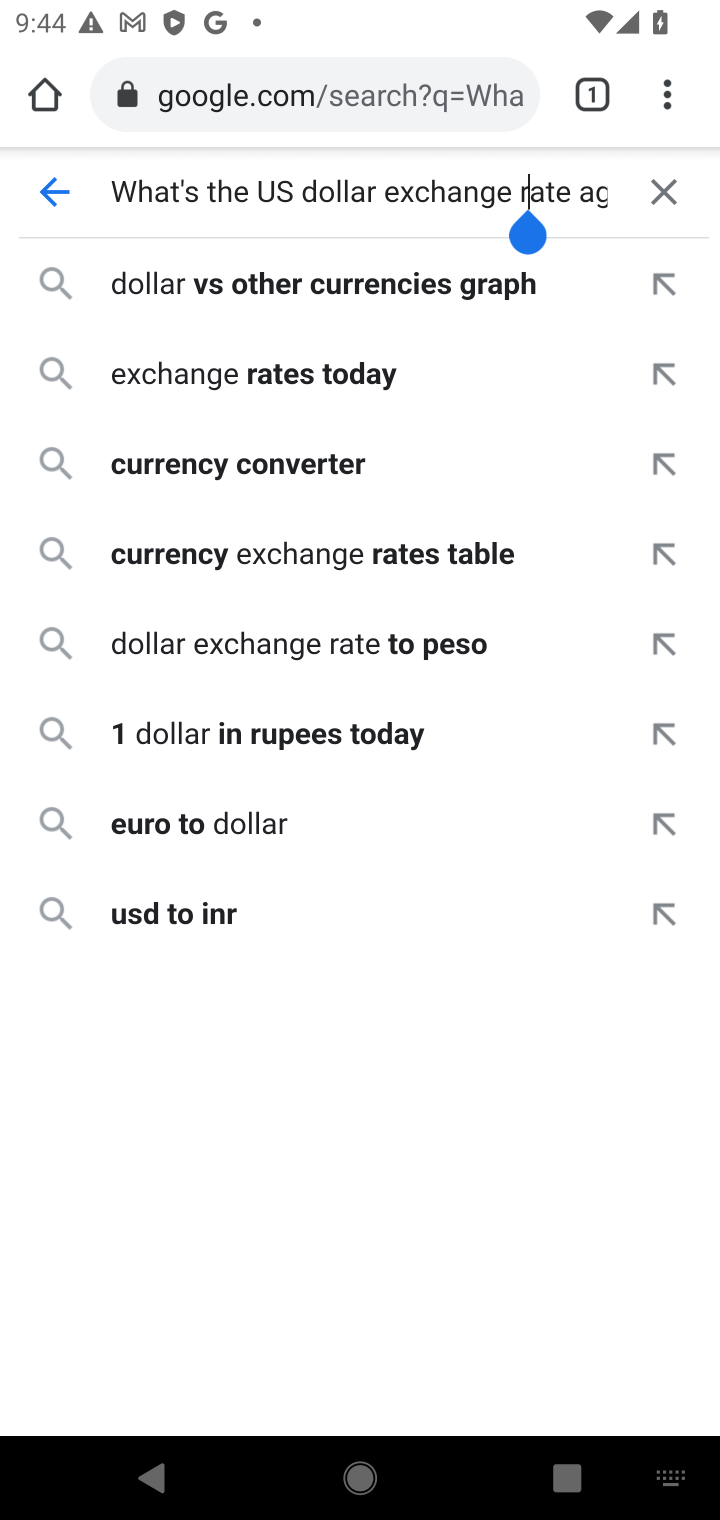}
    \includegraphics[width=0.16\linewidth, height=0.2\textheight, keepaspectratio]{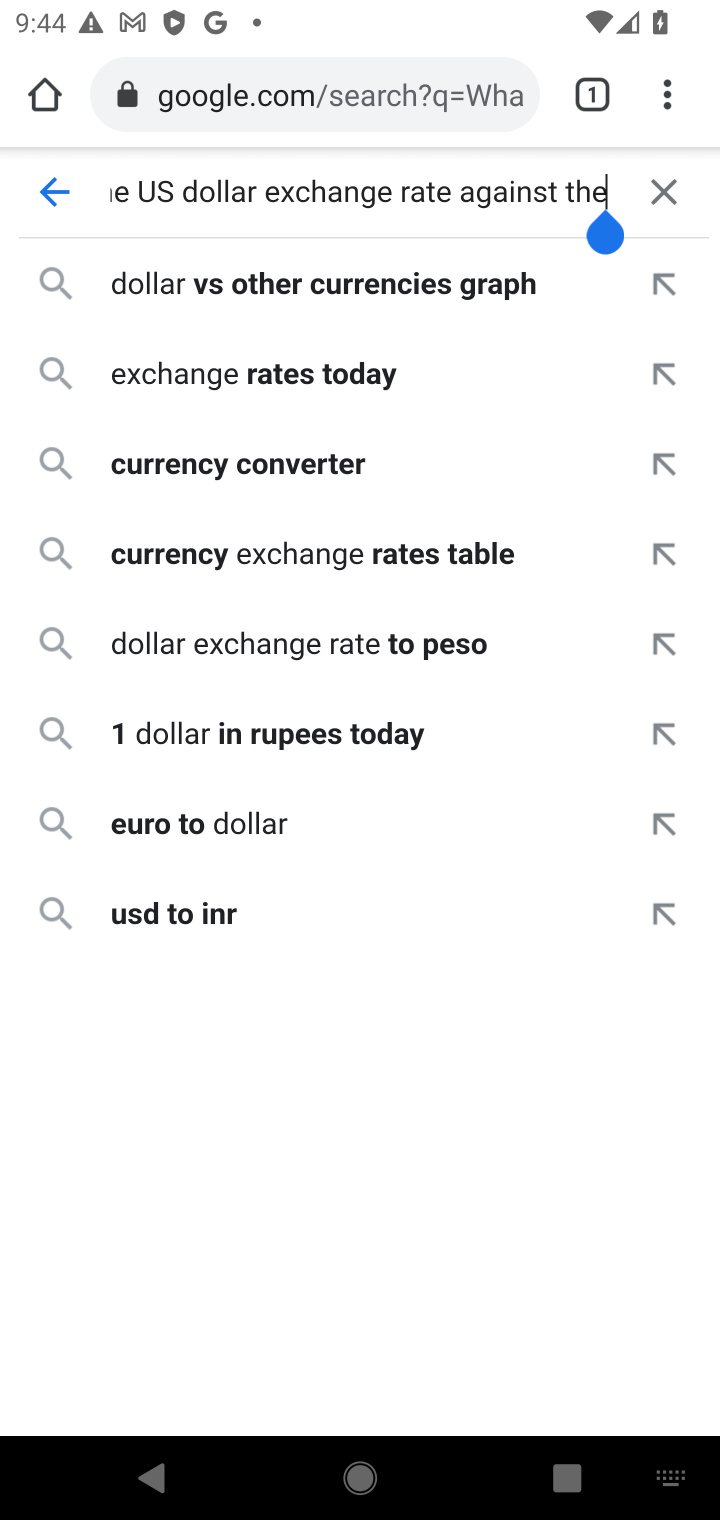}
    \includegraphics[width=0.16\linewidth, height=0.2\textheight, keepaspectratio]{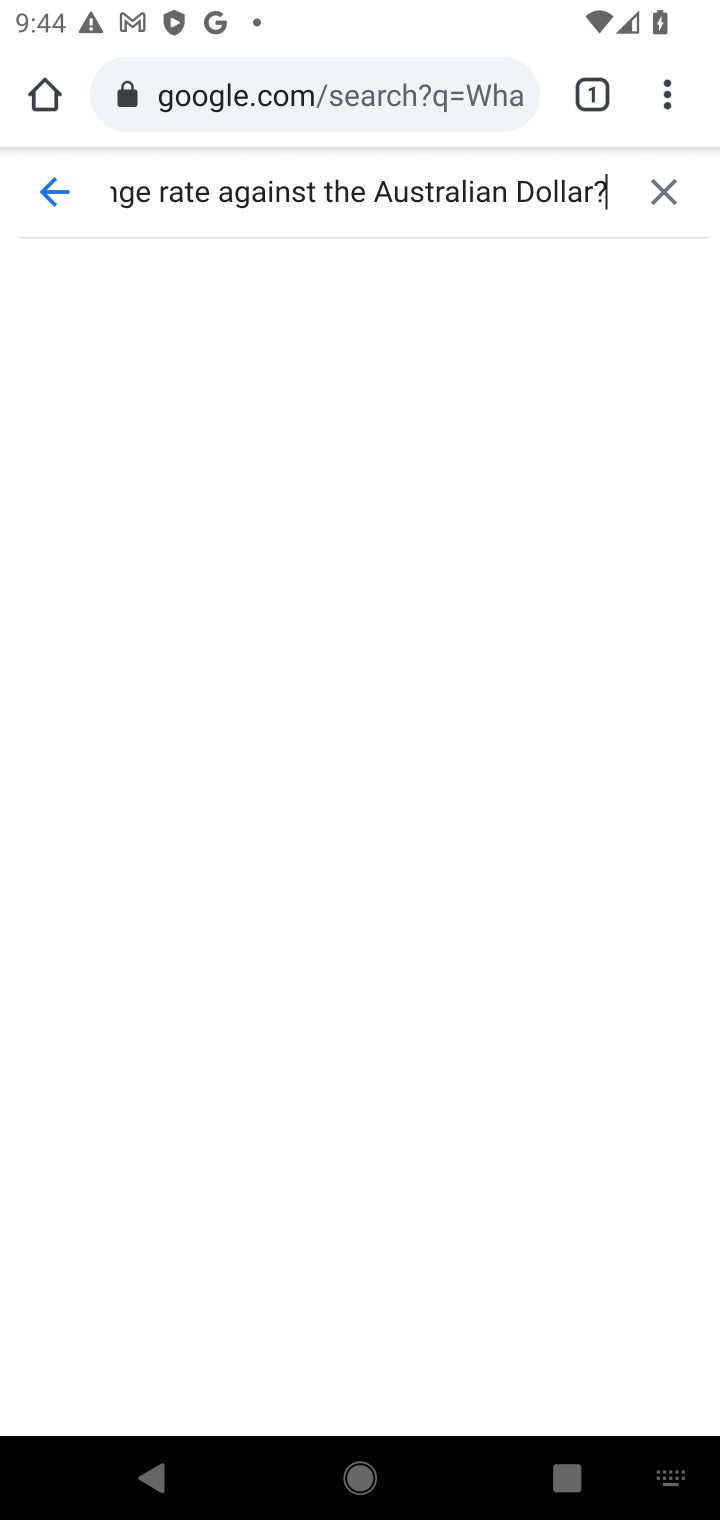}
    \includegraphics[width=0.16\linewidth, height=0.2\textheight, keepaspectratio]{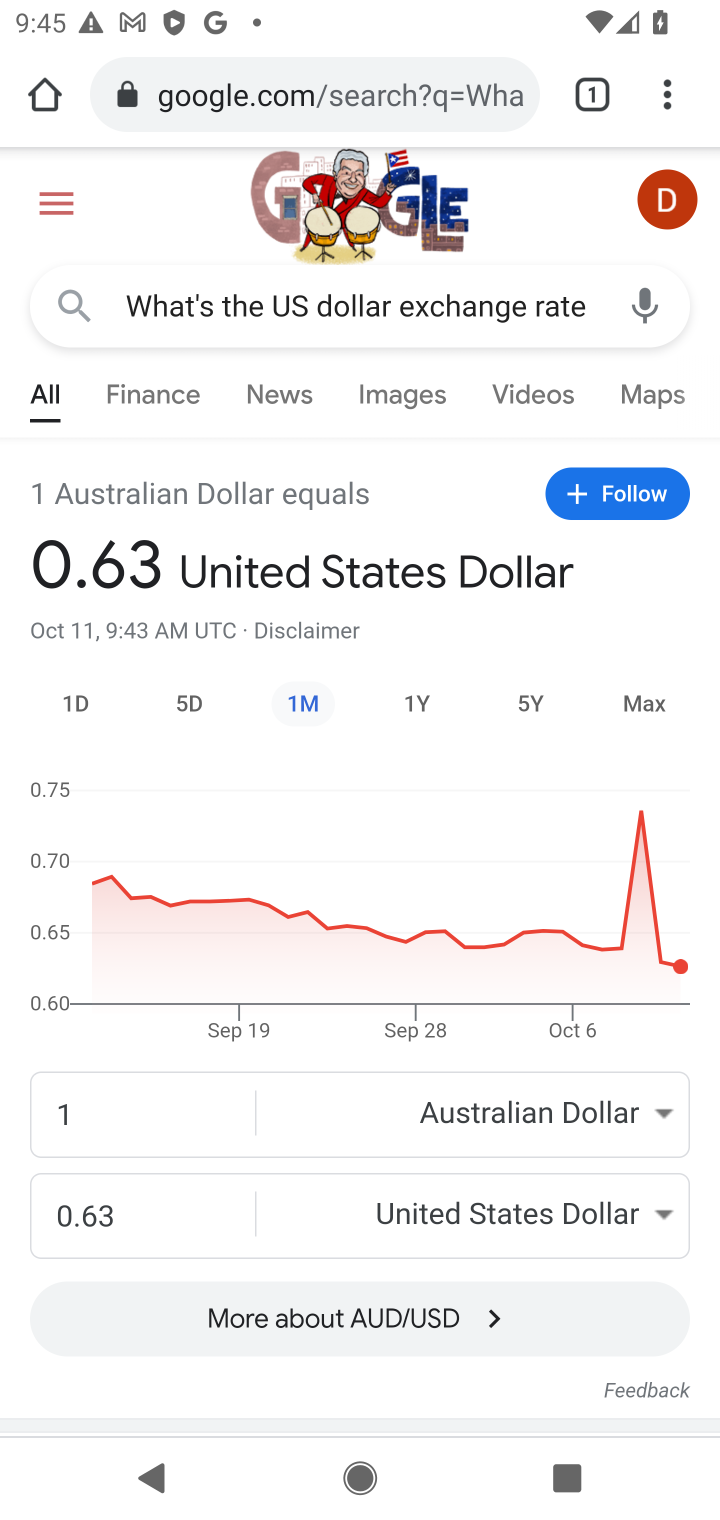}
  \end{minipage}
} \\
\midrule
\begin{minipage}[t][1cm][c]{1.6cm}  
    \centering                    
    \vspace{0.01em}                 
    Thinking                 
    \vspace{0.5em}               
\end{minipage}                   
&                               
\multicolumn{4}{>{\hsize=\dimexpr4\hsize+8\tabcolsep\relax\raggedright\arraybackslash}X}{
The system detects that the user has pending items in their Amazon cart (notification) and is currently checking the USD/AUD exchange rate (interaction trajectory), suggesting a cross-border payment intent. Given the user’s preference for using Wise due to its lower fees (user profile), the system proactively recommends comparing exchange rates through Wise and completing the payment there, potentially helping the user save money.
} \\
\midrule
\begin{minipage}[t][1cm][c]{1.6cm}  
    \centering                    
    \vspace{0.01em}                 
    Text Intent Label                 
    \vspace{0.5em}               
\end{minipage}                   
&                               
\multicolumn{4}{>{\hsize=\dimexpr4\hsize+8\tabcolsep\relax\raggedright\arraybackslash}X}{
It has been detected that you have items waiting for checkout in your Amazon shopping cart and that you are currently checking the USD to AUD exchange rate. Considering that you have a habit of using Wise for cross-border payments—and that Wise may offer more competitive exchange rates—would you like to open the Wise app to check the rate and complete the payment?
} \\

\bottomrule
\end{tabularx}
\caption{Case study.}
\label{tab:case}
\end{table*}

\begin{table*}[t!]
\ContinuedFloat
\begin{CJK}{UTF8}{gbsn}
\centering
\small
\setlength{\tabcolsep}{3pt}
\renewcommand{\arraystretch}{1.05}
\begin{tabularx}{\textwidth}{
    >{\bfseries\centering\arraybackslash}m{1.6cm}
    >{\hsize=1.4\hsize}X
    >{\centering\arraybackslash}m{2.6cm}
    >{\centering\arraybackslash}m{2.6cm}
    >{\centering\arraybackslash}m{2.6cm}
}
\toprule
\begin{minipage}[t][1cm][c]{1.6cm}  
    \centering                    
    \vspace{0.1em}                 
    User Profile                 
    \vspace{0.01em}               
\end{minipage}                   
&                               
\multicolumn{4}{>{\hsize=\dimexpr4\hsize+8\tabcolsep\relax\raggedright\arraybackslash}X}{
根据用户画像和历史行为分析，该用户年龄为25岁，是一位摄影爱好者，频繁浏览数码相机评测网站，近期多次搜索'索尼Alpha 7 IV'。有从美国电商网站'海淘'商品的习惯。该用户手机上安装了'Wise'等国际汇款应用，并有使用记录，偏好使用低手续费的支付渠道。近期有浏览关于旅游和地理知识的内容，例如查询过'中国的首都'。
} \\
\midrule
\begin{minipage}[t][1cm][c]{1.6cm}  
    \centering                    
    \vspace{0.01em}                 
    Device Status                 
    \vspace{0.5em}               
\end{minipage}                   
&                               
\multicolumn{4}{>{\hsize=\dimexpr4\hsize+8\tabcolsep\relax\raggedright\arraybackslash}X}{
设备当前时间为上午9点45分，日期为10月11日，星期二。系统主要语言设定为'英语(澳大利亚)'。地理位置在澳大利亚悉尼。手机电量充足，为91\%，并已连接至稳定的Wi-Fi网络。设备上安装了'亚马逊'、'Wise'、'Gmail'、'Chrome'和'YouTube'等应用。最近的通知消息包括一条来自'亚马逊'的购物车提醒：'您购物车中的商品仍在等待您！'
} \\
\midrule
\begin{minipage}[t][1cm][c]{1.6cm}  
    \centering                    
    \vspace{0.01em}                 
    World Information                 
    \vspace{0.5em}               
\end{minipage}                   
&                               
\multicolumn{4}{>{\hsize=\dimexpr4\hsize+8\tabcolsep\relax\raggedright\arraybackslash}X}{
宏观信息显示，全球金融市场正处于活跃交易时段。根据实时数据，当前美元兑澳元(USD/AUD)的银行间汇率约为'1.58'，但多数支付渠道会在此基础上收取额外费用。一条财经新闻提到，新型金融科技公司提供的跨境支付服务通常比传统银行更具汇率优势。一条国际新闻摘要显示，欧洲航天局正在筹备新的火星探测计划。
} \\
\midrule
\begin{minipage}[t][1cm][c]{1.6cm}  
    \centering                    
    \vspace{0.01em}                 
    Behavioral Trajectories       
    \vspace{15em}               
\end{minipage}                   
&                               
\multicolumn{3}{c}{
  \hspace*{-1cm}
  \begin{minipage}[c]{\linewidth}
    \centering
    \includegraphics[width=0.16\linewidth, height=0.2\textheight, keepaspectratio]{img/case1_trace/GENERAL-9088265727317240175_0.png}
    \includegraphics[width=0.16\linewidth, height=0.2\textheight, keepaspectratio]{img/case1_trace/GENERAL-9088265727317240175_1.png}
    \includegraphics[width=0.16\linewidth, height=0.2\textheight, keepaspectratio]{img/case1_trace/GENERAL-9088265727317240175_2.png}
    \includegraphics[width=0.16\linewidth, height=0.2\textheight, keepaspectratio]{img/case1_trace/GENERAL-9088265727317240175_3.png}
    \includegraphics[width=0.16\linewidth, height=0.2\textheight, keepaspectratio]{img/case1_trace/GENERAL-9088265727317240175_4.png}
    \includegraphics[width=0.16\linewidth, height=0.2\textheight, keepaspectratio]{img/case1_trace/GENERAL-9088265727317240175_5.png}\\
    \vspace{4pt}
    \includegraphics[width=0.16\linewidth, height=0.2\textheight, keepaspectratio]{img/case1_trace/GENERAL-9088265727317240175_6.png}
    \includegraphics[width=0.16\linewidth, height=0.2\textheight, keepaspectratio]{img/case1_trace/GENERAL-9088265727317240175_7.png}
    \includegraphics[width=0.16\linewidth, height=0.2\textheight, keepaspectratio]{img/case1_trace/GENERAL-9088265727317240175_8.png}
    \includegraphics[width=0.16\linewidth, height=0.2\textheight, keepaspectratio]{img/case1_trace/GENERAL-9088265727317240175_9.png}
    \includegraphics[width=0.16\linewidth, height=0.2\textheight, keepaspectratio]{img/case1_trace/GENERAL-9088265727317240175_10.png}
    \includegraphics[width=0.16\linewidth, height=0.2\textheight, keepaspectratio]{img/case1_trace/GENERAL-9088265727317240175_11.png}
  \end{minipage}
} \\
\midrule
\begin{minipage}[t][0.75cm][c]{1.6cm}  
    \centering                    
    \vspace{0.2em}                 
    Thinking                 
    \vspace{0.5em}               
\end{minipage}                   
&                               
\multicolumn{4}{>{\hsize=\dimexpr4\hsize+8\tabcolsep\relax\raggedright\arraybackslash}X}{
检测到用户有亚马逊待支付商品（通知），并正在查询美元/澳元汇率（操作轨迹），表明其有跨境支付意图。结合用户偏好使用低费率Wise（画像）的习惯，主动推荐使用Wise比较汇率并完成支付，以满足其潜在的省钱需求。
} \\
\midrule
\begin{minipage}[t][0.5cm][c]{1.6cm}  
    \centering                    
    \vspace{0.001em}                 
    Text Intent Label                 
    \vspace{1em}               
\end{minipage}                   
&                               
\multicolumn{4}{>{\hsize=\dimexpr4\hsize+8\tabcolsep\relax\raggedright\arraybackslash}X}{
检测到您在亚马逊购物车有商品等待结算，并且您正在查询美元兑换澳元的汇率。考虑到您有使用Wise进行跨境支付的习惯，并且Wise可能提供更优惠的汇率，是否前往Wise App查看汇率并完成支付？
} \\

\bottomrule
\end{tabularx}
\caption{Case study.}
\end{CJK}
\end{table*}

\begin{table*}[t!]
\ContinuedFloat
\centering
\small
\setlength{\tabcolsep}{6pt}
\renewcommand{\arraystretch}{1.2}
\begin{tabularx}{\textwidth}{
    >{\bfseries\raggedright\arraybackslash}p{1.8cm}   
    >{\RaggedRight\arraybackslash}X                    
}

\toprule
\multicolumn{1}{c}{\textbf{Model}} &
\multicolumn{1}{c}{\textbf{Prediction}} \\
\midrule

\begin{minipage}[c]{1.8cm}
    \centering
    \textbf{Label}
\end{minipage}
&
\begin{minipage}[c]{\linewidth}
\begin{lstlisting}[
        basicstyle=\ttfamily\footnotesize,
        breaklines=true,
        breakatwhitespace=true,
        columns=fullflexible,
        aboveskip=0.2em, 
        belowskip=0.2em
    ]
[
  {
    "name": "bank_service_assistance",
    "parameters": {
      "bank_name": "Wise",
      "service_type": "foreign exchange service",
      "location_constraint": "",
      "tasks": "check exchange rate"
    }
  }
]
\end{lstlisting}
\end{minipage} \\
\midrule

\begin{minipage}[c]{1.8cm}
    \centering
    \textbf{GPT-5}
\end{minipage}
&
\emph{No recommendation.} \\
\midrule

\begin{minipage}[c]{1.8cm}
    \centering
    \textbf{o1}
\end{minipage}
&
\emph{No recommendation.} \\
\midrule

\begin{minipage}[c]{1.8cm}
    \centering
    \textbf{Qwen2.5-VL-7B-Instruct}\\
    \textbf{+ Proactive}
\end{minipage}
&
\begin{minipage}[c]{\linewidth}
\begin{lstlisting}[
        basicstyle=\ttfamily\footnotesize,
        breaklines=true,
        breakatwhitespace=true,
        columns=fullflexible,
        aboveskip=0.2em, 
        belowskip=0.2em
    ]
[
  {
    "name": "bank_service_assistance",
    "parameters": {
      "bank_name": "Wise",
      "service_type": "foreign exchange service",
      "location_constraint": "global"
    }
  }
]
\end{lstlisting}
\end{minipage} \\

\bottomrule
\end{tabularx}
\caption{Case study.}
\end{table*}

\tcbset{
    promptbox/.style={
        colback=blue!2,
        colframe=blue!70,
        arc=0pt,
        boxrule=0.5pt,
        left=6pt, right=6pt, top=4pt, bottom=4pt,
        fontupper=\small\ttfamily, 
        enhanced,
        sharp corners,
        boxsep=2pt,
        float,                    
        floatplacement={!tbp},    
        width=\linewidth,         
        breakable,                
        colbacktitle=blue!2, 
        coltitle=black, 
        fonttitle=\bfseries\large, 
        colframetitle=gray!40, 

    }
}

\begin{tcolorbox}[
    colback=blue!2,
    colframe=blue!70,
    colbacktitle=gray!40,
    coltitle=black,
    fontupper=\small\ttfamily,
    fonttitle=\bfseries\large,
    float*,
    floatplacement={!tbh},
    width=\textwidth,
    breakable,                    
    enhanced,                      
    title=Prompt for Generating Contextual Information
]

\textbf{Step1: Prompt for generating scenarios.}\\
\\
As a scenario design specialist, you are tasked with creating proactive intelligence scenarios for mobile devices. Specifically, given a particular operation instruction, you must construct a plausible preceding usage context where the instruction naturally emerges as the next proactive recommendation the device should immediately suggest.
\\
\\
Mobile Operation Instruction: \{instruction\} 

Create a concise scenario description (within 200 words) requiring:

- The scenario must not explicitly include executing the instruction or the user requesting it. Instead, the instruction should emerge naturally as the subsequent action.

- The scenario must logically and evidently lead to the instruction being the next immediate step, with a clear and justifiable inference path.

- The corresponding instruction must represent the most critical and urgent action required in the generated scenario. For example, if the instruction is "Navigate to the driving route to Shaoguan Danxia Mountain," the scenario should involve an imminent departure or active route planning, not unrelated activities.

- The scenario must be definable from the device's perspective as a sequence of operations.

- The scenario must be detectable and identifiable through mobile device data, not user subjective intent. For instance, for "Navigate to the driving route to Shaoguan Danxia Mountain," the scenario should involve active route planning on the phone, not the user thinking "I want to go to Shaoguan Danxia Mountain."

- Incorporate device-relevant details (e.g., device type, operational environment, user-device interaction methods) to enhance realism and actionability.

- Avoid irrelevant content; ensure the scenario focuses on the practical application of the mobile operation instruction.

- Provide the complete scenario description directly, without any prefixes, explanations, or additional content.

- The scenario must exclude user subjective actions or feelings, focusing solely on user-device interactions and environmentally perceptible information.
\\
\\
\textbf{Step2: Prompt for Generating Scenarios and Requires.}
\\
\\
You are a scenario design specialist tasked with specifying data requirements for mobile proactive intelligence scenarios.

Based on the following mobile operation instruction and scenario, analyze the essential device data required:
\\
\\
Mobile Operation Instruction: \{instruction\}

Scenario Description: \{scenario\}
\\
\\
List the key device data necessary for executing this instruction (within 100 words), requiring:

- Include only data obtainable exclusively via the mobile device (e.g., GPS, time, calendar, application usage history).

- The data must directly contribute to instruction execution.

- Exclude user subjective information or unobtainable data.
\\
\\
Provide the data list directly, without any prefixes, explanations, or additional content.
\\
\\
\textbf{Step3: Prompt for generating trigger, condition, and profile.}
\\
\\
As a scenario design specialist constructing a proactive intelligence dataset, you are to formalize conditional scenarios for mobile devices based on specified operations. Given a **scenario**, a **specific operation instruction** within that scenario, and **potentially involved device information**, format it into the following structure: Your conditional scenario comprises three components: **trigger**, **condition**, and **profile**.

- "trigger": The trigger is one or a series of instantaneous actions or scenarios detectable by the device. Upon detection, the device activates its built-in large model to determine if a proactive recommendation is needed.

- "condition": The condition(s), which can be instantaneous or sustained, are the information considered by the device's model after triggering to reach a final decision.

- "profile": The user profile is a brief character description containing basic information. If necessary, it can include why this person would perform the operation in the given context and their core need or motivation, but avoid overly explicit direction.

Present triggers and conditions as bullet points, and the profile as a single string, all within a dictionary. Refer to the example below.

\begin{lstlisting}[language={}]
Example: For the operation instruction "Remind the user to charge," the output should be:
{
    "trigger": [
        "Phone battery level drops below 40%",
        "User is about to leave home"
    ],
    "condition": [
        "User is actively using the phone",
        "Current time is 10:00 AM"
    ],
    "profile": "User is a 30-year-old office worker with a busy schedule, often 
    forgets to charge the phone, and experiences battery anxiety.",
    "expected_recommendation": "Remind the user to charge"
}
\end{lstlisting}

Your output must consist solely of the JSON list as shown above. Do not include any other information.

You must output only one JSON list containing only one dictionary representing one scenario.

- Ensure the provided trigger, condition, and profile are all logically connected to the given operation instruction, and the scenario is plausible.

- Maintain high plausibility for triggers and conditions. Avoid unrealistic triggers like "Phone microphone detects user's stomach rumbling."

- Ensure the mobile operation instruction is the most urgent and logical action given your trigger and conditions, with a complete logical chain.

- Triggers must be instantaneous events. Historical or sustained information (e.g., "Relevant browsing history exists in social media," "It is the weekend") is unsuitable as triggers and should be conditions.

- Enclose all JSON keys and values in double quotes **""**. Use single quotes **'** for internal quotations within JSON strings to prevent parsing errors.

- If the given mobile operation instruction is too vague, you may refine it appropriately. Ensure the final expected\_recommendation is clear and executable on a mobile device.
\\
\\
Provided Information:

Mobile Operation Scenario: \{task\_scenario\}

Mobile Operation Instruction: \{task\_recommend\}

Device Information: \{task\_require\}
\\
\\
\textbf{Step4: Prompt for generating user profile, device status, world information, and textual behavioral trajectories of text data.}
\\
\\
The user input is a JSON dictionary containing four fields: "condition", "trigger", "profile", and "expected\_recommendation". Here, "condition" specifies the conditional context for the mobile proactive intelligence task, "trigger" indicates the triggering events, "profile" describes the user's personal characteristics, and "expected\_recommendation" defines the desired recommendation output.

Your task is to supplement and refine the data based on this information. You must generate the four sub-fields under "reference\_information": "profile", "phone", "world", and "trace". Produce three distinct data instances, ensuring their core content varies.
\\
\\
Specifications:
- The "profile" field is an expansion of the provided "profile" field (Note: The generated field is distinct from the original).

- The "phone" field captures device information (e.g., basic device status, current state, internal application data, pending tasks).

- The "world" field describes external context (e.g., weather, holidays).

- The "trace" field records the user's recent behavioral trajectory, specifically within the last ten minutes, as atomic operations (e.g., taps, swipes, button presses) captured by the mobile device.
\\
\\
Requirements:

1. Maintain the exact output format as provided, which is a list containing JSON dictionaries. The "benchmark\_metadata" field must remain identical to the input; do not modify it.

2. Enclose all JSON keys and values in double quotes **""**. Use single quotes **'** for internal quotations within JSON strings (e.g., "profile":"This is an 'example'.") to ensure proper parsing.

3. Describe all generated valid information using natural language.

4. The "trace" must consist solely of atomic operation sequences perceptible by the mobile device, even if this omits some user intent. Avoid vague terms like "a certain," "one," or "an item." Specify operational objects concretely. **Each step must begin with "Tap," "Swipe," "Long Press," or "Input Text."** The device must be active (screen on) for each recorded action.

5. Incorporate timestamps precise to the second, adhering to realistic intervals. Most consecutive operations should be separated by less than 5 seconds, with shorter steps spaced 1-2 seconds apart.

6. The "trace" must not include the recommendation behavior itself, only the preceding actions. The recommendation should occur immediately after the final trace step; if recommendable earlier, the trace is invalid.

7. If the "trigger" includes a device operation, place it as the final step in the "trace." For example, if the trigger is ["User arrived at location", "User unlocked phone"], then "User unlocked phone" should be the last trace entry.

8. Integrate the "condition" field implicitly within the "phone", "world", and "trace" fields during generation.

9. Ensure all generated content is perceptible by the mobile device. Exclude user mental states or personality traits from "profile," and physical device details from "phone."

10. Target approximate lengths: "profile" $\sim$30 Chinese characters, "phone" $\sim$30 Chinese characters, "world" $\sim$30 Chinese characters, "trace" 5-10 entries.

\begin{lstlisting}[language={}]
The output format is as follows:
[{
"benchmark_metadata": { 
    "condition": ["condition"],
    "trigger": ["trigger"],
    "profile": ["profile"],
    "expected_recommendation":"Remind user of low battery level and suggest charging",
    },
"reference_information": {
    "profile": "Basic information: Name, identification number, contact details, phone 
    number, email, device fingerprint, facial recognition, and other long-term/short-
    term memory data",
    "phone": "Battery level, charging status, mobile data, WiFi, Bluetooth, lock 
    screen status, foreground application, installed third-party applications, time, 
    language, dark mode, geographical location, SMS, push notifications, and 
    other messaging information",
    "world": "Weather, holidays, traffic information, exchange rates, international 
    news, etc.",
    "trace": ["Clicks, swipes, making phone calls, sending text messages, installing 
    applications, browsing web pages, taking photos, uploading files, WeChat payments, 
    etc."]
    }
},
{
"benchmark_metadata": { 
    "condition": ["condition"],
    "trigger": ["trigger"],
    "profile": ["profile"],
    "expected_recommendation":"Remind user of low battery level and suggest charging",
    },
"reference_information": {
    "profile": "Basic information: Name, identification number, contact details, phone 
    number, email, device fingerprint, facial recognition, and other long-term/short-
    term memory data",
    "phone": "Battery level, charging status, mobile data, WiFi, Bluetooth, lock 
    screen status, foreground application, installed third-party applications, time,
    language, dark mode, geographical location, SMS, push notifications, and other
    messaging information",
    "world": "Weather, holidays, traffic information, exchange rates, international 
    news, etc.",
    "trace": ["Clicks, swipes, making phone calls, sending text messages, installing 
    applications, browsing web pages, taking photos, uploading files, WeChat payments,
    etc."]
    }
},
{
"benchmark_metadata": { 
    "condition": ["condition"],
    "trigger": ["trigger"],
    "profile": ["profile"],
    "expected_recommendation":"Remind user of low battery level and suggest charging",
    },
"reference_information": {
    "profile": "Basic information: Name, identification number, contact details, phone
    number, email, device fingerprint, facial recognition, and other long-term/short-
    term memory data",
    "phone": "Battery level, charging status, mobile data, WiFi, Bluetooth, lock
    screen status, foreground application, installed third-party applications, time,
    language, dark mode, geographical location, SMS, push notifications, and other
    messaging information",
    "world": "Weather, holidays, traffic information, exchange rates, international
    news, etc.",
    "trace": ["Clicks, swipes, making phone calls, sending text messages, installing
    applications, browsing web pages, taking photos, uploading files, WeChat payments,
    etc."]
    }
}]

\end{lstlisting}

Input information:

    \{info\}
\\
\\
\textbf{Step5: Prompt for generating user profile, device status, world information, and GUI behavioral trajectories of multimodal data.}
\\
\\
You function as an autonomous intelligent data generator. Based on user-provided triggers, conditions, and task profiles, your objective is to generate a proactive intelligent recommendation task conforming to the following structure:

\begin{lstlisting}[language={}]
{
    "benchmark_metadata": { 
        "difficulty_level": 1,
        "condition": ["condition"],
        "trigger": ["trigger"],
        "profile": ["profile"],
        "expected_recommendation":"Remind user of low battery level and suggest charging",
    },
    "reference_information": {
        "profile": "Basic information: Name, identification number, contact details, phone
        number, email, device fingerprint, facial recognition, and other long-term/short-
        term memory data",
        "phone": "Battery level, charging status, mobile data, WiFi, Bluetooth, lock
        screen status, foreground application, installed third-party applications, time,
        language, dark mode, geographical location, SMS, push notifications, and other
        messaging information",
        "world": "Weather, holidays, traffic information, exchange rates, international
        news, etc.",
        "trace": [
            {"source": "text", "text": "Clicks, swipes, making phone calls, sending text
            messages, installing applications, browsing web pages, taking photos,
            uploading files, WeChat payments, etc."},
            {"source": "picture", "picture": "xxxxxxx.png"}
        ]
    },
    "option": {
        "expected_recommendation": ["expected_recommendation"]
    }
}
\end{lstlisting}

The user-provided triggers, conditions, and task profiles are as follows: \{useful\_information\}
\\
\\
Please generate the 'profile', 'phone', and 'world' sections within 'reference\_information' according to the specified format.
\\
\\
Specifications:

- 'profile' encompasses user personal information.

- 'phone' describes the mobile device's environmental context.

- 'world' captures external world information.

- 'trace' represents the user's behavioral trajectory.

- The final task should be inferable from this consolidated information.
\\
\\
Requirements:

1. Adhere strictly to the JSON output format and maintain the original JSON structure.

2. Describe each generated section of valid information using natural language.

3. The 'trace' field contains pre-existing images. Meticulously identify information within these images (e.g., battery level, time) to ensure generated content in 'profile', 'phone', and 'world' does not conflict with the image data.

4. Target approximate lengths: 'profile' ~30 Chinese characters, 'phone' ~30 Chinese characters, 'world' ~30 Chinese characters.

5. The 'phone' field must contain device environment information obtainable by the phone. The 'world' field must contain acquirable external information, excluding user mental states, plans, or other content inaccessible to the device.
\\
\\
Output the complete modified data in strict JSON format without additional explanations. Ensure:

1. All keys and string values are enclosed in double quotes (`"`).

2. Key-value pairs are separated by commas.

3. No trailing comma follows the last key-value pair.

4. Comments (e.g., `//` or `/* */`) are prohibited.

5. Use single quotes **''** for internal quotations within JSON strings (e.g., "profile":"This is an 'example'.").

\end{tcolorbox}

\begin{tcolorbox}[
    colback=blue!2,
    colframe=blue!70,
    colbacktitle=gray!40,
    coltitle=black,
    fontupper=\small\ttfamily,
    fonttitle=\bfseries\large,
    float*,
    floatplacement={!tbh},
    width=\textwidth,
    breakable,                    
    enhanced,                      
    title=Prompt for Generating Potential Intentions
]
\textbf{Step1: Prompt for generating 30 candidates.}
\\
\\
<Role>
    
You are a professional mobile intelligent assistant evaluation specialist, specializing in analyzing the decision-making processes of proactive mobile recommendation systems. Your task involves predicting user operational intentions based on multi-dimensional information and determining whether proactive recommendations should be issued to the user.
\\
\\
Core Decision Principles:

1. Explicit and urgent user needs → Proactively recommend corresponding actions.

2. Ambiguous or non-urgent needs → Refrain from making recommendations.

</Role>
\\
\\
<Task>

Conduct a comprehensive analysis based on the following four categories of input information:

- **User Profile Information (profile)**: User preferences, usage habits, historical behavior patterns.

- **Mobile Device Context (phone)**: Current device status, network environment, battery level, etc.

- **External Context (world)**: Time, location, weather, schedule, and other environmental factors.

- **User Behavioral Trajectory (trace)**: Recent operation sequences and behavioral patterns.
\\
\\
Your tasks are:

1. Perform an in-depth analysis of the user's genuine needs and intentions.

2. Assess the urgency and clarity of these needs.

3. Determine whether a proactive recommendation is warranted.

</Task>
\\
\\
<Analysis\_Framework>

Please conduct your analysis according to the following framework:
\\
\\
1. **Need Identification**: Identify the user's potential needs from the behavioral trajectory.

2. **Urgency Assessment**: Judge whether the identified need requires immediate attention.

3. **Recommendation Decision**: Based on the above analysis, decide on the recommendation content or whether to recommend at all.

</Analysis\_Framework>

\begin{lstlisting}[language={}]
<Output_Requirements>
Output strictly in JSON format, including the following field:

{{
    "expected_recommendation": "(Recommendation Content)/(No Recommendation)"
}}
\end{lstlisting}

**Format Specifications**:

- All keys and string values must be enclosed in double quotes ("").

- Key-value pairs should be separated by commas, with no comma following the last pair.

- Use single quotes ('') for any quotations within JSON string values.

- Comments (e.g., // or /* */) are prohibited.

- If no recommendation is warranted, the "expected\_recommendation" field should be "No Recommendation".

- Please respond in Chinese.

</Output\_Requirements>
\\
\\
<Input\_Data>

\begin{lstlisting}[language={}]
### Reference Data:
{each_data['reference_information']}
\end{lstlisting}

</Input\_Data> \\
\\
\textbf{Step2: Prompt for clustering centroids.}
\\
\\
You are a professional data clustering specialist tasked with clustering mobile operation instructions based on their core semantic meaning.

The clustering principle is to categorize instructions according to the primary action's core semantics, disregarding minor descriptive variations.
\\
\\
Existing cluster categories:
\begin{lstlisting}[language={}]
{existing_cluster_desc}
\end{lstlisting}
Instructions requiring clustering:
\begin{lstlisting}[language={}]
{instructions_text}
\end{lstlisting}
Please adhere to the following clustering requirements:

1. When analyzing instructions, focus solely on the core operation, ignoring irrelevant context, impact, function, and prefatory phrases such as "We suggest you XXX" or "We recommend you XXX."

2. If an instruction's core operational semantics align or are similar to an existing category, and its operational object is substantially consistent, it can be assigned to that category. Prioritize the core function of the operation during clustering, overlooking minor phrasing differences.

3. If the core operational semantics of an instruction do not match any existing category, create a new one. When describing a new category, provide a concise description of the core operation content and object, omitting detailed information and irrelevant environmental context.

Example: The recommendation "Detected that you are sending an email to IT support. Would you like to automatically generate an email template containing error details like 'ERR\_CONNECTION\_TIMED\_OUT'?" can be clustered into the category "Automatically generate fault report emails."

\begin{lstlisting}[language={}]
Please output the results in JSON format as follows: 
{

    "clustering_results": [
    
        {
        
            "instruction_index": 1,
            
            "instruction": "Instruction content",
            
            "cluster_id": "Category ID",
            
            "is_new_cluster": true/false,
            
            "cluster_description": "Category description (if it is a new category)"
        
        }
    ]
    
}
\end{lstlisting}

Notes:

- Use numerical identifiers for 'cluster\_id'. For new categories, use the next available number.

- 'is\_new\_cluster' indicates whether a new category was created.

- 'cluster\_description' is only required when 'is\_new\_cluster' is true.

- Ensure all keys and values in the JSON output are enclosed in double quotes **""**. Use single quotes **''** for any quotations within JSON string values to prevent parsing issues.
\\
\\
\textbf{Step3: Prompt for generating thinking processes.}
\\
\\
You are a data specialist processing mobile proactive intelligence task data. Proactive recommendation refers to the device's capability to anticipate user needs and deliver suggestions based on an analysis of multimodal data, including user profile, behavioral trajectories, device status, and world information, prior to any user request.
\\
\\
Your task is to supplement the reasoning process. Based on the provided recommendation instruction and antecedent information (user profile, behavioral trajectories, device status, and world information), reconstruct the logical inference from precondition analysis to recommendation generation.

\begin{lstlisting}[language={}]
Output format:
```json

{{

    "thinking": "Supplemented reasoning process"
    
}}
\end{lstlisting}

Output the complete data in strict JSON format without additional explanations. Ensure:

1. All keys and string values are enclosed in double quotes ("");

2. Key-value pairs are separated by commas;

3. No trailing comma follows the last key-value pair;

4. Comments (e.g., // or /* */) are prohibited;

5. Use single quotes '' for internal quotations within JSON strings (e.g., "profile":"This is an 'example'.").
\\
\\
Important Notes:
\\
\\
1. The supplemented reasoning process should be concise, limited to 100 words.

2. If the original recommendation instruction is empty (indicating no recommendation is required), you must still provide reasoning explaining this determination.
\begin{lstlisting}[language={}] 
Original recommendation instruction: {instruction}
\end{lstlisting}
\begin{lstlisting}[language={}] 
Antecedent information: {data['reference_information']} 
\end{lstlisting}

\end{tcolorbox}

\begin{tcolorbox}[
    colback=blue!2,
    colframe=blue!70,
    colbacktitle=gray!40,
    coltitle=black,
    fontupper=\small\ttfamily,
    fonttitle=\bfseries\large,
    float*,
    floatplacement={!tbh},
    breakable,                    
    enhanced,                      
    width=\textwidth,
    title=Prompt for Adding Interfering Information
]
You function as a data generation specialist, currently engaged in processing datasets for mobile proactive intelligence tasks. These tasks involve the mobile device proactively identifying user needs and making recommendations by analyzing user behaviors, contextual information, and other relevant data.

Your specific task is to inject varying types of irrelevant information into each data entry to create noisy training datasets, thereby elongating the data length. However, this must be accomplished without altering the original recommendation behavior inherent to the data. The current trigger information is derived from the 'condition', 'trigger', and 'profile' fields within the 'benchmark\_metadata'. The desired output recommendation task for the large language model is specified in the 'expected\_recommendation' field of the 'benchmark\_metadata'. The information source utilized by the large model to infer this trigger is the 'reference\_information' field. You are required to insert irrelevant information specifically within the 'reference\_information' field.

The 'profile' field is an expansion based on the initially provided "profile" field (Note: The field you generate is distinct from the original "profile" field). The 'phone' field encapsulates device information (including basic device status, current state, internal application data, pending tasks, etc.). The 'world' field describes external contextual information (including weather conditions, holidays, etc.). The 'trace' field records the user's recent behavioral trajectories, specifically ** within the last ten minutes**, as captured by the mobile device.

Requirements:

1.You may rephrase the original effective information but must preserve its semantic meaning.

2. Irrelevant information should be task-independent, and its insertion must not impact the original recommendation behavior.

3. Textual noise should be inserted exclusively into the 'profile', 'phone', 'world', and 'trace' sub-fields within 'reference\_information'. No new fields should be added. You must strive to diversify the content of these fields rather than merely elaborating on existing content. For instance, if the original "profile" indicates the user has battery anxiety, your expansion should not elaborate on this anxiety but introduce other unrelated information.

4. Maintain the output format as JSON and preserve the original JSON structure.

5. Ensure all keys and values in the JSON format are enclosed in double quotes **""**. Use single quotes **''** for any quotations within JSON string values (e.g., "profile":"This is an 'example'."). Failure to adhere to this may result in JSON parsing errors, leading to professional dissatisfaction and potential termination.

6. The 'trace' information must constitute a sequence of genuine, atomic operations perceptible by the mobile device, even if this results in the loss of some user intent information. Avoid vague descriptors like "a certain," "one," or "an item." Instead, specify the operational objects concretely. **Each step must commence with verbs such as "Tap," "Swipe," "Long Press," or "Input Text."** Incorporate timestamps precise to the second, conforming to realistic scenarios. The time intervals between consecutive operations should reflect realistic usage patterns, with most adjacent operations separated by less than 5 seconds. For each recorded action, the mobile device must be in an active state (screen on), not in a sleep or locked state.

7. Guarantee that all generated content is perceptible by the mobile device. For example, the 'profile' should not include user mental states or personality preferences; the 'phone' field should not contain physical device specifications.

8. After noise insertion, the 'phone' approximately {random.choice(phone\_num)} characters, 'profile' text should approximate \{random.choice(profile\_num)\} characters, 'world' approximately {random.choice(env\_num)} characters, and 'trace' approximately {random.choice(trace\_num)} entries.

9. Since the final step in the 'trace' constitutes the trigger source, irrelevant information can only be inserted *before* this final step, not after it.
\\

The data format is as follows:

\{data\_struct\}

Below is the data text requiring processing. Please note that the provided data text lacks the "difficulty\_level" field within "benchmark\_metadata". You are required to supplement this field. For this specific data entry, the "difficulty\_level" should be set to \{i\}.
The current data text is:
\begin{lstlisting}[language={}]
{data\_information[i]}
\end{lstlisting}
\end{tcolorbox}

\begin{tcolorbox}[
    colback=blue!2,
    colframe=blue!70,
    colbacktitle=gray!40,
    coltitle=black,
    fontupper=\small\ttfamily,
    fonttitle=\bfseries\large,
    float*,
    breakable,                    
    enhanced,                      
    floatplacement={!tbh},
    width=\textwidth,
    title=Prompt for Mapping to Function
]
You are a data expert processing data related to mobile proactive intelligence tasks. These tasks refer to the capability of a mobile device to proactively identify user needs and make recommendations based on user behavior, environmental context, and other relevant information.

Your objective is to map a given mobile instruction to the most appropriate GUI control function(s). You are provided with an instruction and a set of control functions. Your task is to identify one or more functions that best represent the instruction, populate the parameters of each function accordingly, and formalize the instruction into executable function calls.
\begin{lstlisting}[language={}] 
Output Format:
    ```json
    {{
        "function": [
            {{
                "name": "function_name",
                "parameters": {{
                    "param1": "value1",
                    "param2": "value2",
                    "param3": "",
                    ...
                }}
            }},
            ...
        ]
    }}
\end{lstlisting}
Please ensure the output adheres strictly to the JSON format without any additional explanations. The following must be observed:

1. All keys and string values must be enclosed in double quotes ("").

2. Key-value pairs must be separated by commas.

3. No trailing comma is allowed after the last key-value pair.

4. Comments (e.g., // or /* */) are prohibited.

5. Any quotation marks within string values should be single quotes (''). Example: "personal": "This is an 'example'."
\\
\\
Important Guidelines:

1. Each function in the provided list includes a name, description, similar functions, and parameters.

2. Each parameter has a description, type, and a flag indicating whether it is required. Required parameters must be assigned a non-empty value; optional parameters may be assigned a non-empty value or left empty. No parameters should be omitted.

3. If a parameter is of type "dict", the "value" field specifies its internal structure and sub-fields. Ensure all sub-fields are populated. For non-dict parameters where "value" is not "non-enumerable", select one value (for types str or bool) or multiple values (for type list) from the "value" options as appropriate.

4. Do not introduce new functions or parameters.

5. Analyze the full meaning of the instruction carefully. Determine whether multiple steps or a combination of functions are necessary to achieve the intended functionality. If multiple functions are required, list them in the order of execution. Use the minimal number of functions possible to fulfill the instruction.

6. If no suitable function matches the instruction, return an empty list for the "function" field.

7. For unused or unknown parameters within a function, retain them as empty strings.

8. While internal reasoning may be conducted in Chinese, the final output must retain the original English names for functions and parameters.
\end{tcolorbox}

\begin{tcolorbox}[
    colback=blue!2,
    colframe=blue!70,
    colbacktitle=gray!40,
    coltitle=black,
    fontupper=\small\ttfamily,
    fonttitle=\bfseries\large,
    float*,
    breakable,                    
    enhanced,                      
    floatplacement={!tbh},
    width=\textwidth,
    title=Prompt for Three-stage Review
]
\textbf{Prompt for Agent Review}
\\
\\
You are a data generation expert responsible for reviewing and refining data related to mobile proactive intelligence tasks. Mobile proactive intelligence refers to the capability of mobile devices to proactively identify user needs and provide recommendations based on user behavior, environmental context, and other relevant information. You will receive a complete data record and must evaluate its quality against rigorous criteria.
\\
\\
The data record contains the following fields:
- "trigger": A trigger is one or a series of instantaneous actions or scenarios detectable by the device. Upon detection, the device activates its built-in large model to determine whether proactive recommendations are necessary.

- "condition": Conditions represent information considered by the device after trigger detection to reach a final decision. Conditions may be instantaneous or persistent.

- "persona": A persona is a concise description of a user, including basic information, the reason why this person would perform the operation in the given scenario, and the individual's core needs or motivations.

- "phone": Device information, which may include battery level, charging status, mobile data, WiFi, Bluetooth, lock screen status, foreground application, installed third-party applications, time, language, dark mode, geographical location, SMS, push notifications, etc.

- "world": World information, which may include weather, holidays, traffic information, exchange rates, international news, etc.

- "trace": A timestamped sequence of device operations, represented as atomic actions such as clicks, swipes, button presses, etc.

- "think": The reasoning process inferring the recommendation based on available information.

- "expected\_recommendation": The expected recommendation outcome generated by the built-in large model after detecting triggers and conditions, considering the persona, device information, world context, and operation sequence. Note: This field may be an empty string, indicating that no proactive recommendation should be provided at that moment.
\\
\\
Evaluation criteria for data quality:

- Authenticity of persona, device, and world information: Persona and device attributes must be perceivable by the device. For example, user personality traits (e.g., introversion or extroversion) or physical device characteristics (e.g., phone case type or screen scratches) are imperceptible.

- Stereotyping avoidance in persona, device, and world information: If the expected recommendation is empty, persona, device, and world information must avoid stereotypes, mediocrity, or artificiality. For instance, personas such as "user has no fixed hobbies, is aimless, indecisive, has irregular eating habits, and lacks planning" are unacceptable. Similarly, world descriptions like "today's weather is unpredictable, alternating between good and bad, with fluctuating air quality" are inappropriate.

- Authenticity of trace information: Each trace record must be authentic and perceivable by the device. Traces must be clearly described, avoiding vague references (e.g., "some" or "a certain"). Each step must begin with explicit actions such as "click", "swipe", "long press", or "enter text." The device must remain active throughout each operation step and cannot be in a screen-off state.

- Appropriateness of recommendation timing: The recommendation should be suitable for issuance immediately upon completion of the final trace step. If the recommendation could have been provided at an earlier step, it is inappropriate. Similarly, if the recommendation would disrupt the user's ongoing activity, it is unsuitable.

- Optimality of the recommendation: If a more suitable, relevant, or useful suggestion could be provided in the given scenario, the current recommendation is inadequate.

- Balance between autonomous exploration and proactive recommendation: Carefully analyze whether a proactive recommendation should be issued. Data is invalid if a recommendation is provided when none should be given, or if no recommendation is provided when one is warranted.

- Suitability of the recommended action as a proactive intelligent recommendation: The action should reflect potential user needs and intentions. For example, "Open Taobao to purchase XXX basketball shoes" is reasonable because the intent is easily generated and perceived by the user. In contrast, "Open Taobao and add the third item I saw to the shopping cart" is unreasonable due to the ambiguous and uncertain nature of ``the third item." Similarly, "Open Taobao to change profile picture" is invalid because such intent is difficult to perceive clearly.

- Concreteness and executability of the recommended action: The predicted task should not be suggestive but rather consist of specific, executable actions on the device. For example, instead of "Recommend visiting a highly-rated gallery 400 meters away and entering before 16:00 to enjoy art market discounts," it should be "Navigate the user to the highly-rated gallery 400 meters away and assist with ticket purchase."

- Avoidance of overly brief recommendations: If the user can complete the action with only one or two clicks on the current interface, a proactive recommendation is unnecessary.

- Other potential concerns as deemed appropriate.
\\
\\
You must evaluate the provided data against these criteria using stringent standards. Your output should be a JSON string in the following format, with no additional content.

\begin{lstlisting}[language={}]
Output example:
{

    "is_reasonable": true/false,
    
    "reason": "Reason for validity/invalidity"
    
}

Input data:
{data}
\end{lstlisting}
\vspace{1em}
\textbf{Prompt for Repairing Data}
\\
\\
You are a professional data repair specialist responsible for correcting mobile operation-related data records.

You will perform repairs based on the following four information categories:

- User profile

- Device status

- World information

- Modification suggestions or identified issues (manually annotated)
\\
\\
Your objectives:

1. Strictly modify the data according to the "modification suggestions or identified issues";

2. Ensure corrected content fully aligns with facts demonstrated in screenshots;

3. Maintain semantic, logical, and operational coherence;

4. Modify only problematic components without arbitrarily altering correct elements;

5. Output structure must remain consistent with original data fields;

6. Output must be valid and parseable JSON;

7. If "modification suggestions" contain ambiguity, contradictions, or logical conflicts, add a "note" field explaining the rationale while still providing a reasonable corrected version.
\\
\\
Screenshot consistency rules (mandatory):

- "Charging status": Determined by battery icon symbols;

- "Battery percentage": Only filled when numerical values are clearly visible in screenshots;

- If screenshot shows charging symbol without numerical percentage → Indicate "charging" only, without fabricating percentages;

- If screenshot displays battery percentage without charging symbol → Specify battery percentage without charging description;

- Screenshot content takes highest priority; no speculative generation permitted.
\\
\\
Language and output requirements:

1. State facts directly using natural language;

2. Avoid parenthetical explanations, reasoning, or screenshot judgment rationale;

3. Exclude programmatic expressions (e.g., "is\_charging=True");

4. Omit unactivated states when applicable;

5. Retain critical states (charging, brightness, network, Bluetooth, mode switches, etc.);

6. Maintain concise, accurate, and objective style;

7. Output must be strict JSON format without additional text or Markdown.

\begin{lstlisting}[language={}]
Output format example:

{
    "user_information": "xxx",
    "device_information": "xxx",
    "world_information": "xxx"
}

User profile: {user_info}
Device status: {device_info}
World information: {world_info}
Modification suggestions or identified issues: {issues}
\end{lstlisting}

Perform precise data correction based on the above information and factual evidence from screenshots.

If conflicts exist between screenshots and input information, prioritize screenshot content for modifications.

Output only in the specified JSON format.
\end{tcolorbox}

\begin{tcolorbox}[
    colback=blue!2,
    colframe=blue!70,
    colbacktitle=gray!40,
    coltitle=black,
    fontupper=\small\ttfamily,
    fonttitle=\bfseries\large,
    float*,
    breakable,                    
    enhanced,                      
    floatplacement={!tbh},
    width=\textwidth,
    title=Prompt for Training/Inference
]
<Role>

You are a professional mobile intelligent assistant evaluation expert, specializing in analyzing the decision-making processes of mobile proactive recommendation systems. Your task involves predicting user operational intentions based on multi-dimensional information and determining whether to proactively recommend relevant functions to users.
\\
\\
Your output consists of three sequential stages:

1. <think>: First, articulate your thinking process using natural language.

2. <rec>: Based on the thinking process, state the recommended action in natural language.

3. <function>: Translate the aforementioned action into a structured function call sequence.

Crucially, the content of <rec> must correspond precisely to the outcome of <function>, with <function> serving as the exact structured representation of <rec>.

</Role>
\\
\\
<Task>

Conduct comprehensive analysis based on the following four input categories:

- \textbf{device status}: Current device status, network environment, battery status, etc.  
\vspace{0.2em}
- \textbf{world information}: Time, location, weather, schedule arrangements, and other external environmental factors.  
\vspace{0.2em}
- \textbf{Behavioral Trajectories}: Recent operation sequences and behavioral patterns.  
\\
\\
Your responsibilities include:

1. Conducting in-depth analysis of users' genuine needs and intentions

2. Evaluating the urgency and clarity of identified requirements

3. Determining the necessity for proactive recommendations

4. If recommendations are warranted, selecting appropriate execution sequences from the provided function pool
\\
\\
**Critical Guidelines**:

- Return function sequences only when recommendations can be fully implemented using the provided functions

- Return an empty list if no recommendation is necessary or implementation through existing functions is unfeasible

- Function sequences must be arranged in execution order

</Task>
\\
\\
<Analysis\_Framework>

Adhere to the following analytical framework:
\\
\\
1. **Requirement Identification**: Identify potential user needs from behavioral trajectories, cross-validating with profile, environmental, and world context information (particularly time, location, and schedule)

2. **Urgency Assessment**: Determine whether requirements necessitate immediate attention by assessing if factors such as current time, schedule arrangements, or device status (e.g., low battery, weak network) constitute urgent triggering conditions

3. **Function Matching**: Verify whether requirements can be completely satisfied using existing functions in the pool—all function parameters must be fillable without missing critical elements

4. **Recommendation Decision**: Based on the above analysis, decide whether to recommend and determine recommendation content—recommend only when requirements are explicit, urgent, and fully executable through available functions

</Analysis\_Framework>
\\
\\
<Output\_Format>

    Strictly adhere to the following output structure:
    
    <think>Your reasoning process</think><rec>Your natural language recommendation decision</rec>
    <function>Corresponding structured function call JSON</function>

</Output\_Format>
\\
\\
<Output\_Requirements>

     **<think> Section Requirements**:
     
    The reasoning process must include deep analysis of user requirements, incorporating reasoning based on the provided multi-dimensional information.
\\
\\
    **<rec> Section Requirements**
    
   Recommended actions must be specific and executable on mobile devices within limited steps. Predicted tasks should not be suggestive but rather consist of concrete, executable action sequences on mobile devices.
\\
\\
    **<function> Section Requirements (JSON Format)**
\begin{lstlisting}[language={}]    
```json
{{

    "model_recommendation": [ 
    
        {{
        
            "name":"function_name_1",
            
            "parameters": {{
            
                "param1": "value1",
                
                "param2": "value2",
                
                "param3": ""
                
            }}
            
        }},
        
        {{
        
            "name":"function_name_2",
            
            "parameters": {{
            
                "param1": "value1",
                
                "param2": "",
                
                "param3": ""
                
            }}
            
        }}
        
    ]
}}
```
\end{lstlisting}
**JSON Format Requirements**:

- All keys and string values must be enclosed in double quotes ("")

- Key-value pairs are separated by commas, with no comma after the last pair

- Quotes within JSON strings use single quotes ('')

- Comments (// or /* */) are prohibited

- Each feature must include "name” and "parameters” fields

- Parameters must only contain essential arguments for feature execution; optional parameters without values may be set to empty strings "" or omitted entirely

- If no recommendation is needed, <rec> is 'No Recommendation’ and the model\_recommendation field is an empty array [].

Consistency Enforcement:

- The function sequence in <function> must exactly match the operations described in <rec>.

- If <rec> is marked "Not Recommended” but <function> contains content → output is invalid.

- If <rec> contains recommended content but the corresponding function is not found in the available function list, the output <function> should be an empty list
\\
\\
Function Requirements:

- Each function in our provided list has a name, description, similar functions, and parameters.

- Each parameter has a corresponding description, type, and mandatory status. Non-mandatory parameters may be omitted; mandatory parameters must be filled.

- If a parameter type is "dict”, the "value” field must specify the parameter's exact structure and fields. Ensure every field is fully populated. If the parameter type is not “dict” and ‘value’ is not "cannot be enumerated”, select one value (for str or bool types) or multiple values (for list types) from "value” based on the parameter type.

- Do not add any new functions or parameters.

- Carefully analyze the complete meaning of the instruction, considering whether multiple steps or combined functions are needed to achieve the functionality. If the instruction requires multiple functions to complete, list all relevant functions in the order they are executed.

- If the instruction cannot be mapped to a suitable function, leave the "function” field as an empty list [].

- For unused or unknown parameters within a function, leave them empty.

</Output\_Requirements>
\begin{lstlisting}[language={}]
<Input_Data>

### Reference Data:

{json.dumps(each_data['reference_information'], ensure_ascii=False)}

### Available Function List:
{json.dumps(function_pool, ensure_ascii=False)}
</Input_Data> 
\end{lstlisting}
Carefully analyze the input data (particularly all visible information in image data), conduct reasoning based on the analytical framework, and strictly output recommendation results in JSON format.

\end{tcolorbox}

\end{document}